\documentclass[12pt]{article}
\usepackage[margin=1in]{geometry}

\usepackage{makecell,colortbl}
\usepackage{microtype}
\usepackage{graphicx}
\usepackage{subfigure}
\usepackage{booktabs} 

\usepackage{times,rotating}
\usepackage{url}
\usepackage{booktabs, multicol, multirow}
\usepackage{morenotations2}
\usepackage{caption}

\usepackage{hyperref}
\usepackage[capitalize,noabbrev]{cleveref}
\usepackage{natbib}

\usepackage[capitalize]{cleveref}

\title{\papertitle}
\author{
  Richard Nock \qquad Mathieu Guillame-Bert \\
 Google Research\\
{\normalsize $\{$richardnock,gbm$\}$@google.com} \\
}
\begin{document}

\date{}

\maketitle

\begin{abstract}
While Generative Adversarial Networks (GANs) achieve spectacular results on unstructured data like images, there is still a gap on \textit{tabular data}, data for which state of the art \textit{supervised learning} still favours to a large extent decision tree (DT)-based models. This paper proposes a new path forward for the generation of tabular data, exploiting decades-old understanding of the supervised task's best components for DT induction, from losses (properness), models (tree-based) to algorithms (boosting). The \textit{properness} condition on the supervised loss -- which postulates the optimality of Bayes rule -- leads us to a variational GAN-style loss formulation which is \textit{tight} when discriminators meet a calibration property trivially satisfied by DTs, and, under common assumptions about the supervised loss, yields "one loss to train against them all" for the generator: the $\chi^2$. We then introduce tree-based generative models, \textit{generative trees} (GTs), meant to mirror on the generative side the good properties of DTs for classifying tabular data, with a boosting-compliant \textit{adversarial} training algorithm for GTs. We also introduce \textit{copycat training}, in which the generator copies at run time the underlying tree (graph) of the discriminator DT and completes it for the hardest discriminative task, with boosting compliant convergence. We test our algorithms on tasks including fake/real distinction, training from fake data and missing data imputation. Each one of these tasks displays that GTs can provide comparatively simple -- and interpretable -- contenders to sophisticated state of the art methods for data generation (using neural network models) or missing data imputation (relying on multiple imputation by chained equations with complex tree-based modeling).
\end{abstract}

\section{Introduction}\label{sec-int}

Generative Adversarial Networks have early established a gold standard for both neural networks as generative models \textit{and} the loss to train generative models via a variational measure-based distortion \citep{gpmxwocbGA,nctFG,ncmqwFG}. While they have achieved spectacular results on a variety of unstructured data \citep{nkhnML}, the quality of outcomes on \textit{tabular data} is still lagging behind with the sentiment that new approaches are needed \citep{cshOT}. This is an important problem: recently, tabular data was still representing the most prevalent data type in real world AI \citep[pp. 15]{cmm+NF}.  Interestingly, this chasm separating astonishing generation on unstructured data to suboptimal generation on tabular data mirrors another one, \textit{on the supervised side}, where neural nets can achieve superhuman recognition on unstructured data \citep{lll+SC} but require massive amounts of sophistication to compete against standard libraries using decision-tree (DT) based models on tabular data \citep{apTA}. DT induction has been perfected over decades, starting with core supervised loss functions known as \textit{proper} \citep{sEO,rwID}, using particularly fit and simple graph-based \textit{tree models} \citep{bfosCA,qC4}, and culminating with a powerful algorithmic machinery to learn them, \textit{boosting} \citep{kmOT,fhtAL,ssIB}. One would expect that potential generative approaches for tabular data would "mirror" those three key components on the generative side, but to our knowledge, none has been achieved. Such is our objective, and our paper thus contains three main technical contributions:

\textbf{On losses}, the GAN approach formulates the generator's loss from a variational measure-based divergence, unveiling the discriminator's loss \citep{nctFG}. Instead, we start from the \textit{discriminator's} side and a general proper loss, \textit{i.e.} a loss for which Bayes prediction is optimal, which is standard for DT induction since \citet{bfosCA}. We relate the corresponding information \citep{dgUI} to a GAN-style formulation which provides us with the generator's loss. A difference with GANs' variational formulation is there is \textit{no slack} in the characterisation if the discriminator meets a calibration condition trivially satisfied by DTs: unlike \textit{e.g.} \citet[Ineq. (4)]{nctFG}, we get identities all the way through. A surprising corollary follows. If the discriminator's \textit{partial losses} meet a property that most popular choices meet, then to minimize the generator's loss, it is \textit{sufficient} to minimize the $\chi^2$ between real and fake data: we get \textit{one} loss to "train generators against them all". This first contribution \textit{is not specific to DTs} as it holds for all calibrated discriminators in the properness framework.

\textbf{On models}, we introduce generative trees (GTs). In the same way as generator and discriminator in GANs include a similar functional form (a neural net), our GTs include a tree (graph) structure like DTs, differences being stochastic activations at the arcs and leaf-dependent data generation.

\textbf{On algorithms},  we propose a top-down induction algorithm to adversarially train GTs with provable boosting-compliant geometric convergence of the $\chi^2$, the weak generative learning assumption being a weak statistical dependence between the generator and discriminator. We propose a second way to train generative trees, extremely efficient and that we think has no equivalent yet in neural networks. In this setting, that we nickname \textit{copycat}, the generator tracks and copies the discriminator's tree (graph) at training time, and completes it for the hardest generative model given the discriminator\footnote{The generator turns out to compute boosting's balanced distribution of \citet{kmOT}.}. The geometric convergence in density ratio loss of the generator \citep{moLL} directly follows from a seminal result of \citet{kmOT}. 

In order not to laden this draft, we then summarise four series of experiments on missing data imputation, training from synthetic data, fake/real discrimination and synthetic data augmentation. Experiments were made on a series of domains including simulated domains and domains from the UCI, Kaggle and the Stanford Open Policing project (experiments are given \textit{in extenso} in an Appendix, \supplement, also containing all proofs). The experiments display that GTs can be very efficient contenders against sophisticated state of the art methods: on fake/real discrimination, GTs tend to get better results than neural networks (\textsc{ct-gan}s, \citet{xscvMT}) and on missing data imputation, GTs can beat on low-dimensional problems the \texttt{mice} approach \citep{vgMM}, even when \texttt{mice} relies on tree-based imputation using thousands+ of tree models -- against a GT essentially relying on a single one.

\section{Basic definitions}\label{sec-def}

$\forall k\in \mathbb{N}_*$, we let $[k] \defeq \{1, 2, ..., k\}$.
$\mathcal{X}$ denotes a domain, $\mathcal{S} \defeq \{\ve{x}_i : i\in [m]\} \subset \mathcal{X}$ is a sample of real observations. The associated supervised learning problem is a binary labeled problem where labels $\mathcal{Y} \defeq \{-1,1\} \defeq \{\mbox{fake}, \mbox{real}\}$ distinguish between a fake and a real observation. The objective of the supervised problem is to learn a \textit{posterior} computing $\pr[\Y = 1 | \X]$, denoted $\posterior \in [0,1]^{\mathcal{X}}$. With slight variations, many notations follow from \cite{rwID}. $\prior \defeq \pr[\Y = 1]$ is the prior. In the generative game, the prior is user-fixed.
$({\mathcal{X}}, \meas{P})$ and $({\mathcal{X}}, \meas{N})$ are measure spaces for 'positive/real' and 'negative/fake' observations respectively -- to avoid notation overloads, we leave implicit the $\sigma$-algebra. $({\mathcal{X}} \times \{-1,1\}, \meas{D})$ is the product measure space of labeled examples following the (supervised) \textit{binary task} $(\prior, \meas{P}, \meas{N})$ \citep[Section 4]{rwID}; we let $\binartask \defeq (\prior, \meas{P}, \meas{N})$
for short. We also have the \textit{mixture} space $({\mathcal{X}}, \meas{M})$ with $\meas{M} \defeq \prior \cdot \meas{P} + (1-\prior) \cdot \meas{N}$. A posterior is particularly interesting for $\binartask$, \textit{Bayes posterior}, which is:
    \begin{eqnarray}
      \bayespo & = & \prior\cdot \frac{\dmeas{P}}{\dmeas{M}},\label{defBAYES}
    \end{eqnarray}
and is optimal for \textit{proper} losses (more on this in Section \ref{sec-loss}).

\section{Models}\label{sec-models}

\begin{figure}
  \centering
  \begin{tabular}{cc}
\hspace{-0.6cm}     \includegraphics[trim=80bp 50bp 140bp 170bp,clip,width=0.5\columnwidth]{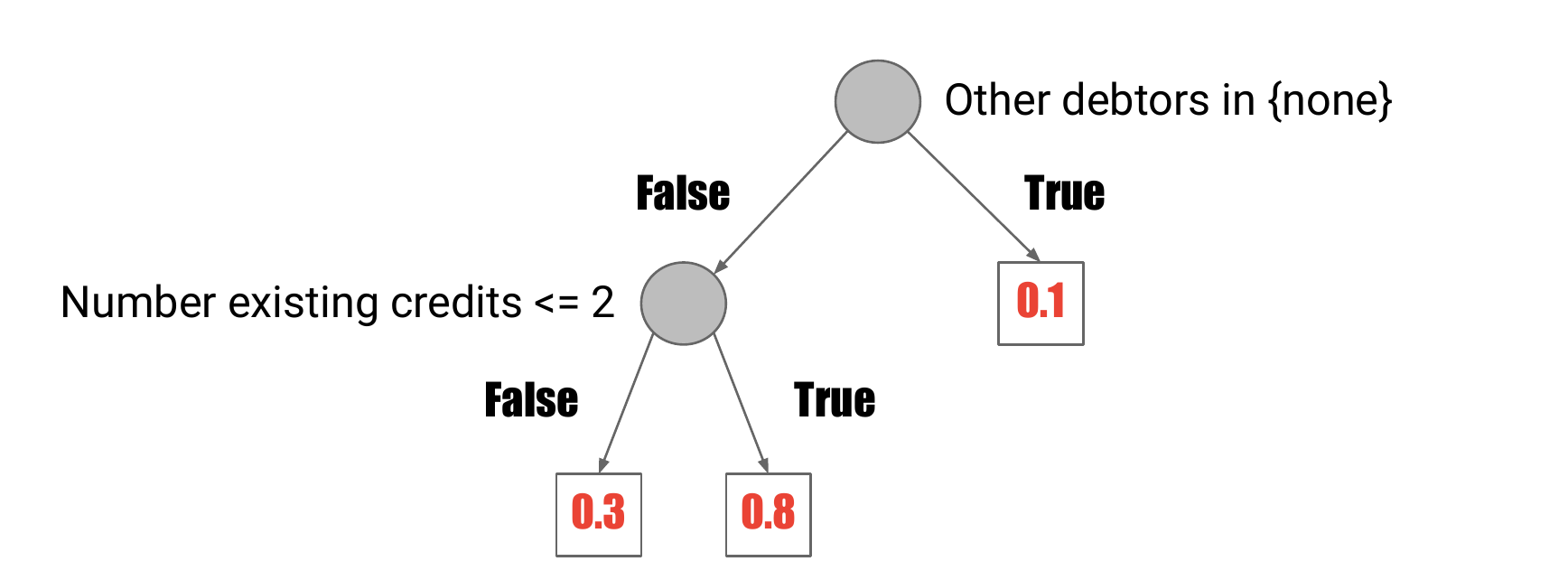} & \hspace{-1.05cm} \includegraphics[trim=80bp 50bp 140bp 170bp,clip,width=0.5\columnwidth]{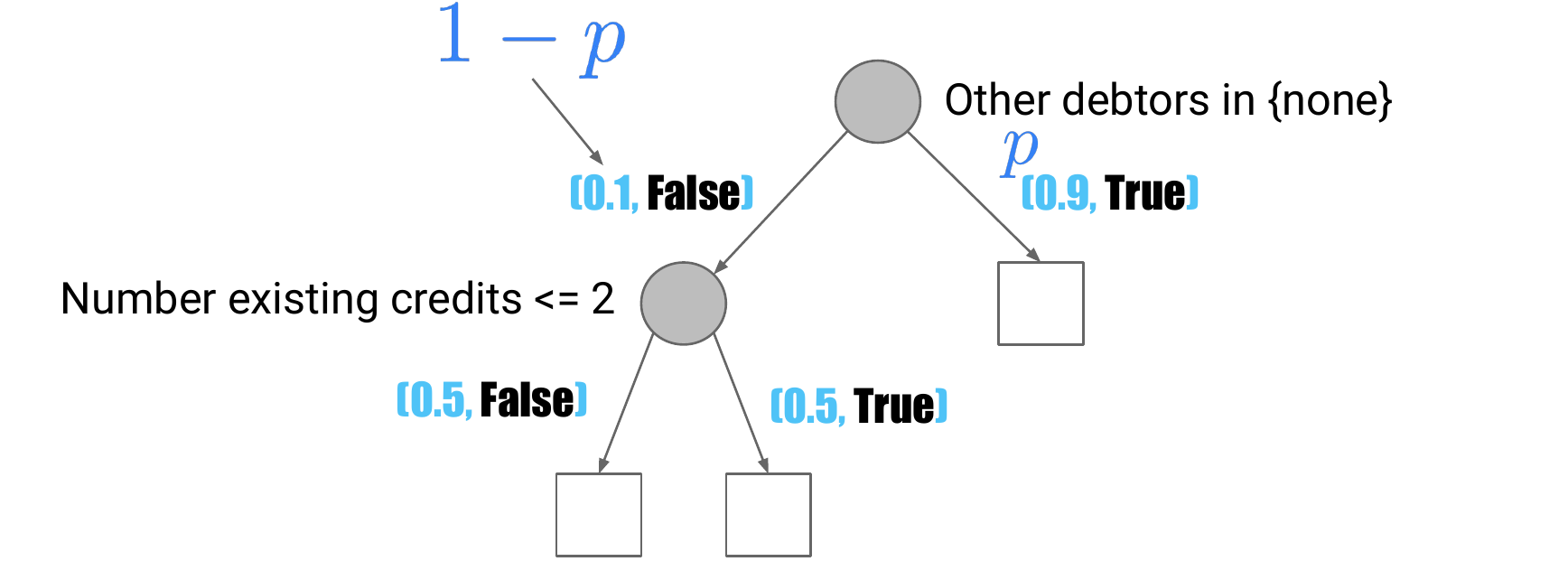}
  \end{tabular}
    \caption{Decision tree (left), generative tree (right) with the same underlying tree (graph).}
    \label{fig:dt-gt-trees}
  \end{figure}

We present the tree-based models we use as architectures for both the discriminator and the generator. 
\paragraph{Architectures} we start by the commonpoint between both, that we denote a tree for short.
  \begin{definition}\label{def-tree}
A \textbf{tree} is a rooted, directed binary tree whose internal nodes are labeled with binary tests over observation variables and outgoing arcs are labeled with truth values. For any internal node, the left outgoing arc is labeled with truth value \texttt{false} and the right outgoing arc is labeled with \texttt{true}. Leaves are blank nodes.
\end{definition}
  \begin{definition}\label{def-dt-gt}
A \textbf{decision tree} (DT) is a tree with leaves labeled in $[0,1]$. A \textbf{generative tree} (GT) is a tree in which truth values at arcs are associated to Bernoulli events $\meas{B}(.)$.
\end{definition}
Figure \ref{fig:dt-gt-trees} presents examples of DT and GT with the same underlying tree. We assume without loss of generality that trees are
binary but our definitions could trivially be extended to trees of any arity.
Hereafter, low caps like $h$ are used to represent DTs while high-caps like $G$ are used to represent GTs. $\leafset(.)$ denotes the set of leaves of a tree.
\paragraph{Access routines} an important routine needed for a DT $h$ is, for any observation $\ve{x}$, the leaf $\leaf(\ve{x}) \in \leafset(h)$ reached by $\ve{x}$. This is the leaf whose path from the root involves tests satisfied by $\ve{x}$. If $\ve{x}$ contains no unknown feature values, this path is unique. 
The main access routine for a GT $G$ is the generation of an observation. To do so, we simply stochastically traverse the tree using the Bernoulli events at the internal nodes. Once a leaf $\leaf\in\leafset(G)$ is reached, sampling an observation is done by a uniform sampling in the complete domain \textit{that satisfies the tests traversed} to reach $\leaf$. In Figure \ref{fig:dt-gt-trees}, the center leaf $\leaf$ of the GT $G$ is reached with probability $q = 0.1 \cdot 0.5 = 0.05$. If we do reach it, according to the UCI German Credit data domain \citep{dgUM}, 
then we sample uniformly at random an observation for which attribute 'Number existing credits' is in $\{0, 1, 2\}$ and 'Other debtors' is in $\{$co-applicant, guarantor$\}$ \textit{and} all other attributes are chosen uniformly at random in their full domain, since they do not appear in the path to $\leaf$. 
\begin{remark}\label{rem-unif}
  Uniform sampling imposes a finite length domain for real or integer features, which is a reasonable assumption for standard features like \textit{e.g.} age, salary. Alleviating the constraint can be done using specific transformations, such as the Box-Muller transform, generating Normal deviates from uniform distributions \citep{bmAN}. 
  \end{remark}

\section{Loss functions involved} \label{sec-loss}

Departing from (W)GAN-style approaches, we design the losses involved from the discriminator's. 
\paragraph{Calibrated posteriors} For any function $f \in \mathbb{R}^{\mathcal{X}}$ and measure $\meas{Q}$ over measurable space $({\mathcal{X}}, \sigmaalgebra)$, $\meas{Q}_{f}$ is the restriction of $\meas{Q}$ to the sub-$\sigma$-algebra $\sigmaalgebra_{f}$ induced by the level set of $f$. A similar notation $\meas{Q}_{f}$ with identical definition is used in \citet[Section II]{vehRD}. It can be interpreted as the marginal of $\meas{Q}$ on the subset of events of $\sigmaalgebra_f$, each of which is a union of events from $\sigmaalgebra$ having the same $f$-value. We now define a property of a posterior for class probability estimation that shall be fundamental to analyse our losses.
\begin{definition}\label{defCAL}
  Posterior $\calposterior$ is said \textit{calibrated} with task $\binartask$ (or just calibrated for short) iff $\calposterior = \prior\cdot \frac{\dmeas{P}_{\calposterior}}{\dmeas{M}_{\calposterior}}$, and we let $\binartask_{\calposterior} \defeq (\prior, \meas{P}_{\calposterior}, \meas{N}_{\calposterior})$.
\end{definition}
There are three important examples of calibrated posteriors:
\begin{itemize}
\item [{[1]}] \textit{the constant posterior} $\constantpo \defeq \prior$ is (the only constant) calibrated (posterior). To see it, it yields $\Omega_\constantpo = \{\emptyset, \mathcal{X}\}$. The RHS in Def. \eqref{defCAL} gives $\prior \cdot \dmeas{P}_{\constantpo}(\mathcal{X}) / \dmeas{M}_{\constantpo}(\mathcal{X}) \defeq \prior \cdot \int_{\mathcal{X}} \dmeas{P} / \int_{\mathcal{X}} \dmeas{M} \defeq  \prior \cdot 1 / 1 = \prior$ and we have for this posterior $\pr[\Y | \X] = \pr[\Y | \X \in \mathcal{X}]= \constantpo\defeq \prior$;
\item [{[2]}] \textit{Bayes posterior} $\bayespo$ is calibrated; it follows from \eqref{defBAYES};
\item [{[3]}] let $h$ be a DT. Any DT induces a partition of $\mathcal{X}$ at its leaves, $\{\mathcal{X}_\leaf: \leaf \in \leafset(h)\}$; suppose without loss of generality that all leaves' predictions are different, and consider $\Omega_h = \{\emptyset\} \cup \{\mathcal{X}_\leaf: \leaf \in \leafset(h)\}$. Without further correction, \textit{the posterior prediction} at a leaf $\leaf$ of $h$ is classically computed as the ratio of the total weight of real observations reaching $\leaf$ over the total weight of observations reaching $\leaf$. In mathematical form, we have here $\pr[\Y | \X] = \pr[\Y | \X \mbox{ reaches } \leaf] = \prior \dmeas{P}_{h} / \dmeas{M}_h$, which is by definition the prediction of $\posterior$ at $\leaf$ and shows that the posterior prediction of any DT is calibrated.
  \end{itemize}
We note that [1] is a particular case of [3] when $h$ is reduced to its root, and if the domain $\mathcal{X}$ is finite, then [2] is a particular case of [3] for $h$ being \textit{any} complete (finite) DT. Hereafter, a tilda like $\calposterior$ denotes a calibrated posterior. Notation $\posterior$ denotes any posterior, disregarding eventual additional properties.
\begin{table}[t]
  \centering
  \resizebox{\columnwidth}{!}{
    \begin{tabular}{cc|acc}\hline\hline
      Loss & $\partialloss{-1}(u)$ & \multicolumn{3}{|a}{$\drloss(\densrat), \densrat \geq 0$} \\
           & & Eq. & cvx & $\searrow$ \\ \hline
      Log & $-\log_2(1-u)$ & $\log_2\left(1+\frac{1}{\densrat}\right)$ & \tickYes  & \tickYes  \\
      Square & $\frac{u^2}{2}$ & $\frac{1}{2} \cdot \left(\frac{1}{1+\densrat}\right)^2$ & \tickYes & \tickYes   \\
      Matusita & $\sqrt{\frac{u}{1-u}}$ & $\frac{1}{\sqrt{\densrat}}$ & \tickYes & \tickYes  \\\hline
      Jeffreys & $2\cdot\left( \log\left(\frac{u}{1-u}\right)+\frac{1}{1-u}\right)$ & $2\cdot\left( \log\left(\frac{1}{\densrat}\right)+\frac{1}{\densrat} + 1\right)$ & \tickYes & \tickYes  \\
      KL & $2\cdot\left(2\log(1-u) + \frac{u}{1-u}\right)$ & $2\cdot \left(2\log\left(\frac{\densrat}{1+\densrat}\right) + \frac{1}{\densrat}\right)$ & \tickYes & \tickYes  \\ 
      Normalized$^{*}$ $\chi^2$ & $2\cdot \frac{u^2}{(1-u)^2}$ & $\frac{2}{\densrat^2}$ & \tickYes & \tickYes  \\ \hline\hline
    \end{tabular}
    }
      \caption{Differentiable partial losses for class $-1$ for symmetric losses, along with their corresponding $\drloss$ and its properties  (see text; cvx = convex). (*) add +2 to Pearson $\chi^2$ to have $\partialloss{-1}(0)=0$.}
    \label{tab:crit}
  \end{table}
\paragraph{Losses for class probability estimation} A \textit{loss for class probability estimation}, $\loss : \mathcal{Y} \times [0,1]
\rightarrow \mathbb{R}$, is expressed as
\begin{eqnarray}
\loss(y,u) & \defeq & \iver{y=1}\cdot \partialloss{1}(u) +
                     \iver{y=-1}\cdot \partialloss{-1}(u), \label{eqpartialloss}
\end{eqnarray}
where $\iver{.}$ is Iverson's bracket \citep{kTN}. Functions $\partialloss{1}, \partialloss{-1}$ are called \textit{partial} losses. A loss is
\textit{symmetric} when $\partialloss{1}(u) = \partialloss{-1}(1-u),
\forall u \in [0,1]$ \citep{nnOT} and \textit{differentiable} when both partial losses are differentiable. Table \ref{tab:crit} presents examples partial losses of symmetric losses.
The pointwise conditional risk of posterior ${\posterior} \in [0,1]$ with respect to ground truth $\optpo \in [0,1]$ is $\poirisk({\posterior},\optpo) \defeq  \expect_{\Y \sim \bernoulli(\optpo)}\left[\loss(\Y,{\posterior})\right]$, \textit{i.e.},
\begin{eqnarray}
  \poirisk({\posterior},\optpo) & = & \optpo\cdot \partialloss{1}({\posterior}) + (1-\optpo)\cdot \partialloss{-1}({\posterior}). \label{eqpoirisk}
\end{eqnarray}
$\mathrm{B}(.)$ denotes a Bernoulli for picking label $\Y = 1$.
The associated (pointwise) \textit{Bayes} risk is 
\begin{eqnarray}
\poibayesrisk(\optpo) & \defeq & \inf_{{\posterior}} \poirisk({\posterior},\optpo). \label{pbr}
\end{eqnarray}
The interesting case is when the argument of the $\inf$ reduces to $\{\optpo\}$, because then minimizing \eqref{eqpoirisk} for $\posterior$ 'encourages' to pick ground truth $\optpo$. Formally, when (i) $\poibayesrisk(\posterior) = \poirisk(\posterior,\posterior), \forall \posterior \in [0,1]$ and (ii)  $\poirisk({\posterior},\optpo) > \poibayesrisk(\optpo), \forall {\posterior}\neq \optpo$, we say that the loss is \textit{strictly proper}, and \textit{proper} when (i) holds. The population version of \eqref{eqpoirisk}, when both $\posterior, \optpo \in [0,1]^{\mathcal{X}}$, is the (\textit{full}) risk \citep[pp 747]{rwID},
\begin{eqnarray}
  \poprisk({\posterior},\bayespo, \meas{M}) & \defeq &  \expect_{\X \sim \meas{M}} \left[\poirisk({\posterior}(\X),\bayespo(\X))\right]\label{eqpoprisk}.
\end{eqnarray}
We now assume that all losses for class probability estimation used hereafter are strictly proper, symmetric and differentiable (\spsd) and satisfy the additional technical assumption that $\partialloss{-1}(0) = 0$ (all but Jeffreys in Table \ref{tab:crit} are \spsd), which makes $\optpo$ Bayes posterior in \eqref{defBAYES}.
\begin{definition}\label{def-si-posterior}
  The \textbf{information} of calibrated $\calposterior \in [0,1]^{\mathcal{X}}$ is:
  \begin{eqnarray*}
    \statinf(\calposterior,  \meas{M})  \defeq  \popbayesrisk (\constantpo ,  \constantpo ,  \meas{M})  -  \popbayesrisk (\calposterior,  \calposterior,  \meas{M})  =  \poibayesrisk(\prior)   -   \popbayesrisk (\calposterior,  \calposterior,  \meas{M}).
    \end{eqnarray*}
  \end{definition}
  This definition is a convenient restriction to calibrated posteriors of the original definition in \citet[Eq. (2.2)]{dgUI} and \citet[Eq. (20)]{rwID}. It represents how much 'information' $\calposterior$ brings compared to the constant calibrated posterior $\constantpo$. Decision tree induction would traditionally maximize $\statinf(\calposterior, \meas{M})$ \textit{via} the minimisation of some $\popbayesrisk (\calposterior, \calposterior, \meas{M})$, where $\calposterior$ is the calibrated posterior at the leaves of the decision tree: CART's uses the square loss \citep{bfosCA}, C4.5 uses the log-loss \citep{qC4}, etc. .
  \paragraph{Losses for measure estimation and binary task information} A substantial body of work has tightened the GAN loss to variational $f$-divergences \citep{nctFG,ncmqwFG}. Here, we are also interested in such a formulation \textit{but} for a very specific set of $f$ introduced decades ago \citep[Theorem 2]{ovSI}:
\begin{eqnarray*}
  \fprior (t) \defeq \poibayesrisk(\prior) - (\prior t + 1 - \prior)\cdot\poibayesrisk\left(\frac{\prior t}{\prior t + 1 - \prior}\right), \forall t\in \mathbb{R}_+,\label{deffprior}
\end{eqnarray*}
which involves prior $\prior$ (under control in the generative game).
\begin{definition}\label{def-si-binartask}
  The \textbf{information} of binary task $\binartask \defeq (\prior, \meas{P}, \meas{N})$, $\idiv{}(\binartask)$, is the $\fprior$-divergence
\begin{eqnarray}
  \idiv{}(\binartask) & \defeq & \idiv{\fprior}(\meas{P}, \meas{N}), \label{defBININF}
\end{eqnarray}
where we recall $\idiv{\fprior}(\meas{P}, \meas{N}) \defeq \int \fprior\left(\frac{\dmeas{P}}{\dmeas{N}}\right) \dmeas{N}$. 
\end{definition}
Given $\idiv{}(\binartask)$, we could directly dig into the variational formulation of the $\fprior$-divergence to design the generative modelling game and loss at the expense of an eventual slack due to the variational argument \citep[Ineq. (4)]{nctFG}. We avoid the slack \textit{via} a trick using calibrated posteriors.
\paragraph{Losses for the adversarial generative game} We need to define two additional functions, for any posterior $\posterior$:
\begin{eqnarray} 
  \densrat \defeq \frac{1-\posterior}{\posterior} \quad;\quad \likelihood \defeq \frac{1}{\densrat} \cdot \frac{1-\prior}{\prior} \label{defDENSRATANDRATORI}.
\end{eqnarray}
We call $\densrat$ the \textbf{density ratio} and $\likelihood$ the \textbf{likelihood ratio}, following conventions in \citet[pp 746]{rwID}\footnote{Names can otherwise vary in the literature.}. To take an example, if we consider Bayes posterior $\bayespo$ in \eqref{defBAYES}, then it follows $\bayesli = \dmeas{P} / \dmeas{N}$, justifying the name. Let $f^\star(z) \defeq \sup_t \{zt - f(t)\}$ denote the convex conjugate of $f$. We note that for any \spsd~loss $\loss$, $\fprior$ is differentiable.
\begin{definition}\label{def-gd-risks}
  Let $\binartask$ and $\likelihood$ be any binary task and likelihood ratio. Let $\genloss(z) \defeq (f^\prior)^\star\circ {f^\prior}' (z)$ and
  \begin{eqnarray}
    \genrisk(\meas{N}| \likelihood) & \defeq & -\expect_{\X \sim \meas{N}}[\genloss(\likelihood (\X))]\label{defGENRISK},\\
  \disrisk(\likelihood| \binartask) & \defeq & -\left(\expect_{\X \sim \meas{P}}\left[{f^\prior}' \circ \likelihood (\X)\right] + \genrisk(\meas{N}| \likelihood)\right)\label{defDISRISK}
  \end{eqnarray}
  respectively denote the \textbf{generator and discriminator risks}.
\end{definition}
For any $f$-divergence, we have the $f$-GAN defining inequality \citep[eqs. (4-6)]{nctFG} similar to -\eqref{defDISRISK}:
\begin{eqnarray}
  \idiv{f}(\meas{P}, \meas{N}) \geq \sup_{\likelihood} \{\expect_{\X \sim \meas{P}}\left[f' \circ \likelihood (\X)\right]  - \expect_{\X \sim \meas{N}}[ f^\star \circ  f' \circ \likelihood (\X)]\},\label{vfGAN}
\end{eqnarray}
so both \eqref{defGENRISK} and \eqref{defDISRISK} define the corresponding functions to minimise for the generator and discriminator in this variational inequality after the change $f \rightarrow \fprior $. While the change is anecdotical with respect to the inequality \eqref{vfGAN}, it conceptually operates a radical shift with respect to classical ($f$-)GANs: the generator's loss is completely determined in our case \textit{by the loss of the discriminator} as it appears in $\fprior$, a loss whose design heavily relies on properness. The change also has a key fortunate mathematical consequence: we can replace the inequality \eqref{vfGAN} by a \textit{chain of equalities} involving all key risks, as we now show.
\begin{theorem}\label{th-all-main}
  For any \spsd~loss $\ell$, any binary task $\binartask$, any calibrated posterior $\calposterior$ whose likelihood ratio is denoted $\callikelihood$, the following holds:
  \begin{eqnarray}
  \statinf(\calposterior, \meas{M}_{\calposterior}) = -\disrisk(\callikelihood| \binartask _{\calposterior}) = \idiv{}(\binartask_{\calposterior}). \label{reducgen2}
  \end{eqnarray}
  Furthermore, we have the expression for function $\genloss$ in Definition \ref{def-gd-risks}:
  \begin{eqnarray}
\genloss(\likelihood) & = & - \poibayesrisk(\prior) + (1-\prior)\cdot \drloss(\densrat),\label{eqgenloss}
  \end{eqnarray}
  with
  \begin{eqnarray*}
\drloss(\densrat) & \defeq & \partialloss{-1}\left(\frac{1}{1+\densrat}\right),
    \end{eqnarray*}
  and the transformation $\likelihood \leftrightarrow \densrat$ is obtained \textit{via} \eqref{defDENSRATANDRATORI}.
\end{theorem}
The proof (in \sm, Section \ref{proof-th-all-main}) also provides the conjugate $({\fprior})^{\star}$, of potential independent interest.
\begin{remark}\label{rem-whyCalibrated}
Since $f$-divergences satisfy the data processing inequality, we also have for any calibrated posterior $\calposterior$, $\idiv{}(\binartask_{\calposterior}) \leq \idiv{}(\binartask)$.
  Together with \eqref{reducgen2}, this gives a precise way of how the GAN game operates with calibrated posteriors and proper losses: training a discriminator to maximise its statistical information $\statinf(\calposterior, \meas{M}_{\calposterior})$, \textit{e.g.} as done with DT induction algorithms, increases as well the information of the binary task $\idiv{}(\binartask_{\calposterior})$. On the other hand, training in turn the generator to minimize $\genrisk(\meas{N}_{\calposterior}|.)$ reduces the information of the binary task $\idiv{}(\binartask_{\calposterior})$. In the case of DT algorithms, as the tree grows, its calibrated posterior $\calposterior$ converges to an 'empirical Bayes' best posterior (based on training real data). Disregarding generalisation issues, as long as the generator 'stands' the growth of the discriminator by keeping $\idiv{}(\binartask_{\calposterior})$ small enough, it is guaranteed to improve with iterations.
\end{remark}
\paragraph{A generative loss 'to learn against them all' (almost)} Table \ref{tab:crit} shows that $\drloss$ has several invariant properties for the losses shown. We formalise some of them.
  \begin{lemma}\label{lempropDRLOSSsmall}
    For any $\ell$ proper symmetric and differentiable, (i) $\drloss$ is decreasing and (ii) $\drloss$ is convex $\densrat$ or in $1/\densrat$, $\forall \densrat$.
  \end{lemma}
  Proof in \sm, Section \ref{proof-lempropDRLOSSsmall}. In the examples of Table \ref{tab:crit}, the 'or' in Lem. \ref{lempropDRLOSSsmall} is in fact an 'and'. Convexity is important because it yields a single loss to efficiently train the generator against any 'proper' trained discriminator: the $\chi^2$, \textit{i.e.} the $f$-divergence whose generator is $f(t) \defeq (t-1)^2$.
  \begin{lemma}\label{lembinf-lossg-ub}
    For any \spsd~loss $\ell$ for which $\drloss$ is convex, any binary task $\binartask$, calibrated posterior $\calposterior$ (likelihood ratio $\defeq \callikelihood$), the following bound holds on the generator's risk $\genrisk$:
    \begin{eqnarray}
     \genrisk(\meas{N}_{\calposterior}| \callikelihood) & \leq & \poibayesrisk(\prior)- (1-\prior)\cdot \partialloss{-1}\left(\frac{\prior}{1+(1-\prior)\cdot \chi^2(\meas{N}_{\calposterior}||\meas{P}_{\calposterior})}\right) \label{boundGenrisk}.
    \end{eqnarray}
  \end{lemma}
  Proof in \sm, section \ref{proof-lembinf-lossg-ub}. For any \spsd~loss, $\partialloss{-1}$ is increasing (\textit{Cf} proof of Theorem \ref{th-all-main}). Therefore, if we train the generator to reduce $\chi^2(\meas{N}_{\calposterior}||\meas{P}_{\calposterior})$, it reduces the RHS in \eqref{boundGenrisk} and provides a smaller bound on the generator's risk, \textit{regardless} of the proper loss used as long as $\drloss$ is convex.

\section{Training $h$ and $G$}\label{sec-train}

\begin{figure*}[h]
  \centering
  \begin{tabular}{c|c|c|c|c}\hline\hline
    \includegraphics[trim=520bp 30bp 53bp 145bp,clip, height=0.17\textheight]{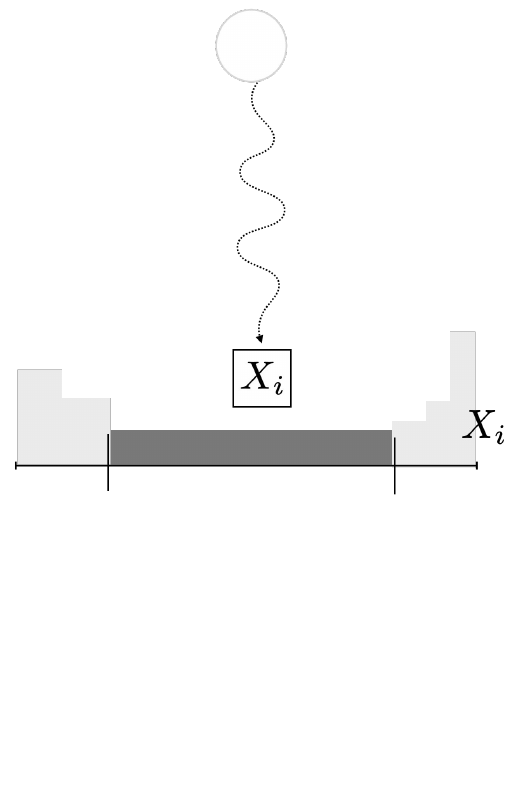}& \includegraphics[trim=520bp 30bp 53bp 145bp,clip, height=0.17\textheight]{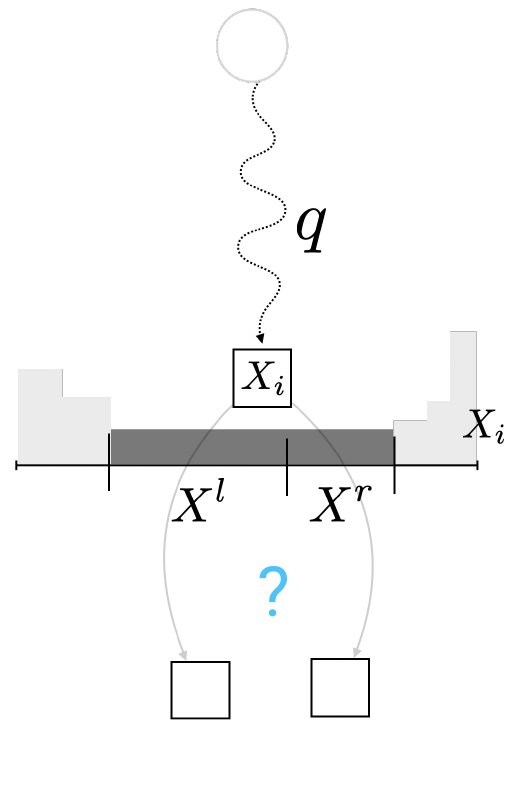} & \includegraphics[trim=520bp 30bp 20bp 145bp,clip, height=0.17\textheight]{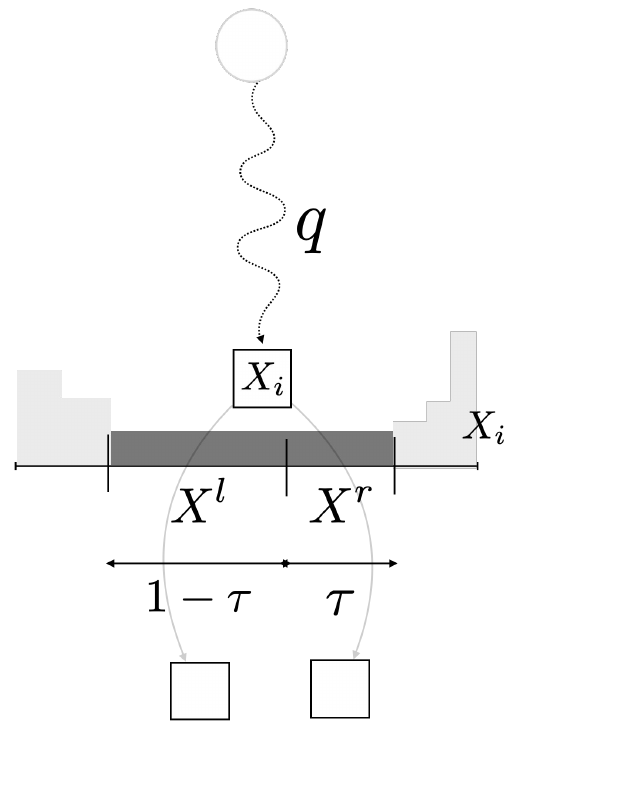} & \includegraphics[trim=520bp 30bp 20bp 145bp,clip, height=0.17\textheight]{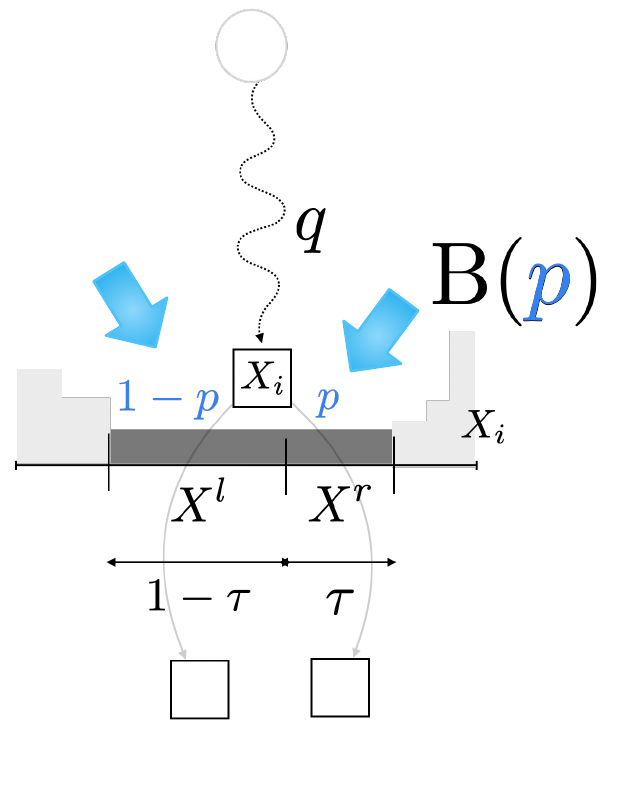}& \includegraphics[trim=520bp 30bp 20bp 145bp,clip, height=0.17\textheight]{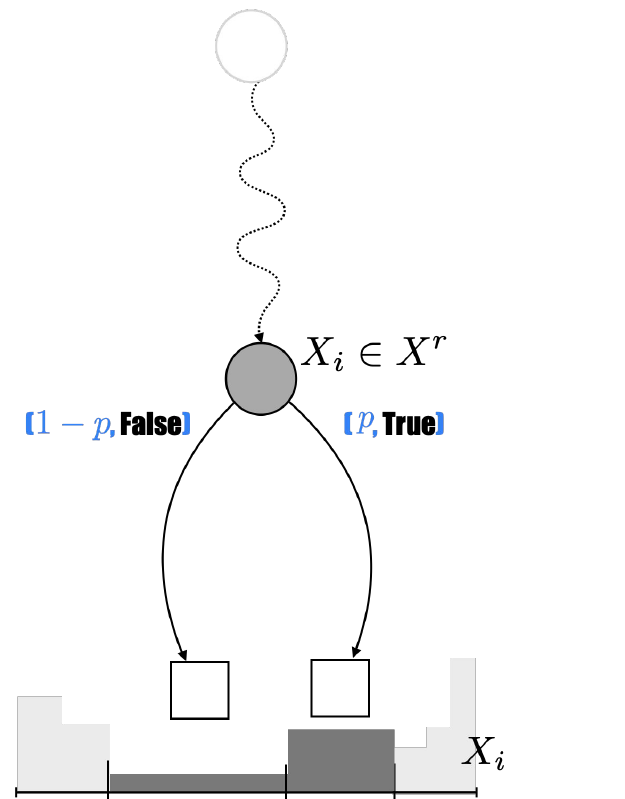}\\
    (A) & (B) & (C) & (D) & (E)\\ \hline\hline
    \end{tabular}
    \caption{Splitting a sampling leaf in $G$ to create a subtree with two new sampling leaves. (A): we pick a current leaf and decide on a variable $X_i$ whose local density (therefore uniform, in dark gray) is going to be split in two at the new leaves. (B) a potential split creates two local intervals $X^l, X^r$, and we can compute the relative local proportion of examples that would be generated from $X^r$ ($\tau$) and from $X^l$ ($1-\tau$), (C). Finally, we compute Bernoulli's $p$ (D). Note that $\tau$ does not appear after split (E), it is just used to compute $p$.}
    \label{fig:gt-topdown}
  \end{figure*}

The most popular way to train both the DT $h$ and the GT $G$ is to proceed as in generative adversarial networks \citep{gpmxwocbGA}. The DT can be trained using any commercial package \citep{bfosCA,qC4} or more generally any greedy induction of a tree minimizing a \spsd~loss with convex $\drloss$. We can then train the adversarial GT by minimizing $\chi^2(\meas{N}_{\calposterior}||\meas{P}_{\calposterior})$ and alternate between phases of training the DT and training the GT. We call this setting '\textit{adversarial}' for short.
Due to the architecture of the models, there is a more specific training available for generative trees, more constrained than the adversarial setting but with a straightforward implementation and direct convergence guarantees coming from the convergence of the DT training. In this case, the GT \textit{copies} the tree architecture of the DT and fits the probabilities to keep $\chi^2(\meas{N}_{\calposterior}||\meas{P}_{\calposterior}) = 0$. We call this setting the '\textit{copycat}' setting. We detail them.

\subsection{Adversarial training of the generator}

We adopt a greedy induction of the GT. The current calibrated posterior of the generator $h$ is $\calposterior$. Let $\lambda$ denote a general leaf of $h$. $\samplingnode$ denotes the current sampling node at the generator $G$ that we are going to split to create a subtree with two sampling leaves and associated Bernoulli probability $p$ to compute the new arcs at $\samplingnode$. Figure \ref{fig:gt-topdown} provides an overview of the process, pointing to a new variable, $\tau$, which is the local (relative) proportion of examples generated from the right sub-domain at the candidate split. For any $\leaf \in \leafset(h)$ in the discriminator and candidate split at leaf $\samplingnode \in \leafset(T)$ in the generator, we define:\\

  \noindent $\hookrightarrow$ $p_\leaf$, the total weight of real examples reaching $\leaf$;\\
  \noindent $\hookrightarrow$ $n_{\leaf} \defeq \int_{\domain{\leaf}}\dmeas{N}_{\calposterior}$, the theoretical proportion of fake examples reaching $\leaf$, where $\domain{\leaf}$ is the subset of $\mathcal{X}$ of observations that reach $\leaf$ in $h$;\\
  \noindent $\hookrightarrow$ $n^0_\leaf$, the total weight of fake examples reaching $\leaf$ \textit{but} generated by $\leafset(G) \backslash \{\samplingnode\}$ -- these weights do not change after the split at $\samplingnode$;\\
  \noindent $\hookrightarrow$ $n^l_\leaf$, the total weight of fake examples reaching $\leaf$, generated by $\samplingnode$ \textit{and} whose value for attribute $X_i$ (the one considered for the split) is in $X^l$;\\
  \noindent $\hookrightarrow$ $n^r_\leaf$, the total weight of fake examples reaching $\leaf$, generated by $\samplingnode$ \textit{and} whose value for attribute $X_i$ (the one considered for the split) is in $X^r$.\\
  
It is worth noticing that $n_{\leaf}, n^0_\leaf, n^l_\leaf, n^r_\leaf$ can all be calculated \textit{exactly} from the trees of $h$ and $G$. After the split, 'only' the proportions in $\cup_\leaf \{n^l_\leaf\} \cup_\leaf \{n^r_\leaf\} $ are potentially changed by the split. We can compute $\tau$ as a function of these quantities:
    \begin{eqnarray}
\tau & = &  \frac{\sum_{\leaf \in \leafset(h)} n_\leaf^r}{\sum_{\leaf \in \leafset(h)} n_\leaf^l+n_\leaf^r}.
    \end{eqnarray}
    Define three more quantities:
    \begin{eqnarray*}
l_\leaf \defeq {n}^0_\leaf + \frac{{n}^l_\leaf }{1-\tau} ; r_\leaf \defeq {n}^0_\leaf + \frac{{n}^r_\leaf }{\tau}; \vardelta_\leaf \defeq r_\leaf - l_\leaf .
    \end{eqnarray*}
    These quantities are interpretable as follows: $l_\leaf$ would be the new $n_{\leaf}$ after split \textit{if} we were to pick $p=0$; $r_\leaf$ would be the new $n_{\leaf}$ after split \textit{if} we were to pick $p=1$ and $\vardelta_\leaf $ quantifies the difference in generation between these two extreme strategies. These strategies \textit{are} extreme because for example if we choose $p=0$, then we discard the support at $\samplingnode$ covering observations whose value for $X_i$ is in $X^r$. Some coefficients are particularly important to compute $p$:
    \begin{eqnarray}
\mu_{\LL\LL} \defeq \sum_{\leaf \in \leafset(h)} \frac{l_\leaf^2}{p_\leaf} ; \mu_{\R\R} \defeq \sum_{\leaf \in \leafset(h)} \frac{r_\leaf^2}{p_\leaf} ; \mu_{\LL\R} \defeq \sum_{\leaf \in \leafset(h)} \frac{l_\leaf r_\leaf}{p_\leaf} .\label{eqALLMU}
    \end{eqnarray}
    These are also interpretable: if we let $\chi^2\left(\meas{N}'_{\calposterior}(p)||\meas{P}_{\calposterior}\right)$ denote the new $\chi^2$ after the split at $\samplingnode$ with Bernoulli $p$, then $\mu_{\LL\LL} = 1 + \chi^2\left(\meas{N}'_{\calposterior}(0)||\meas{P}_{\calposterior}\right)$, $\mu_{\R\R} = 1 + \chi^2\left(\meas{N}'_{\calposterior}(1)||\meas{P}_{\calposterior}\right)$,
    and $\mu_{\LL\R} $ is a correlation between both strategies. The proof of Lemma \ref{lembinf-lossg-ub} shows those identities.
  \begin{algorithm}[t]
\caption{\topdowngen $(G,h)$}\label{splitMF}
\begin{algorithmic}
  \STATE  \textbf{Input:} current generator $G$, current discriminator $h$;
  \STATE  \textbf{Output:} $G$ with a new split;
  \STATE  Step 1 : pick $\samplingnode \in \leafset(G)$, $i\in [d]$; // leaf and variable for the current split
  \STATE Step 2 : choose $(X^l, X^r)$ and compute $\tau$; // split choice
  \STATE Step 3 : compute $p$ as
    \begin{eqnarray}
      p & \leftarrow & \clamp\left(\frac{\mu_{\LL\LL} -\mu_{\LL\R} }{\mu_{\LL\LL} +\mu_{\R\R} -2 \mu_{\LL\R}}\right); \label{defP}
    \end{eqnarray}
    \STATE Step 4 : replace $\samplingnode$ by a split as designed in Steps 1,2 w/ Bernoulli probability $p$ as in \eqref{defP};
\end{algorithmic}
\end{algorithm}
Algorithm \topdowngen~summarizes the steps to split one leaf, without giving specific constraint on the choice of leaf to split $\samplingnode$, feature $i$, and split parameters $(X^l, X^r)$. We leave these open because general convergence rates can be obtained for \topdowngen ~that do not constraint those choices. Function $\clamp(z)$ is
\begin{eqnarray*}
  \clamp(z) & \defeq & \max\{\min\{z,1\},0\}.
\end{eqnarray*}
We have two different regimes for the convergence of the $\chi^2$, depending on whether $p\in(0,1)$ or $p\in \{0,1\}$ (that latter case means that we \textit{discard support} for the generation of examples). We give those results in two different Theorems. For our first Theorem, let $\mathbb{I}_\updelta \defeq [0, \updelta)$ denote a range of 'acceptable' values for the $\chi^2$s. The question we ask is what is the guaranteed convergence rate when we are \textit{not} in this favorable case, that is, when the $\chi^2$s (before and after update) are not in $\mathbb{I}_\updelta$, a situation we refer to as \topdowngen~being 'outside regime $\mathbb{I}_\updelta$'. We show \topdowngen~exhibits geometric convergence rate related to $\updelta$ and the proximity of $p$ to $\tau$. 
    \begin{theorem}\label{thboost1}
      Suppose $p\in (0,1)$. For any $\epsilon > 0, \updelta > 0$, if (i) \topdowngen~is outside regime $\mathbb{I}_\updelta$ and (ii) $p$ in Step 3 satisfies $|\tau-p| \geq \epsilon$, then after one iteration of \topdowngen, we have:
    \begin{eqnarray}
\chi^2\left(\meas{N}'_{\calposterior}(p)||\meas{P}_{\calposterior}\right) & \leq & \frac{1}{1+\updelta \epsilon^2} \cdot \chi^2\left(\meas{N}_{\calposterior}||\meas{P}_{\calposterior}\right).\label{sumCase1S}
    \end{eqnarray}
  \end{theorem}
Condition (ii) does make sense because if $p = \tau$, then there is no change in the $\chi^2$ as $\chi^2\left(\meas{N}'_{\calposterior}(\tau)||\meas{P}_{\calposterior}\right) = \chi^2\left(\meas{N}_{\calposterior}||\meas{P}_{\calposterior}\right)$. To cover the case $p\in \{0,1\}$, we need an additional assumption that mirrors the weak learning assumption that governs the convergence of the discriminator $h$ in the boosting framework. We call it a \textit{weak generating assumption}.
\begin{definition} \label{defWGA}(\textbf{$\updelta$-WGA}) Let $\updelta > 0$ be a constant. We say that the split at $\samplingnode$ meets the $\updelta$-Weak Generating Assumption iff $\mu_{\D\D} \geq \updelta \cdot \max\{\mu_{\LL\LL}, \mu_{\RR\RR} \}$, where $\mu_{\D\D} \defeq \sum_{\leaf \in \leafset(h)} \delta_\leaf^2/p_\leaf$. 
\end{definition}
  \begin{theorem}\label{thboost2}
Suppose $p\in \{0,1\}$ and the $\updelta$-WGA holds. Then after one iteration of \topdowngen, we have:
    \begin{eqnarray*}
\chi^2\left(\meas{N}'_{\calposterior}(p)||\meas{P}_{\calposterior}\right)  \leq \frac{1}{1+\updelta (\tau+(1-2\tau)p)^2} \cdot \chi^2\left(\meas{N}_{\calposterior}||\meas{P}_{\calposterior}\right).
    \end{eqnarray*}
  \end{theorem}
  \begin{figure}
  \centering
    \includegraphics[trim=60bp 20bp 480bp 235bp,clip, height=0.17\textheight]{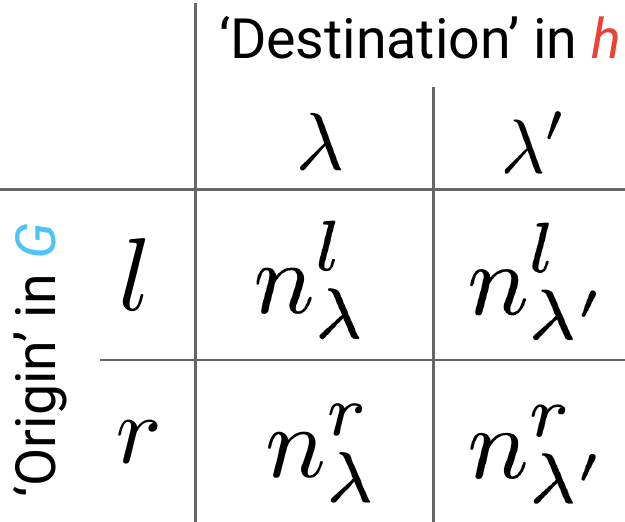}
    \caption{2$\times$2 contingency table for the $\chi^2$ independency test on which the 'weak' generative assumption relies, see Remark \ref{remWGA}.}
    \label{fig:chi2-ind}
  \end{figure}
  \begin{remark}\label{remWGA}
   Definition \ref{defWGA} is 'weak' in a generative sense. Indeed, the only case where $\mu_{\D\D} = 0$ is when ${n}^l_\leaf /(1-\tau) = {n}^r_\leaf / \tau, \forall \leaf \in \leafset(h)$, which brings after solving for $\tau$, the relationships ${n}^r_\leaf/( {n}^r_\leaf + {n}^l_\leaf) = {n}^r_{\leaf'}/( {n}^r_{\leaf'} + {n}^l_{\leaf'})$ for any two leaves in $\leafset(h)$, and after simplifying, ${n}^r_\leaf \cdot{n}^l_{\leaf'} = {n}^l_\leaf \cdot{n}^r_{\leaf'}, \forall \leaf, \leaf' \in \leafset(h)$. Filling any 2$\times$2 contingency table with 'destination' leaves in the discriminator ($\leaf, \leaf'$) versus 'provenance' in the generator ($l, r$) (Figure \ref{fig:chi2-ind}) immediately leads to a Pearson's $\chi^2 = 0$. What the WGA prevents is thus the extreme independence where the generator's examples would be randomly attributed to the leaves in the discriminator.
 \end{remark}
 
\subsection{Copycat training of the generator}

\paragraph{Algorithm} When using a (decision) tree as discriminator, both the GT and DT have an underlying tree (graph). Copycat training takes advantage of this scenario as the GT $G$ copies the tree of the DT $h$ as it is learned: if $h$ involves the usual top-down induction scheme, after each of the new splits in $h$, the generative tree $G$ replicates the same split in its tree, computing the Bernoulli probabilities in such a way that the new proportion of fake observations is going be the \textit{same} as that of real observations at the new leaves of $h$. In other words, after the update of $G$, the new $h$ performs as badly as a fair coin.\\
We see two substantial downsides to copycat vs adversarial training: the generator 'peeks' in the discriminator's tree, which can be a problem for privacy or fairness issues, and it has zero freedom to grow its own tree. There is, however, a major upside of copycat training over adversarial training: it requires no additional expensive computation for the new Bernoulli's $p$ in $G$. Denote $m_\leaf$ the number of positive examples at the leaf $\leaf \in \leafset(h)$ to be split in $h$, and $m_\leaf^r$ the number of positive examples ending up in its right sub-leaf after split. Then in $G$ we have at the same $\leaf$: $p = m_\leaf^r/m_\leaf$.
\paragraph{Convergence} There is another benefit of copycat training: in the boosting model of \citet[Section 5.1]{kmOT}, the convergence rates for $G$ towards the distribution of observed real data directly follow \textit{from the boosting rates of $h$ on the supervised task}. To show this, we proceed in two steps: the first introduces and conveniently decomposes a risk quantifying the discrepancy between measures in the density ratio model \citep{moLL} -- for this objective, we introduce indexes in notations, and let $h_T$ denote the generator $h$ after $T$ splits, so that $h_0$ is the single-root DT. Similarly, we let $\calposterior_T$ denote the corresponding calibrated posterior and $\meas{P}^T$ denote the measure induced on $\mathcal{X}$ by $\meas{P}$ and $h_T$ by (i) ensuring it is locally uniform at each leaf and (ii) it locally sums to the local weight of $\meas{P}$. In equation, it satisfies, $\meas{U}$ denoting the uniform measure,
\begin{eqnarray}
  \frac{\dmeas{P}^T}{\dmeas{U}} (\ve{x})& = & \frac{\int_\leaf \dmeas{P}}{\int_\leaf \dmeas{U}}, \forall \ve{x}\mbox{ reaching }\leaf.
\end{eqnarray}
Denote $p_\leaf \defeq \int_\leaf \dmeas{P}$ and $u_\leaf \defeq \int_\leaf \dmeas{U}$. For differentiable and convex $F: \mathbb{R} \rightarrow \mathbb{R}$, the Bregman divergence with generator $F$ is $B_F(z\|z') \defeq F(z) - F(z') - (z-z')F'(z')$. Given function $g : \mathbb{R} \rightarrow \mathbb{R}$, the generalized perspective transform of $F$ given $g$ is \citep{mOAI,mOAII,nmoAS} $\check{F}(z) \defeq g(z) \cdot F  \left(\frac{z}{g(z)}\right)$.
$g$ is implicit in notation $\check{F}$. 
\begin{definition}\label{riskLH}
  The \textbf{Likelihood ratio risk} of $\meas{P}^T$ with respect to $\meas{P}$ for \spsd~loss $\ell$ is (with $g(z) \defeq z + (1-\prior)/\prior$):
  \begin{eqnarray*}
    \likelihoodratiorisk_\ell\left(\meas{P}, \meas{P}^T\right) & \defeq & \prior \cdot \expect_{\meas{U}} \left[ B_{\perspectivepobayesrisk}\left(\frac{\dmeas{P}}{\dmeas{U}}\left\|\frac{\dmeas{P}^T}{\dmeas{U}}\right. \right) \right].
  \end{eqnarray*}
\end{definition}
Risks expressed as in Def. \ref{riskLH} have a history in density ratio estimation \citep{moLL} (and references within). 
\begin{lemma}\label{it-to-dr}
  $\forall$ \spsd~loss $\ell$ and DT $h_T$, $\perspectivepobayesrisk$ is strictly convex and
  \begin{eqnarray*}
    \likelihoodratiorisk_\ell\left(\meas{P}, \meas{P}^T\right)  & = & \idiv{\fprior}(\meas{P}, \meas{U}) - \idiv{\fprior}(\meas{P}_{\calposterior_T}, \meas{U}_{\calposterior_T}).
  \end{eqnarray*}
\end{lemma}
The proof is given in \sm, Section \ref{proof-it-to-dr}. Strict convexity is crucial: in such a case a Bregman divergence zeroes iff its two arguments are equal, implying at the risk level in Def. \ref{riskLH} that $\likelihoodratiorisk_\ell\left(\meas{P}, \meas{P}^T\right) =0 $ iff $\meas{P} = \meas{P}^T$ almost everywhere. $\idiv{\fprior}(\meas{P}, \meas{U})$ is a constant and the data processing inequality satisfied by $f$-divergences brings $0 = \idiv{\fprior}(\meas{P}_{\calposterior_0}, \meas{U}_{\calposterior_0}) \leq ... \leq \idiv{\fprior}(\meas{P}_{\calposterior_T}, \meas{U}_{\calposterior_T}) \leq ... \leq \idiv{\fprior}(\meas{P}, \meas{U})$, 
so regardless of the top-down induction algorithm used for $h$, Lemma \ref{it-to-dr} shows a form of convergence of $\meas{P}^T$ towards $\meas{P}$ as accounted by the likelihood ratio risk $\likelihoodratiorisk_\ell\left(\meas{P}, \meas{P}^T\right)$. The last part of copycat training's convergence is to make those inequalities \textit{strict} with guaranteed slack: this is achieved using the boosting analysis of \citet{kmOT} \textit{as is}. \\

We do not put iteration indexes in $G$, assuming the one we consider is the one after the update of the last discriminator $h_T$. Denote $\leaf$ a leaf to be split in $h$ and $\meas{P}_\leaf$, $\meas{U}_\leaf$ the distributions conditioned to reaching $\leaf$; we denote as 'uniformly generated' the observations sampled from $\meas{U}_\leaf$. Define the local mixture $\meas{M}_\leaf \defeq \prior \cdot\meas{P}_\leaf + (1-\prior) \cdot\meas{U}_\leaf$ and the \textit{balanced} mixture, $\meas{M}'_\leaf$, is defined as $\meas{M}_\leaf \defeq (1/2) \cdot\meas{P}_\leaf + (1/2) \cdot\meas{U}_\leaf$. Let $\texttt{t}$ denote the predicate value of a split chosen for $\leaf$.
    \begin{definition}\citep{kmOT}\label{defKM}
Fix $\updelta > 0$. Predicate $\texttt{t}$ at leaf $\leaf$ satisfies the $\updelta$-Weak Hypothesis Assumption (\textbf{WHA}) iff $\Pr_{\meas{M}'_\leaf}[\texttt{t}(\X) \neq \Y] \leq 1/2 - \updelta$.
\end{definition}
It turns out that the balanced mixture \textit{is the one} against which each new split in $h_.$ is evaluated after the generator is updated in copycat training (the modifications at $G$ are local since the underlying tree defines a partition of $\mathcal{X}$). We use the WHA to ensure that the split chosen at any leaf during copycat training complies with Definition \ref{defKM}. Using a result of \citet{kmOT}, this brings guaranteed rates for the maximisation of the information of its calibrated posterior (Definition \ref{def-si-posterior}). Theorem \ref{th-all-main} then directly yields rates for the maximisation of $\idiv{\fprior}(\meas{P}_{\calposterior_T}, \meas{U}_{\calposterior_T})$, and Lemma \ref{it-to-dr} translates them to convergence for $\meas{P}^T = \meas{G}$ towards $\meas{P}$, where $\meas{G}$ denotes the measure induced by generator $G$ (equality $\meas{P}^T = \meas{G}$ is guaranteed by copycat training). We make those convergence rates explicit for the boosting-optimal splitting criterion, Matusita's loss (Table \ref{tab:crit}), for which $\poibayesrisk(u) = 2\sqrt{u(1-u)}$. For any $\varepsilon \in [0,1]$, we abbreviate $\poibayesrisk_{\varepsilon}(\binartask) \defeq \varepsilon \cdot \poibayesrisk(\prior) + (1-\varepsilon)\cdot\popbayesrisk (\bayespo, \bayespo, \meas{M})$.
\begin{theorem}
  Define the binary task $\binartask \defeq (\prior, \meas{P}, \meas{U})$, $\meas{M}$. Suppose \spsd~loss $\ell$ is Matusita's loss and the WHA is satisfied at each split of $h_.$. Then for any $\varepsilon \in [0,1]$, if the number of splits in $h_.$ satisfies
  \begin{eqnarray*}
    T & \geq & \left(\frac{1}{\poibayesrisk_{\varepsilon}(\binartask)}\right)^{\frac{32}{\updelta^2}},
  \end{eqnarray*}
  then the likelihood ratio risk achieved by generator $G$ with respect to the distribution of real observations satisfies
  \begin{eqnarray*}
    \likelihoodratiorisk_\ell\left(\meas{P}, \meas{G}\right) & \leq & \varepsilon \cdot \idiv{\fprior}(\meas{P}, \meas{U}).
  \end{eqnarray*}
\end{theorem}
  The proof directly follows from the proof of \citet[Theorem 10]{kmOT}, using Defns in \eqref{def-si-posterior}, \eqref{riskLH}, Thm \ref{th-all-main} and Lem. \ref{it-to-dr} to calibrate the bound that is needed on the information of $\calposterior_T$ to guarantee the bound on $\likelihoodratiorisk_\ell\left(\meas{P}, \meas{G}\right)$.

\bgroup
\def\arraystretch{0.2}
\begin{figure}[t]
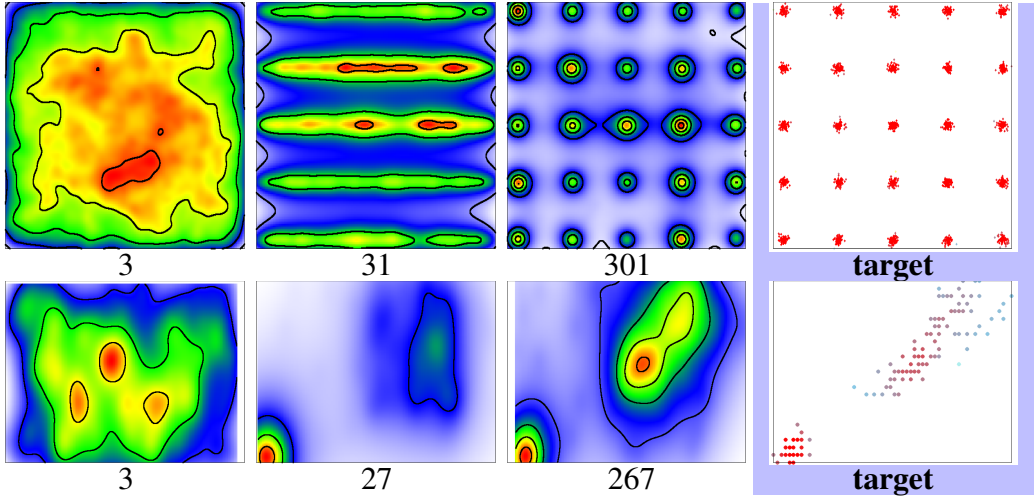

  \centering
  \begin{tabular}{cccc}
   \hspace{-0.3cm} \imagebigmain{gridgauss-density}{Sep_10th__13h_15m_4s}{3}{}{4000} \hspace{-0.3cm} &\hspace{-0.3cm}  \imagebigmain{gridgauss-density}{Sep_10th__13h_15m_4s}{31}{}{4000} \hspace{-0.3cm} &\hspace{-0.3cm}  \imagebigmain{gridgauss-density}{Sep_10th__13h_15m_4s}{301}{}{4000} \hspace{-0.3cm} & \cellcolor{blue!25}{\imagemodelmain{gridgauss/data-gridgauss-plot}}\\
    \hspace{-0.3cm}    3 \hspace{-0.3cm} & \hspace{-0.3cm} 31 \hspace{-0.3cm} &\hspace{-0.3cm}  301 \hspace{-0.3cm} &  \cellcolor{blue!25}{\textbf{target}}\\
    \hspace{-0.3cm} \imagebignosquaremain{iris}{Sep_15th__11h_35m_51s}{3}{_X_petal_length_Y_petal_width}{150} \hspace{-0.3cm} &\hspace{-0.3cm}  \imagebignosquaremain{iris}{Sep_15th__11h_35m_51s}{27}{_X_petal_length_Y_petal_width}{150} \hspace{-0.3cm} &\hspace{-0.3cm}  \imagebignosquaremain{iris}{Sep_15th__11h_35m_51s}{267}{_X_petal_length_Y_petal_width}{150} \hspace{-0.3cm} & \cellcolor{blue!25}{\imagemodelnosquaremain{iris/iris-petal-length-petal-width}}\\
     \hspace{-0.3cm} 3 \hspace{-0.3cm} & \hspace{-0.3cm} 27 \hspace{-0.3cm} & \hspace{-0.3cm} 267\hspace{-0.3cm}  & \cellcolor{blue!25}{\textbf{target}}
  \end{tabular}
    \caption{2D density heatmaps for coordinates $(x,y)$ on \texttt{gridGauss} (top) and (\texttt{petal-length}, \texttt{petal-width}) on \texttt{iris} (bottom). Index is the total number of nodes in the GT.}
    \label{fig:exp-gridgauss-iris-main}
  \end{figure}
  \egroup
  
\section{Experiments}\label{secexp}

We carried out experiments on four topics: missing data imputation, synthetic training (training on fakes vs real), synthetic discrimination (distinguishing fakes from real) and synthetic augmentation (adding fakes to real for training) on a total of 11 readily available datasets, from the UCI \citep{dgUM}, Kaggle and the Stanford Open Policing project, to which we added 4 simulated datasets. For simplicity, all GT experiments use copycat training, implemented in Java. We refer to \supplement, Section \ref{app-exps} for all details. Before embarking on a summary of the main findings, we provide example density plots (a classical rite of passage for generative models, \citet{xzzBG}) on one of our simulated domains (\texttt{gridGauss}, \citet{dbplamcAL}) and on the \texttt{iris} dataset, in Figure \ref{fig:exp-gridgauss-iris-main}. In general, we observe quite a good fitting of the observed data, even for real domains, and specific features about the true density can emerge quite early in the GT induction.

\subsection{Missing data imputation ('\textsc{impute}')}\label{exp-imp-main}

\bgroup
\def\arraystretch{1.0}
\begin{table*}
  \centering
  {\normalsize
\begin{tabular}{cr|lll?c|ll?c|ll}\Xhline{4\arrayrulewidth}
     \multicolumn{2}{c|}{\textsc{u}s vs \textsc{m}ice$|$} & \tagnorm & \tagcart & \tagrf &  & \tagcart & \tagrf & & \tagcart & \tagrf \\
    \multicolumn{2}{c|}{$\#$trees p. fold} & N/A & 10 & 1 000 & & 35 & 3 500 & & 125 & 12 500\\ \hline
    ($q$=)5$\%$ & \multirow{4}{*}{\parbox[h]{2mm}{\rotatebox[origin=c]{90}{{\small \texttt{circGauss}}}}} & \signif{\tagus}{0.003} & \tagus & \signif{\tagus}{0.07} &  \multirow{4}{*}{\parbox[h]{2mm}{\rotatebox[origin=c]{90}{{\small \texttt{led}}}}} & \tagus & \tagus  &  \multirow{4}{*}{\parbox[h]{2mm}{\rotatebox[origin=c]{90}{{\small \texttt{led24}}}}} & \tagthem & \tagus  \\
   10$\%$ & &  \signif{\tagus}{0.03} & \signif{\tagus}{0.002} & \signif{\tagus}{0.007} & &  \tagus & \tagthem  & &  \tagus & \tagthem  \\
   20$\%$ & &\signif{\tagus}{0.0005} & \signif{\tagus}{0.001} & \signif{\tagus}{0.001} & & \signif{\tagus}{0.05} & \tagus & & \tagus & \tagthem  \\
   50$\%$ & &\signif{\tagus}{0.04}  & \tagus & \tagus &  & \signif{\tagthem}{0.02} & \signif{\tagthem}{0.06} & & \signif{\tagthem}{0.01} & \signif{\tagthem}{0.07} \\ \Xhline{4\arrayrulewidth}
\end{tabular}
}
\begin{tabular}{ccc}
  \includegraphics[trim=60bp 0bp 60bp 0bp,clip,width=0.28\textwidth]{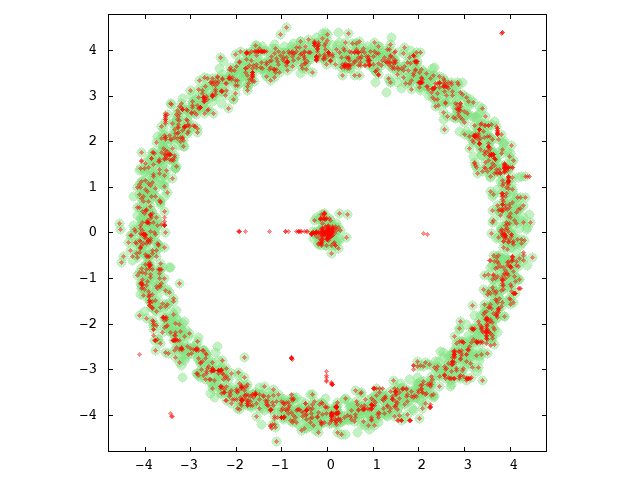} \hspace{-0.5cm}&\hspace{-0.5cm}  \includegraphics[trim=60bp 0bp 60bp 0bp,clip,width=0.28\textwidth]{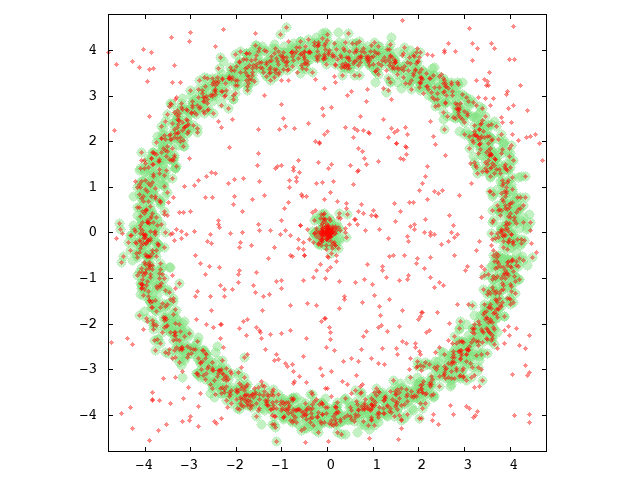} \hspace{-0.5cm}&\hspace{-0.5cm}  \includegraphics[trim=60bp 0bp 60bp 0bp,clip,width=0.28\textwidth]{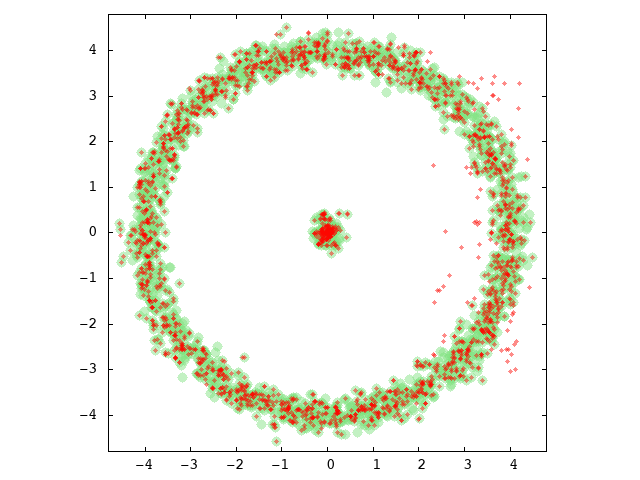}\\
  us \hspace{-0.5cm}&\hspace{-0.5cm} mice$|$\tagnorm \hspace{-0.5cm}&\hspace{-0.5cm} mice$|$\tagcart 
  \end{tabular}
\caption{\textit{Top}: best method (\tagus(s) vs \tagthem(ice) on three domains; $p$-values are shown if $p\leq 0.1$) on \textsc{impute}. \textit{Bottom}: \texttt{circGauss} plots, showing the domain data (green dots) vs imputed data for three methods (red dots, $q = 20\%$).}
\label{tab:imput-gridgauss-vs-mice-main}
\vspace{-0.2cm}
  \end{table*}
  \egroup
  \paragraph{Objective and experimental setting} A GT $G$ is not just useful to generate data: it can trivially be used for missing data imputation \citep{mjbcMD}. For this, we constrain the support of the tree to the observed variables and then sample in region(s) of maximal density. This costs no more than $\mathcal{O}(|\leafset(G)|)$ per observation. We compared against a few powerful alternatives (mostly tree-based) from the R \texttt{mice} package \citep{vgMM}. Such methods rely on round-robin prediction of missing values: after having initialized them, one circles several times (5 in our experiments) through predicting each column from all others using trained models from a specific \texttt{method}. We used \texttt{method} $\in\{$\textsc{norm}, \textsc{cart}, \textsc{rf}$\}$ (\textsc{rf} = random forests with 100 trees each, \textsc{cart} learns regression / decision trees \citep{bfosCA}). It is important to realise that even on a domain like \texttt{led24} with 25 variables, 5 round-robin iterations with \textsc{rf} implies using no less than \textbf{12 500} trees \textit{per fold} when we rely on \textbf{1} in our GT. We grow the GT to its \textsc{max} size (limit: 10 000 nodes) and prevent splits with $p\in \{0,1\}$, thus avoiding discarding support for data generation. We generate Missing Completely At Random (MCAR, \citet{vFIM}) data by removing a fixed proportion $q \in \{5\%, 10\%, 20\%, 50\%\}$, embedded in a 5-fold cross validation for each $q$. Imputing a complete dataset on each fold, we judge imputation's quality with optimal transport's Wasserstein's $W_2^2$ \citep{mjbcMD}.\\
  \paragraph{Results} Table \ref{tab:imput-gridgauss-vs-mice-main} summarises 3 domains (more in \supplement) giving a good panorama of our observations, the first of which being the fact that our simple approach can in fact \textit{beat} \texttt{mice} on problems with restricted number of variables (but the picture is reversed on domains with large number of variables, \supplement). The quality of imputations with respect to \texttt{mice} is clear from Table \ref{tab:imput-gridgauss-vs-mice-main}. While our objective was not to beat such fit-for-purpose imputation methods relying on a comparatively huge number of models, we remark that our results can serve as basis for using GTs in more sophisticated approaches.

  \subsection{Training on synthetic data ('\textsc{train-synth}')}\label{sec-train-synth-main}

  \paragraph{Objective and experimental setting} In this basic experiment, we seek to answer whether generated data can be used \textit{in lieu} of the original data to solve the original data's supervised / regression problem (\textit{e.g.} predicting the variety for \texttt{iris}). We use a 5-folds CV experiment where on each fold a supervised classifier is trained on fake or real data and then used to classify the (fresh) real data's fold. Fake data is obtained from a generator trained on the training data, then sampled for the same data size. We consider 3 GTs with different sizes, with 10, 300 and up to \textsc{max} splits (same as in Section \ref{exp-imp-main}). Our contender is the state of the art CT-GAN \citep{xscvMT}, trained for a number of epochs in $\{$10, 300, 1K$\}$; the original data's supervised problem is then solved by training \textsc{rf}s and gradient boosted decision trees (\textsc{gbdt}) on real or fake data, and comparing the output accuracy / RMSE (details in \supplement, Section \ref{sec-doms}). \\
\paragraph{Results} Table \ref{tab:train-synth-discrim-main} (left) provides a summary of the results on the 10 total domains considered, from which it emerges that when GTs have $600+$ nodes (300+ splits), we tend to beat neural networks (NNs, regardless of the number of epochs considered). What is worse for NNs is that in a total of 4 cases, they are statistically beaten by a \textit{uniform sampling} of the training data, which means they fail at learning the domain's characteristics. This never happens to GTs. Detailed 2D plots display that GTs tend to better learn domain-specific features. Also, the final size of the GT can be tiny compared to the training data, \textit{e.g.} less than $0.5\%$ on \texttt{dna} and \texttt{open policing} (see \supplement, Section \ref{sec-train-synt}).

  \begin{table}
    \centering
    \resizebox{\columnwidth}{!}{
      \begin{tabular}{c?c|c|c?c|c|c?c|c|c}\Xhline{4\arrayrulewidth}
         & \multicolumn{3}{c?}{\textsc{train-synth}} & \multicolumn{3}{c?}{\textsc{synth-discrim}} & \multicolumn{3}{c}{\textsc{synth-aug}} \\\hline
  \backslashbox{\taggt{.}}{\tagctgan{.}} & 10{\color{red}$\star$} & 300{\color{red}$\star$} & 1K{\color{red}$\star$}{\color{red}$\star$} & 10 & 300  & 1K & 10 & 300  & 1K \\  \hline\hline
  10 & \wtl{1}{7}{2} & \wtl{3}{4}{3} & \wtl{2}{3}{5}& \wtl{7}{4}{2} & \wtl{6}{5}{2} & \wtl{6}{5}{2} & \wtl{6}{1}{3} & \wtl{6}{2}{2} & \wtl{4}{1}{5}\\ \hline
  300 & \wtl{8}{1}{1} & \wtl{8}{1}{1} & \wtl{4}{5}{1}& \wtl{8}{4}{1} & \wtl{8}{3}{2} & \wtl{8}{3}{2}& \wtl{9}{0}{1} & \wtl{9}{0}{1} & \wtl{6}{2}{2}\\ \hline
  \textsc{max} & \wtl{8}{1}{1} & \wtl{8}{1}{1} & \wtl{7}{2}{1}& \wtl{8}{4}{1} & \wtl{8}{3}{2} & \wtl{8}{3}{2} & \wtl{9}{0}{1} & \wtl{9}{0}{1} & \wtl{7}{1}{2} \\ \Xhline{4\arrayrulewidth}
\end{tabular}
}
\caption{\textsc{train-synth} (left table), \textsc{synth-discrim} (central table) $\&$ \textsc{synth-aug} (right table): statistical wins / ties / statistical losses for us (\taggt{})~vs CT-GAN (\tagctgan{}). Statistical = significant for $p \leq 0.01$. For example, \wtl{$a$}{$b$}{$c$} means we statistically win $a$ times, lose $c$ times and there is no statistical difference $b$ times. On \textsc{train-synth}, each red star ({\color{red}$\star$}) indicates a domain for which the related technique performed statistically \textit{worse} than uniform sampling (\textsc{unif}) for $p\leq 0.05$.}
\label{tab:train-synth-discrim-main}
\vspace{-0.4cm}
  \end{table}

  \subsection{Fake-real discrimination ('\textsc{synth-discrim}')}

  \paragraph{Objective and experimental setting} While the objective fits in a simple question (\textit{can the generated data look like real data ?}), its treatment necessitated a complex pipeline, in particular to avoid rewarding generators whose output would be a mere copy their training sample. The complete pipeline is detailed in \supplement~(Section \ref{sec-synt-disc}); very briefly, it starts by shuffling a 3-partition of the training data in a $3! = 6$-fold CV and ends up with supervised \textsc{rf} / \textsc{gbdt} classifiers (same as in Section \ref{sec-train-synth-main}) \textit{for a 2-class supervised problem} of fakes vs real distinction. The smaller their accuracy, the \textit{better} is the generator. We use CT-GANs as contenders; all parameters (GTs, CT-GANs) are the same as in Section \ref{sec-train-synth-main}.\\
  \paragraph{Results} Table \ref{tab:train-synth-discrim-main} (center) provides a summary of the results on 13 total domains considered. They display that GTs (regardless of their sizes) achieved a better job at fooling classifiers than neural nets. Much more interesting is perhaps the fact that GTs managed, on 3 (simulated) datasets, to better fool classifier than the \textit{original real data itself}. This never happened for CT-GANs. However, there is still a gap to fill for all techniques: CT-GANs do a statistically worse job at fooling classifiers than \textit{uniformly generated data} on 6 domains while GTs do statistically worse on 3 domains.

  \begin{table*}
  \centering
\begin{tabular}{rc}\hline\hline
  \rotatebox{90}{{\footnotesize \hspace{1cm} \texttt{dna}}} & \includegraphics[trim=0bp 0bp 0bp 300bp,clip,width=0.9\textwidth]{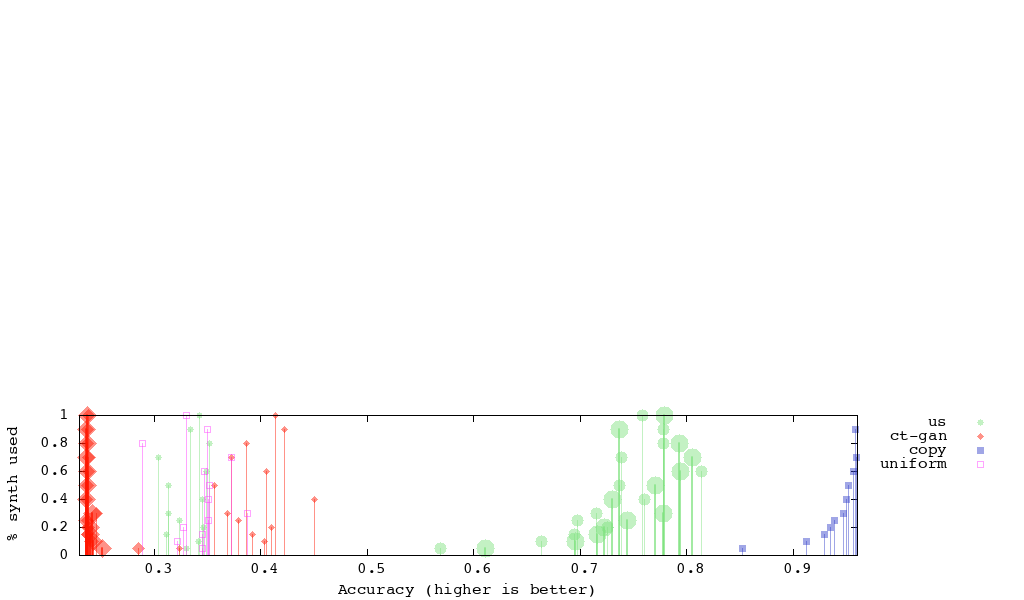} \\ \hline
 \rotatebox{90}{{\footnotesize \hspace{0.1cm} \texttt{house-votes}}} & \includegraphics[trim=0bp 0bp 0bp 300bp,clip,width=0.9\textwidth]{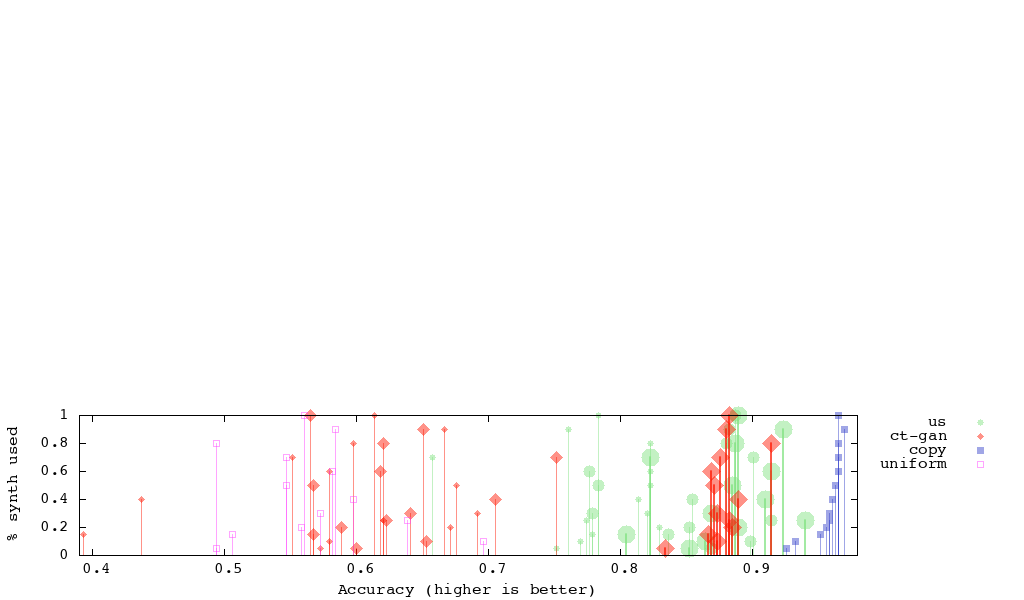} \\ \hline
 \rotatebox{90}{{\footnotesize \hspace{1cm} \texttt{led24}}}& \includegraphics[trim=0bp 0bp 0bp 300bp,clip,width=0.9\textwidth]{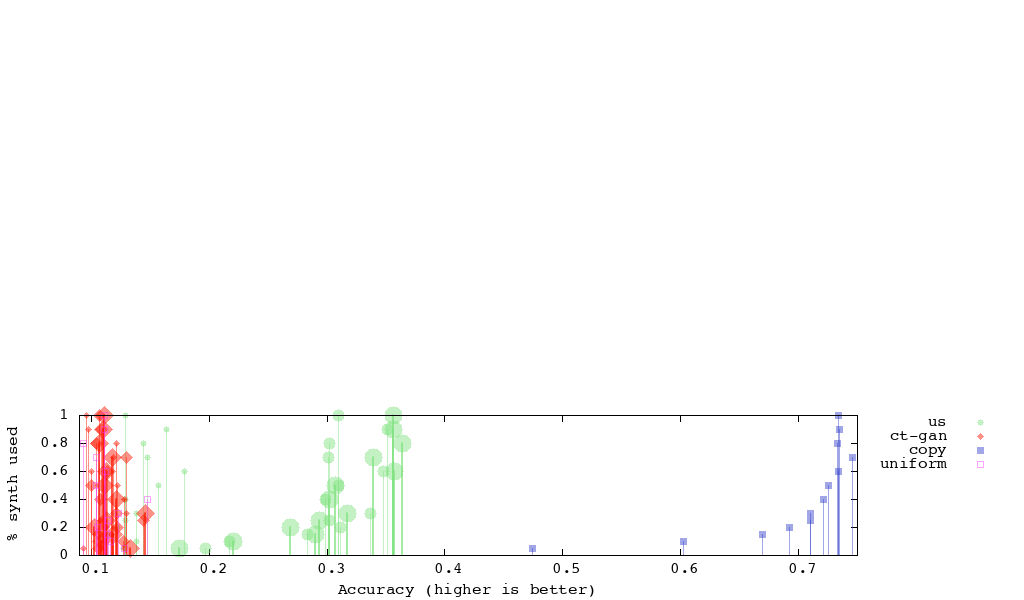} \\\hline\hline
\end{tabular}
\caption{Experiment \textsc{synth-aug}: detailed results on three domains for which the metric is the accuracy (more in \supplement, Section \ref{sec-synt-aug}). In each plot, the $x$ value of a vertical bar indicates a method's accuracy and the height along the $y$ axis indicates the $\%$ of real data that represents generated data used to train the final classifier (up to $100\%$). Green filled circles are GTs results, the size of the circle indicating the number of splits in the GTs (\tikzcircle[green, fill=green]{1pt} = 10, \tikzcircle[green, fill=green]{2pt} = 300, \tikzcircle[green, fill=green]{3pt} = 10K). Red filled diamonds are CT-GANs results, the size of the diamond indicating the number of epochs (\tikzdiamond[red, fill=red]{2pt} = 10, \tikzdiamond[red, fill=red]{4pt} = 300, \tikzdiamond[red, fill=red]{6pt} = 1K). Finally, empty pink squares (\tikzunif[pink]{3pt}) are \textsc{unif}'s results and filled blue squares (\tikzcopy[blue,fill=blue]{3pt}) are those of \textsc{copy}.}
    \label{tab:synth-aug-2-ACC-main}
  \end{table*}
  
  \subsection{'Synthetic augmentation' experiment (\textsc{synth-aug})}

  \paragraph{Objective and experimental setting} Supplementing real data with generative data is a particular case of data augmentation. Among the questions asked are obviously whether the additional generated data can bring better classifiers, but also how much generated data is worth adding and whether increasing the amount of generated data allows to increase the performances of models (regardless of a cross-technique comparison). The setting can be summarized as a variation on the \textsc{train-synth}) experiment, in which we train with \textsc{copy} + \textit{a} generated sample, with two main factors under control: (i) the technique used to supplement the additional data and (ii) the proportion of additional data with respect to the training set size. In the case of (i), we do not just include data generated by our technique or GANs but also consider adding purely \textsc{unif}ormly generated data and also adding real data from \textsc{copy} itself. Ideally, a good generative model should have performances at least in between these two, and the closer possible to the \textsc{copy} metrics.\\
  \paragraph{Results}  Table \ref{tab:train-synth-discrim-main} (right) provides a summary of the results on the 10 total domains considered. From a high level standpoint, it appears that generative trees tend to be a better fit than the neural nets of CT-GAN, even when comparing small trees to nets trained for a larger number of epochs (resulting in this case in a balanced picture among domains). To drill in the impact of the proportion of generated examples added, we have computed per-domain plots providing the full picture of how each technique compares to others. Table \ref{tab:synth-aug-2-ACC-main} displays the results of three domains chosen for their large number of variables (\texttt{dna}), missing values (\texttt{house-votes}) or highly noisy domain (\texttt{led24}). The remaining plots are available in \supplement, Section \ref{sec-synt-aug}. Several observations can be made. First, on \texttt{dna}, CT-GANs clearly overfit the domain as when the number of training epoch exceeds 10, the results are substantially worse than \textsc{unif}; on the contrary, GTs results are much improved when the number of splits in the tree exceeds 10, with also a further tendency to improvement as the quantity of generated data increases. On \texttt{house-votes}, CT-GANs trained with the largest number of epochs get good results, though GTs are the only one managing to beat \textsc{copy} with $5\%$ or $10\%$ additional real data. The variance of accuracies for CT-GANs is much higher than for GTs as almost all runs of CT-GANs with 10 or 300 training epochs lie in the span of \textsc{unif}'s results. On \texttt{led24}, all CT-GANs results are within the span of \textsc{unif}'s results. Only for the smallest trees do GTs achieve such suboptimal performances. Bigger trees result in substantially increases accuracies (by up to $\sim 20\%$), also displaying the same phenomenon as in \texttt{dna} that the more generated data is added, the better the results tend to be. Only on one of the ten domains (\texttt{sigma-cabs}) do we have a reversed picture with CT-GANs clearly beating GTs, yet no domain displays a pattern of GTs being substantially beaten by \textsc{unif} as we observe for \texttt{dna} on CT-GANs. These bad results of CT-GANs do not seems to come from the fact that the domain is Boolean-valued as we observe almost the same extreme results on \texttt{winewhite}, whose attributes are all continuous.

\section{Discussion and Conclusion}\label{sec-con}

Our contributions have different application spectra: while our models obviously fit only to tree-based generators, our contribution on losses has wider a wider applicability to \textit{any} calibrated classifiers. While copycat training is specific to a tree vs tree training procedure, our adversarial algorithm could be used to train generative trees against any calibrated classifier. GTs have advantages that neural nets do not necessarily have: they provide us with an \textit{exact and cheap to compute} expression of the measure learned, they can easily be used for \textit{missing data imputation}, and they also collect many benefits of DTs: interpretability (of the measure); they can be trained using various feature types (numeric, nominal, ordinal, etc.); and they can straightforwardly be trained from data with missing values. They also share some downsides of DTs, such as the fact that the underlying tree graph induces an 'axis-parallel' partition of the support. We anticipate that tricks used to alleviate DTs downsides can also be used for GTs, though maybe in a non-trivial way, like \textit{e.g.} for \citet{hksLO}. Important open problems include extending our formal results in generalisation and scaling the benefits of generative trees to \textit{ensembles} of generative trees. The fact that our generators in the copycat training scheme actually \textit{implement} boosting's modified hard distribution (defining the weak hypothesis assumption) in \citet{kmOT} might signal potential derivations of new training algorithms for generative models based on the use of boosting algorithms on the discriminator's side.\\
We hope our work brings new tools for models, losses and algorithms to train powerful generative models tailored to tabular data, and hope it contributes to fill the persistent gap in data generation quality for tabular data noted in recent work. 

\section*{Acknowledgments}

The authors would like to thank Ehsan Amid, Sercan Ar{\i}k, Olivier Bousquet, Julie Josse, Yishay Mansour, Aditya Krishna Menon, Madeleine Udell, Jean-Philippe Vert, Manfred Warmuth and Bob Williamson for many comments and stimulating discussions.

\bibliography{bibgen}
\bibliographystyle{icml2022}

\newpage
\appendix
\onecolumn
\renewcommand\thesection{\Roman{section}}
\renewcommand\thesubsection{\thesection.\arabic{subsection}}
\renewcommand\thesubsubsection{\thesection.\thesubsection.\arabic{subsubsection}}

\renewcommand*{\thetheorem}{\Alph{theorem}}
\renewcommand*{\thelemma}{\Alph{lemma}}
\renewcommand*{\thecorollary}{\Alph{corollary}}

\renewcommand{\thetable}{A\arabic{table}}

\begin{center}
\Huge{Appendix}
\end{center}

To
differentiate with the numberings in the main file, the numbering of
Theorems, etc. is letter-based (A, B, ...).

\section*{Table of contents}

\noindent \textbf{Supplementary material on proofs} \hrulefill Pg
\pageref{app-proofs}\\

\noindent $\hookrightarrow$ Proof of Theorem \ref{th-all-main}\hrulefill Pg
\pageref{proof-th-all-main}\\
\noindent $\hookrightarrow$ Proof of Lemma \ref{lempropDRLOSSsmall}\hrulefill Pg
\pageref{proof-lempropDRLOSSsmall}\\
\noindent $\hookrightarrow$ Proof of Lemma \ref{lembinf-lossg-ub}\hrulefill Pg
\pageref{proof-lembinf-lossg-ub}\\
\noindent $\hookrightarrow$ Proof of Theorems \ref{thboost1} and \ref{thboost2}\hrulefill Pg
\pageref{proof-thboost1-and-thboost2}\\
\noindent $\hookrightarrow$ Proof of Lemma \ref{it-to-dr}\hrulefill Pg
\pageref{proof-it-to-dr}\\

\noindent \textbf{Supplementary material on experiments} \hrulefill Pg
\pageref{app-exps}\\

\noindent $\hookrightarrow$ Examples of generative trees\hrulefill Pg \pageref{sec-example}\\
\noindent $\hookrightarrow$ Domains\hrulefill Pg \pageref{sec-doms}\\
\noindent $\hookrightarrow$ Data generation experiments\hrulefill Pg \pageref{sec-datagen}\\
\noindent $\hookrightarrow$ Missing data imputation experiments (\textsc{impute})\hrulefill Pg \pageref{sec-impute}\\
\noindent $\hookrightarrow$ 'Training on synthetic' experiment (\textsc{train-synth})\hrulefill Pg \pageref{sec-train-synt}\\
\noindent $\hookrightarrow$ 'Synthetic discrimination' experiment (\textsc{synth-discrim})\hrulefill Pg \pageref{sec-synt-disc}\\
\noindent $\hookrightarrow$ 'Synthetic augmentation' experiment (\textsc{synth-aug})\hrulefill Pg \pageref{sec-synt-aug}

\newpage

\section{Appendix on proofs}\label{app-proofs}

\subsection{Proof of Theorem \ref{th-all-main}}\label{proof-th-all-main}

We proceed in several steps.

\parnew{Proof of the identity between the external elements in \eqref{reducgen2}} We write, as in \citet[Appendix A.3]{rwID}, the second equality of
\begin{eqnarray}
  \idiv{}(\binartask_{\calposterior}) & \defeq & \idiv{f^\prior}(\meas{P}_{\calposterior}, \meas{N}_{\calposterior}) \nonumber\\
                      & = & \poibayesrisk(\prior) - \int \poibayesrisk\left(\prior\cdot \frac{\dmeas{P}_{\calposterior}}{\dmeas{M}_{\calposterior}}\right) \dmeas{M}_{\calposterior}\nonumber\\
                                                                      & = & \poibayesrisk(\prior) - \int \poibayesrisk\left(\calposterior\right) \dmeas{M}_{\calposterior}\label{penul}\\
                                                                      & = & \poibayesrisk(\prior) - \expect_{\X \sim \meas{M}_{\calposterior}} \left[\poibayesrisk(\calposterior)\right]\nonumber\\
  & = & \poibayesrisk(\prior) - \expect_{\X \sim \meas{M}_{\calposterior}} \left[\poibayesrisk({\calposterior}(\X),\calposterior(\X))\right]\label{penul2}\\
  & = & \poibayesrisk(\prior) - \popbayesrisk (\calposterior, \calposterior, \meas{M}_\posterior)\\
  & = & \statinf(\calposterior, \meas{M}_{\calposterior}) ,
\end{eqnarray}
where \eqref{penul} holds because $\calposterior$ is calibrated and \eqref{penul2} holds because the loss is proper and $\calposterior$ matches Bayes posterior on $\binartask_{\calposterior}$.

\parnew{Proof of the rightmost identity in \eqref{reducgen2}} We first prove several helper results. We first remark that $\loss$ being strictly proper differentiable implies $\fprior$ strictly convex and differentiable. We show these results for completeness, starting by showing $\poibayesrisk$ strictly concave: otherwise, we write $\poibayesrisk(\posterior_{1:2}) = (1/2) \cdot (\poibayesrisk(\posterior_1) + \poibayesrisk(\posterior_2))$ for $\posterior_{1:2} \defeq (1/2) \cdot (\posterior_1 + \posterior_2)$. If we had both
\begin{eqnarray}
  \poirisk(\posterior_1, \posterior_{1:2}) & > & \poirisk(\posterior_{1:2},\posterior_{1:2}) \defeq \poibayesrisk(\posterior_{1:2}) = (1/2) \cdot (\poibayesrisk(\posterior_1) + \poibayesrisk(\posterior_2)),\\
  \poirisk(\posterior_2, \posterior_{1:2}) & > & \poirisk(\posterior_{1:2},\posterior_{1:2}) \defeq \poibayesrisk(\posterior_{1:2}) = (1/2) \cdot (\poibayesrisk(\posterior_1) + \poibayesrisk(\posterior_2)),
  \end{eqnarray}
  then making the average of both yields $(1/2) \cdot (\poibayesrisk(\posterior_1) + \poibayesrisk(\posterior_2)) > (1/2) \cdot (\poibayesrisk(\posterior_1) + \poibayesrisk(\posterior_2))$, a contradiction, and yielding for example $\poirisk(\posterior_1, \posterior_{1:2}) \leq \poirisk(\posterior_{1:2},\posterior_{1:2})$, contradicting the strict properness of the loss in $\posterior_{1:2}$. Strict concavity of $\poibayesrisk$ implies strict convexity of $\fprior$ from its definition in \eqref{deffprior}. Also, the differentiability of the partial losses imply the differentiability of $\fprior$.

  For any strictly convex differentiable function $f$, we have $f^\star(z) = z {f'}^{-1}(z) - f({f'}^{-1}(z))$ and $(f^\star)' = {f'}^{-1}$, and if it is lower semicontinuous then $f^{\star\star} = f$. We check that $\fprior$ is indeed lower semicontinuous. Because $\poibayesrisk$ is continuous \citep[Lemma 3.1]{nssBP}, we study the set
  \begin{eqnarray}
\mathcal{I}(\alpha) & \defeq & \left\{t : (\prior t + 1 - \prior)\cdot\poibayesrisk\left(\frac{\prior t}{\prior t + 1 - \prior}\right) \geq \poibayesrisk(\prior) - \alpha \right\},
  \end{eqnarray}
  for $\alpha \in \mathbb{R}$. Denote for short $g(t) \defeq (\prior t + 1 - \prior)\cdot\poibayesrisk\left(\frac{\prior t}{\prior t + 1 - \prior}\right) $, which is continuous and also concave \citet[Appendix A.3]{rwID}. Recalling $B_F(z\|z') \defeq F(z) - F(z') - (z-z')F'(z')$ the Bregman divergence with generator $F$. We have
  \begin{eqnarray}
    g'(t) & = & \prior \cdot \poibayesrisk\left(\frac{\prior t}{\prior t + 1 - \prior}\right) + \frac{\prior(1-\prior)}{\prior t  + 1 - \prior} \cdot \poibayesrisk' \left(\frac{\prior t}{\prior t + 1 - \prior}\right)\nonumber\\
          & = &\prior \cdot \left( (-\poibayesrisk)(1) - (-\poibayesrisk)\left(\frac{\prior t}{\prior t + 1 - \prior}\right) - \left(1-\frac{\prior t}{\prior t + 1 - \prior}\right)\cdot (-\poibayesrisk') \left(\frac{\prior t}{\prior t + 1 - \prior}\right)\right) \label{toshow1}\\
          & = & \prior \cdot B_{-\poibayesrisk}\left( 1 \left\| \frac{\prior t}{\prior t + 1 - \prior}\right.\right)\\
    & = & \prior \cdot \partialloss{1}\left(\frac{\prior t}{\prior t + 1 - \prior}\right), \label{toshow2}
  \end{eqnarray}
  which, since $\partialloss{1}(1) = 0$, shows $\lim_{+\infty} g' = 0^+$; $g$ being concave, $g'$ is decreasing (which also shows $\partialloss{1}$ decreasing and $\partialloss{-1}$ increasing). To conclude, $g$ is increasing; hence, (when it is not empty) $\mathcal{I}(\alpha) = [t_u, +\infty)$ for some finite $t_u$, and so $\overline{\mathcal{I}(\alpha)} = (-\infty, t_u)$ is open, showing the closedness of $\mathcal{I}(\alpha)$ and the closedness of $\{t : \fprior(t) \leq \alpha\}$, and we get \textit{e.g.} from \citet[Theorem 7.1, point (b)]{rCA} that $\fprior$ is lower semicontinuous and thus $(\fprior)^{\star\star} = \fprior$. We complete the proof of the identities: \eqref{toshow1} holds because $\partialloss{-1}(0) = 0$ implies $\poibayesrisk(0) \defeq 0 \cdot \partialloss{1}(0) + 1\cdot \partialloss{-1}(0) = 0$ and because $\ell$ is also symmetric, then $\poibayesrisk(1) \defeq 1 \cdot \partialloss{1}(1) + 0\cdot \partialloss{-1}(1) = \partialloss{1}(1) = \partialloss{-1}(0) = 0$. Hence, $\poibayesrisk(0) = \poibayesrisk(1) = 0$. We state \eqref{toshow2} as a standalone Lemma.
\begin{lemma}\label{simpleLem}
  Suppose $\loss$ is proper differentiable and satisfies $\partialloss{-1}(0) = \partialloss{1}(1) = 0$. Then
  \begin{eqnarray*}
    B_{-\poibayesrisk}\left(0||u\right) & = & \partialloss{-1}(u).\\
    B_{-\poibayesrisk}\left(1||u\right) & = & \partialloss{1}(u).
    \end{eqnarray*}
  \end{lemma}
\begin{myproof}
The following two relationships hold because $\loss$ is proper, $\forall u \in [0,1]$ (using \eqref{pbr}):
\begin{eqnarray}
  \poibayesrisk(u) & = & u \cdot \partialloss{1}(u) + (1-u) \cdot \partialloss{-1}(u),\\
  u \cdot \partialloss{1}'(u) + (1-u) \cdot \partialloss{-1}'(u) & = & 0.\label{idderiv}
\end{eqnarray}
The second identity expresses the fact that properness implies (from \eqref{eqpoirisk}),
\begin{eqnarray}
\left. \frac{\partial}{\partial v} \poirisk(v,u) \right|_{v=u} & = & 0.
  \end{eqnarray}
We derive $\forall u \in [0,1]$, using $\poibayesrisk(0) = 0$,
\begin{eqnarray*}
  B_{-\poibayesrisk}\left(0||u\right) & = & \poibayesrisk(u) - u \poibayesrisk' (u)\\
                                      & = & u \cdot \partialloss{1}(u) + (1-u) \cdot \partialloss{-1}(u) - u\cdot(\partialloss{1}(u) + u \cdot \partialloss{1}'(u) - \partialloss{-1}(u) + (1-u) \cdot \partialloss{-1}'(u))\\
                                      & = & u \cdot \partialloss{1}(u) + (1-u) \cdot \partialloss{-1}(u) - u\cdot(\partialloss{1}(u) - \partialloss{-1}(u))\\
  & = & \partialloss{-1}(u), 
\end{eqnarray*}
as anticipated. We also have 
$\forall u \in [0,1]$, using $\poibayesrisk(1) = 0$,
\begin{eqnarray*}
  B_{-\poibayesrisk}\left(1||u\right) & = & \poibayesrisk(u) + (1-u) \poibayesrisk' (u)\\
  & = & \poibayesrisk(u) -u \poibayesrisk' (u) + \poibayesrisk' (u)\\
                                     & = & u \cdot \partialloss{1}(u) + (1-u) \cdot \partialloss{-1}(u) - u\cdot(\partialloss{1}(u) - \partialloss{-1}(u)) + \partialloss{1}(u) - \partialloss{-1}(u)\\
                                      & = & \partialloss{-1}(u) + \partialloss{1}(u) - \partialloss{-1}(u)\\
  & = & \partialloss{1}(u), 
\end{eqnarray*}
which completes the proof of the Lemma.
\end{myproof}\\
We now finish the proof of \eqref{reducgen2}.  Since $\calposterior$ is calibrated, then the corresponding likelihood satisfies 
  \begin{eqnarray}
    \callikelihood & \defeq & \frac{\calposterior}{1-\calposterior} \cdot \frac{1-\prior}{\prior}\\
                  & = & \frac{\prior\cdot \frac{\dmeas{P}_{\calposterior}}{\dmeas{M}_{\calposterior}}}{ (1-\prior)\cdot \frac{\dmeas{N}_{\calposterior}}{\dmeas{M}_{\calposterior}}} \cdot \frac{1-\prior}{\prior}\nonumber\\
    & = & \frac{\dmeas{P}_{\calposterior}}{\dmeas{N}_{\calposterior}}.
\end{eqnarray}
  Putting all this together, we get:
  \begin{eqnarray}
    -\disrisk(\callikelihood | \binartask_{\calposterior}) & \defeq & \expect_{\X \sim \meas{P}_h}\left[{f^\prior}' \circ \callikelihood (\X)\right] -\expect_{\X \sim \meas{N}_{\calposterior}}[\genloss(\callikelihood(\X))]\nonumber\\
                                           & = & \expect_{\X \sim \meas{P}_{\calposterior}}\left[{f^\prior}' \circ \callikelihood (\X)\right] -\expect_{\X \sim \meas{N}_{\calposterior}}[ (f^\prior)^\star\circ {f^\prior}' \circ \callikelihood(\X)]\nonumber\\
                                           & = & \int \frac{\dmeas{P}_{\calposterior}}{\dmeas{N}_{\calposterior}} \cdot {f^\prior}'\left(\frac{\dmeas{P}_{\calposterior}}{\dmeas{N}_{\calposterior}}\right) \dmeas{N}_{\calposterior} - \int (f^\prior)^\star\circ {f^\prior}' \left(\frac{\dmeas{P}_{\calposterior}}{\dmeas{N}_{\calposterior}}\right) \dmeas{N}_{\calposterior}\label{avoidEq}\\
                                           & = & \int \left(\frac{\dmeas{P}_{\calposterior}}{\dmeas{N}_{\calposterior}} \cdot {f^\prior}'\left(\frac{\dmeas{P}_{\calposterior}}{\dmeas{N}_{\calposterior}}\right) -  (f^\prior)^\star\circ {f^\prior}' \left(\frac{\dmeas{P}_{\calposterior}}{\dmeas{N}_{\calposterior}}\right)\right) \dmeas{N}_{\calposterior}\nonumber\\
                                           & = & \int \left(\frac{\dmeas{P}_{\calposterior}}{\dmeas{N}_{\calposterior}} \cdot ({{{f^\prior}^\star}'})^{-1}\left(\frac{\dmeas{P}_{\calposterior}}{\dmeas{N}_{\calposterior}}\right) -  (f^\prior)^\star\circ ({{{f^\prior}^\star}'})^{-1} \left(\frac{\dmeas{P}_{\calposterior}}{\dmeas{N}_{\calposterior}}\right)\right) \dmeas{N}_{\calposterior}\nonumber\\
                                           & = & \int (f^\prior)^{\star\star}\left(\frac{\dmeas{P}_{\calposterior}}{\dmeas{N}_{\calposterior}}\right) \dmeas{N}_{\calposterior}\nonumber\\
                                           & = & \int f^\prior \left(\frac{\dmeas{P}_{\calposterior}}{\dmeas{N}_{\calposterior}}\right) \dmeas{N}_{\calposterior}\nonumber\\
                                                           & = & \idiv{f^\prior}(\meas{P}_{\calposterior}, \meas{N}_{\calposterior}) \nonumber\\
    & = & \idiv{}(\binartask_{\calposterior}), \nonumber
  \end{eqnarray}
  as claimed. The key to avoiding the inequality \eqref{vfGAN} is \eqref{avoidEq}, which holds only because when the $\sigma$-algebras of the measures involved are coarsened with the level set of a calibrated posterior, it becomes Bayes posterior in the measure spaces obtained. This ends up the proof of \eqref{reducgen2}.

\parnew{Proofs of \eqref{eqgenloss} and expression of $(f^\prior)^\star(z)$}  We now compute $(f^\prior)^\star(z)$. Let
\begin{eqnarray}
  t & \defeq & \frac{p(1-\prior)}{(1-p)\prior}.
\end{eqnarray}
We have:
\begin{eqnarray*}
  (f^\prior)^\star(z) & \defeq & \sup_{t\geq 0} \left\{tz - \poibayesrisk(\prior) + (\prior t + 1 - \prior)\cdot\poibayesrisk\left(\frac{\prior t}{\prior t + 1 - \prior}\right) \right\}\nonumber\\
  & = & - \poibayesrisk(\prior) + \sup_{t\geq 0} \left\{tz + (\prior t + 1 - \prior)\cdot\poibayesrisk\left(\frac{\prior t}{\prior t + 1 - \prior}\right) \right\}\nonumber\\
  & = & - \poibayesrisk(\prior) + \sup_{p\in [0,1]} \left\{\frac{p(1-\prior)}{(1-p)\prior} \cdot z + \frac{1-\prior}{1-p}\cdot\poibayesrisk\left(p\right) \right\}\nonumber\\
  & = & - \poibayesrisk(\prior) + \frac{1-\prior}{\prior}\cdot \sup_{p\in [0,1]} \left\{\frac{pz + \prior \cdot\poibayesrisk\left(p\right)}{1-p} \right\}\nonumber.
\end{eqnarray*}
We see that $ (f^\prior)^\star$ is unbounded if $z>0$. Otherwise, we have
\begin{eqnarray*}
  \frac{\partial}{\partial p} \left(\frac{pz + \prior \cdot\poibayesrisk\left(p\right)}{1-p}\right)& = & \frac{(z + \prior \cdot\poibayesrisk'\left(p\right))(1-p) + (pz + \prior \cdot\poibayesrisk\left(p\right))}{(1-p)^2}\nonumber\\
  & = & \frac{z + \prior\cdot (\poibayesrisk\left(p\right)+(1-p)\poibayesrisk'\left(p\right))}{(1-p)^2}\nonumber\\
  & = & \frac{z + \prior\cdot (-\poibayesrisk\left(1\right) - (-\poibayesrisk\left(p\right)) - (1-p)(-\poibayesrisk)'\left(p\right)}{(1-p)^2}\nonumber\\
  & = & \frac{z + \prior\cdot B_{-\poibayesrisk}\left(1\|p\right)}{(1-p)^2}\nonumber\\
  & = & \frac{z + \prior\cdot \partialloss{1}(p)}{(1-p)^2},
\end{eqnarray*}
where we have used the fact that $\poibayesrisk\left(1\right) = \partialloss{-1}(0) = 0$ and. Zeroing the derivative, we thus seek $p_\pi(z)$ such that 
\begin{eqnarray}
  \partialloss{1}(p_\pi(z)) & = & -\frac{z}{\prior}\label{bregconst},
\end{eqnarray}
and since $\loss$ is strictly proper, $\partialloss{1}$ is invertible and we get
\begin{eqnarray}
  p_\pi(z) & = & \partialloss{1}^{-1}\left(-\frac{z}{\prior}\right)\label{bregconst2},
\end{eqnarray}
and so
\begin{eqnarray*}
  (f^\prior)^\star(z) & = & - \poibayesrisk(\prior) + \frac{1-\prior}{\prior}\cdot \frac{z p_\pi(z) + \prior \poibayesrisk\left(p_\pi(z)\right)}{1-p_\pi(z)}\nonumber\\
  & = & - \poibayesrisk(\prior) + (1-\prior)\cdot \frac{-p_\pi(z) \cdot B_{-\poibayesrisk}\left(1\|p_\pi(z)\right) + \poibayesrisk\left(p_\pi(z)\right)}{1-p_\pi(z)}\nonumber\\
  & = & - \poibayesrisk(\prior) + (1-\prior)\cdot \frac{-p_\pi(z) \cdot \poibayesrisk\left(p_\pi(z)\right) - p_\pi(z) (1-p_\pi(z)) \cdot \poibayesrisk '\left(p_\pi(z)\right) + \poibayesrisk\left(p_\pi(z)\right)}{1-p_\pi(z)}\nonumber\\
  & = & - \poibayesrisk(\prior) + (1-\prior)\cdot \left(\poibayesrisk\left(p_\pi(z)\right) - p_\pi(z) \poibayesrisk '\left(p_\pi(z)\right)\right) \nonumber\\
  & = & - \poibayesrisk(\prior) + (1-\prior)\cdot \left( -\poibayesrisk\left(0\right) - (-\poibayesrisk)\left(p_\pi(z)\right) - (0 - p_\pi(z)) (-\poibayesrisk) '\left(p_\pi(z)\right)\right) \nonumber\\
  & = & - \poibayesrisk(\prior) + (1-\prior)\cdot B_{-\poibayesrisk}\left(0\|p_\pi(z)\right) \\
  & = & - \poibayesrisk(\prior) + (1-\prior)\cdot \partialloss{-1} (p_\pi(z))\\
  & = & - \poibayesrisk(\prior) + (1-\prior)\cdot \partialloss{-1} \circ \partialloss{1}^{-1}\left(-\frac{z}{\prior}\right),
\end{eqnarray*}
This shows the expression of $(f^\prior)^\star(z)$. 
To compute $(f^\prior)^\star\circ {f^\prior}' (z)$, letting $\eta \defeq \prior z/(\prior z + 1 - \prior)$, we observe
\begin{eqnarray}
  -\frac{{f^\prior}' (v)}{\prior} & = &  \poibayesrisk\left(\frac{\prior z}{\prior z + 1 - \prior}\right)+\frac{1-\prior}{\prior z + 1 - \prior}\cdot \poibayesrisk'\left(\frac{\prior z}{\prior z + 1 - \prior}\right) \nonumber\\
  & = & \poibayesrisk\left(\eta\right)+(1-\eta)\cdot \poibayesrisk'\left(\eta\right) \nonumber\\
  & = & (-\poibayesrisk)\left(1\right)-(-\poibayesrisk)\left(\eta\right)-(1-\eta)\cdot (-\poibayesrisk)'\left(\eta\right) \nonumber\\
  & = & B_{-\poibayesrisk}\left(1\|\eta\right) \nonumber\\
  & = & \partialloss{1}(\eta),\label{relFPR}
\end{eqnarray}
hence
\begin{eqnarray}
  (f^\prior)^\star\circ {f^\prior}' (z) & = & - \poibayesrisk(\prior) + (1-\prior)\cdot \partialloss{-1} \left(\eta\right) \\
  & = & - \poibayesrisk(\prior) + (1-\prior)\cdot \partialloss{-1} \left(\frac{\prior z}{\prior z + 1 - \prior}\right) \\
  & = & - \poibayesrisk(\prior) + (1-\prior)\cdot \partialloss{-1} \left(\frac{1}{1 + \frac{1 - \prior}{\prior z }}\right),
\end{eqnarray}
and we remark that when $z \defeq \likelihood$ is a likelihood ratio,
\begin{eqnarray}
\frac{1 - \prior}{\prior z } & = & \densrat,
\end{eqnarray}
a density ratio, as claimed.

\subsection{Proof of Lemma \ref{lempropDRLOSSsmall}}\label{proof-lempropDRLOSSsmall}

From the chain rule and change of variable $\posterior \defeq 1/(1+\densrat)$, we get
\begin{eqnarray}
  {\partialloss{-1}}'(\posterior) & = & \frac{\mathrm{d} \partialloss{-1}(\posterior)}{\mathrm{d} \posterior}\\
  & = & \frac{\mathrm{d} \densrat}{\mathrm{d} \posterior} \cdot \frac{\mathrm{d} \partialloss{-1}(1/(1+\densrat))}{\mathrm{d} \densrat}\\
& = & - \frac{1}{\posterior^2} \cdot {\drloss}'(\densrat)\\
                                      & = &  - (1+\densrat)^2 \cdot {\drloss}'(\densrat). \label{eqdiffDRLOSS}
\end{eqnarray}
\eqref{toshow2} shows that $\partialloss{1}$ is decreasing and by symmetry, $\partialloss{-1}$ is increasing. So, ${\partialloss{-1}}'(\posterior) \geq 0$ and \eqref{eqdiffDRLOSS} shows then ${\drloss}'(\densrat) \leq 0$, thereby proving $\drloss$ is decreasing. We get from \eqref{idderiv} after working all parameters using \ref{eqdiffDRLOSS},
    \begin{eqnarray}
      \frac{1}{1+\densrat} \cdot - \left(1+\frac{1}{\densrat}\right)^2 \cdot {\drloss}'\left(\frac{1}{\densrat}\right) & = & \frac{\densrat}{1+\densrat} \cdot - (1+\densrat)^2 \cdot {\drloss}'(\densrat),
    \end{eqnarray}
    which becomes after simplification
    \begin{eqnarray}
      {\drloss}'\left(\frac{1}{\densrat}\right) & = & \densrat^3 \cdot {\drloss}'(\densrat), \forall \densrat\geq 0.
    \end{eqnarray}
Fix $\densrat, \epsilon > 0$, we also have 
\begin{eqnarray}
      {\drloss}'\left(\frac{1}{\densrat + \epsilon}\right) & = & (\densrat + \epsilon)^3 \cdot {\drloss}'(\densrat + \epsilon),
\end{eqnarray}
From which we get
\begin{eqnarray}
 \frac{{\drloss}'(\densrat + \epsilon) - {\drloss}'(\densrat)}{\epsilon} & = &  - \frac{1}{\epsilon} \cdot \left(\frac{1}{\densrat^3} \cdot {\drloss}'\left(\frac{1}{\densrat}\right) - \frac{1}{(\densrat + \epsilon)^3}\cdot  {\drloss}'\left(\frac{1}{\densrat + \epsilon}\right)\right)\label{stEPS}.
\end{eqnarray}
Suppose the LHS is $\leq 0$, which indicates a secant with negative slope at the right of $\densrat$. From the RHS, we then get the first inequality of
\begin{eqnarray}
  {\drloss}'\left(\frac{1}{\densrat}\right) & \geq & \frac{\densrat^3}{(\densrat + \epsilon)^3}\cdot  {\drloss}'\left(\frac{1}{\densrat + \epsilon}\right)\nonumber\\
  & \geq & {\drloss}'\left(\frac{1}{\densrat + \epsilon}\right)
\end{eqnarray}
and the second is due to the fact that ${\drloss}'(.) \leq 0$. Letting $\densrat' \defeq 1 / (\densrat+\epsilon)$ and $\epsilon' \defeq \epsilon/(\densrat (\densrat + \epsilon))$, we then get
\begin{eqnarray}
\frac{{\drloss}'\left(\densrat' + \epsilon'\right) - {\drloss}'\left(\densrat'\right)}{\epsilon'} & \geq & 0,
\end{eqnarray}
which indicates a secant with positive slope at the right of $\densrat'$ or equivalently at the left of $1/\densrat$. Taking the limits for $\epsilon \rightarrow 0$, we see that if the right derivative at $\densrat$ is negative, then the left derivative at $1/\densrat$ is positive. Switching $\epsilon < 0$ from \eqref{stEPS} switches the directional derivatives and shows that if ${\drloss}'$ is not convex in $\densrat$, then it is convex in $1/\densrat$.

    \subsection{Proof of Lemma \ref{lembinf-lossg-ub}}\label{proof-lembinf-lossg-ub}

    The proof follows from Jensen's inequality: if $\drloss$ is convex then we have by definition of $\callikelihood$:
    \begin{eqnarray}
      \genrisk(\meas{N}_{\calposterior}| \callikelihood) & \defeq & \poibayesrisk(\prior) - (1-\prior)\cdot \expect_{\meas{N}_{\calposterior}}\left[\partialloss{-1}\left(\frac{1}{1+\frac{1-\prior}{\prior}\cdot \frac{\dmeas{N}_{\calposterior}}{\dmeas{P}_{\calposterior}}}\right)\right] \label{defORcal} \\
      & =  &  \poibayesrisk(\prior) - (1-\prior)\cdot \expect_{\meas{N}_{\calposterior}}\left[\drloss\left(\frac{1-\prior}{\prior}\cdot \frac{\dmeas{N}_{\calposterior}}{\dmeas{P}_{\calposterior}}\right)\right] \nonumber \\
      & \leq  &  \poibayesrisk(\prior) - (1-\prior)\cdot \drloss\left(\frac{1-\prior}{\prior}\cdot \expect_{\meas{N}_{\calposterior}}\left[\frac{\dmeas{N}_{\calposterior}}{\dmeas{P}_{\calposterior}}\right]\right) \label{eqCHI2-1} ,
    \end{eqnarray}
    and we check that
    \begin{eqnarray}
      \expect_{\meas{N}_{\calposterior}}\left[\frac{\dmeas{N}_{\calposterior}}{\dmeas{P}_{\calposterior}}\right] & = & \expect_{\meas{P}}\left[\left(\frac{\dmeas{N}_{\calposterior}}{\dmeas{P}_{\calposterior}}\right)^2\right] \nonumber\\
      & = & \expect_{\meas{P}}\left[\left(\frac{\dmeas{N}_{\calposterior}}{\dmeas{P}_{\calposterior}}\right)^2\right]-2\cdot \expect_{\meas{P}}\left[\frac{\dmeas{N}_{\calposterior}\dmeas{P}_{\calposterior}}{\dmeas{P}_{\calposterior}^2}\right] + \expect_{\meas{P}}\left[\left(\frac{\dmeas{P}_{\calposterior}}{\dmeas{P}_{\calposterior}}\right)^2\right] + 1 \nonumber\\
                                                                                                              & = & \expect_{\meas{P}}\left[\left(\frac{\dmeas{N}_{\calposterior}-\dmeas{P}_{\calposterior}}{\dmeas{P}_{\calposterior}}\right)^2\right] + 1 \nonumber\\
      & = & \chi^2\left(\meas{N}_{\calposterior}||\meas{P}_{\calposterior}\right)+1,
    \end{eqnarray}
    from which we get
    \begin{eqnarray}
      \drloss\left(\frac{1-\prior}{\prior}\cdot \expect_{\meas{N}_{\calposterior}}\left[\frac{\dmeas{N}_{\calposterior}}{\dmeas{P}_{\calposterior}}\right]\right) & = & \drloss\left(\frac{1-\prior}{\prior}\cdot \left(\chi^2\left(\meas{N}_{\calposterior}||\meas{P}_{\calposterior}\right)+1\right)\right) \nonumber\\
      & = & \partialloss{-1}\left(\frac{1}{1+\frac{1-\prior}{\prior}\cdot \left(\chi^2\left(\meas{N}_{\calposterior}||\meas{P}_{\calposterior}\right)+1\right)}\right)  \nonumber\\
      & = & \partialloss{-1}\left(\frac{\pi}{1+(1-\prior)\cdot \chi^2\left(\meas{N}_{\calposterior}||\meas{P}_{\calposterior}\right)}\right),
    \end{eqnarray}
    and obtain the lowerbound of \eqref{boundGenrisk} after combining with \eqref{eqCHI2-1}.

    \subsection{Proof of Theorems \ref{thboost1} and \ref{thboost2}}\label{proof-thboost1-and-thboost2}

\begin{figure}
  \centering
    \includegraphics[trim=60bp 40bp 40bp 90bp,clip, width=0.8\textwidth]{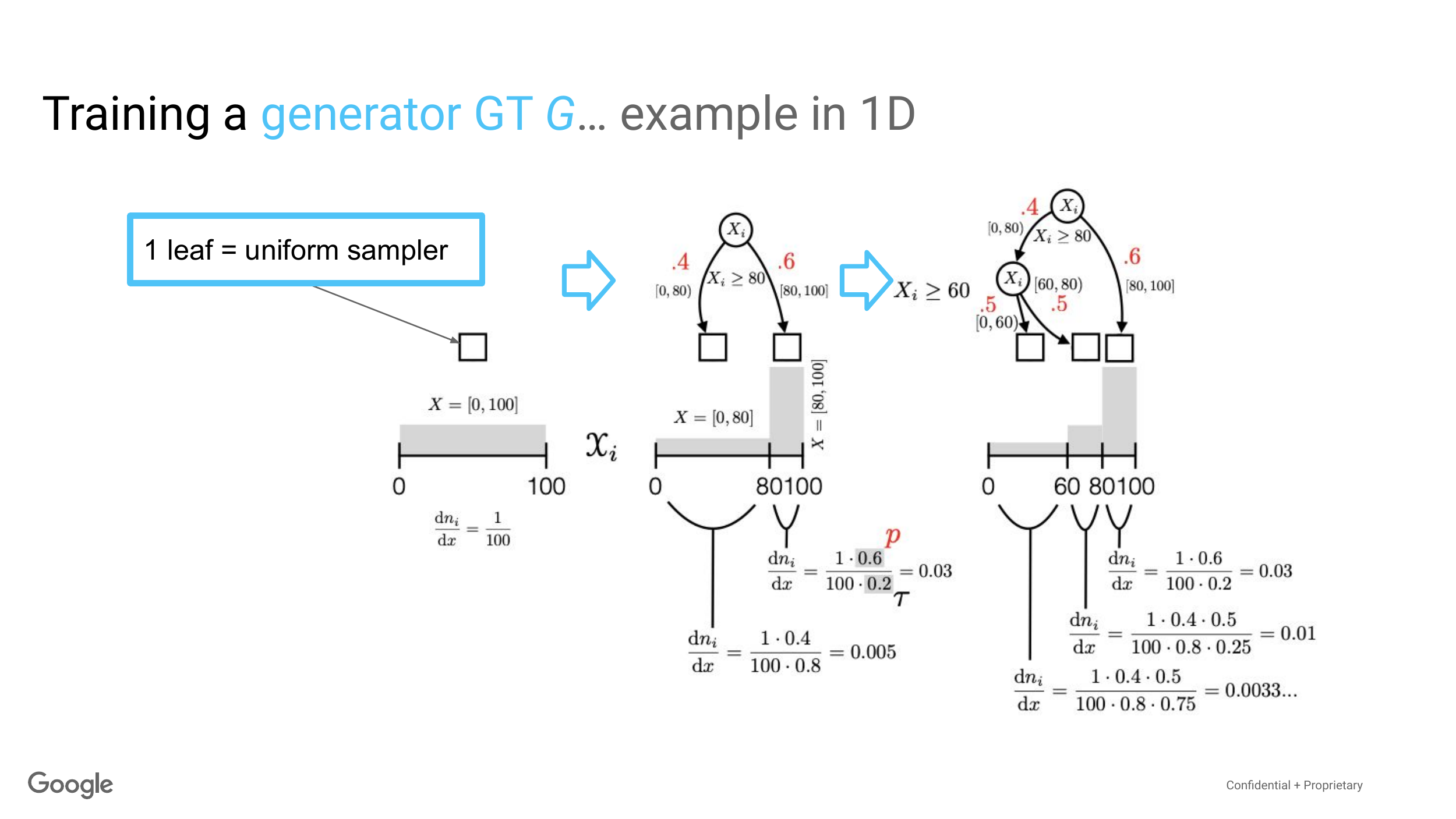}
    \caption{Explanation of \eqref{newsplit-appendix} on a 1D example: two splits on the same variable in a generative tree creating a piecewise constant but non uniform density for the variable.}
    \label{fig:scale-dens}
  \end{figure}

    We start by a simple technical Lemma.
    \begin{lemma}\label{defF}
      Let $f(u) \defeq au^2 +bu+c$; suppose $f(u) \geq 0, \forall u\in \mathbb{R}$, implying $a\geq 0, 4ac - b^2\geq 0$. Consider $u^* \defeq \arg\min_{\mathbb{R}} f(u) = -b/(2a)$. Let $Q \defeq (4ac - b^2)/4a$. For any $\epsilon > 0$ and any $v\in \mathbb{R}$,
      \begin{eqnarray}
      |v-u^*| \geq \epsilon & \Rightarrow & f(u^*) \leq \frac{1}{1+\epsilon^2 Q} \cdot f(v).
      \end{eqnarray}
      \end{lemma}
    \begin{myproof}
    Fix $\delta \in [0,1)$. We want to compute the $v$s such that $f(u^*) \leq (1-\delta) f(v)$. Equivalently, we want
    \begin{eqnarray}
av^2 +bv + \frac{b^2-4\delta ac}{4(1-\delta)a} & \geq & 0.
    \end{eqnarray}
    We have the discriminant
    \begin{eqnarray}
\Delta & = & b^2 - \frac{b^2-4\delta ac}{1-\delta} = \frac{\delta (4ac - b^2)}{1-\delta},
    \end{eqnarray}
    and we have $\Delta \geq 0$ because of the constraints on $f$. The $v$s we seek therefore satisfy
    \begin{eqnarray}
|v-u^*| & \geq & \sqrt{\frac{\delta}{1-\delta} \cdot \left(\frac{4ac - b^2}{4a}\right)}
      \end{eqnarray}
     Solving the RHS$=\epsilon$ for $\delta$ thus yields that if $|v-u^*| \geq \epsilon$ then
      \begin{eqnarray}
      f(u^*) & \leq & \frac{1}{1+\epsilon^2 Q} \cdot f(v),
      \end{eqnarray}
      where $Q$ is defined in the statement of the Lemma.
    \end{myproof}\\
We have the expressions related to split of $\samplingnode$ using feature $X$:
    \begin{eqnarray}
{n'}^0_\leaf = {n}^0_\leaf ;  \quad  {n'}^l_\leaf =  {n}^l_\leaf \cdot \frac{1-p}{1-\tau}  \quad  {n'}^r_\leaf =  {n}^r_\leaf \cdot \frac{p}{\tau},\label{newsplit-appendix}
    \end{eqnarray}
    and $n'_\leaf  = {n'}^0_\leaf + {n'}^l_\leaf + {n'}^r_\leaf$, $n_\leaf  = {n}^0_\leaf + {n}^l_\leaf + {n}^r_\leaf$. Figure \ref{fig:scale-dens} gives an example of the way the scaling factors are obtained on a simple example. These come from scaling the densities after the split (which partitions further the support) and the computation of the associated Bernoullis $\B(p)$ (which defines the stochastic activation of the generative tree). Since this process does not change the way the support is split by the discriminator, the weights -- integrals of these densities -- are scaled by the same factors as depicted in \eqref{newsplit-appendix}.
    We recall notations
    \begin{eqnarray*}
l_\leaf \defeq {n}^0_\leaf + \frac{{n}^l_\leaf }{1-\tau} \quad;\quad r_\leaf \defeq {n}^0_\leaf + \frac{{n}^r_\leaf }{\tau}\quad;\quad \vardelta_\leaf \defeq \frac{{n}^r_\leaf}{\tau}  - \frac{{n}^l_\leaf}{1-\tau} = r_\leaf - l_\leaf .
    \end{eqnarray*}
For any $p \in [0,1]$, the new $\chi^2$ after split of $\samplingnode$ using feature $X$ admits the simplified expression using \eqref{newsplit-appendix} and $\mu_{\LL\LL} \defeq \sum_{\leaf \in \leafset(h)} l^2_\leaf / p_\leaf$, $\mu_{\D\D} \defeq \sum_{\leaf \in \leafset(h)} \delta^2_\leaf / p_\leaf$, $\mu_{\LL\D} \defeq \sum_{\leaf \in \leafset(h)} l_\leaf \delta_\leaf / p_\leaf$,
    \begin{eqnarray}
      \chi^2\left(\meas{N}'_{\calposterior}(p)||\meas{P}_{\calposterior}\right) & = & \sum_{\leaf \in \leafset(h)} \frac{\left(p_\leaf- {n}^0_\leaf  - {n}^l_\leaf \cdot \frac{1-p}{1-\tau}  -{n}^r_\leaf \cdot \frac{p}{\tau} \right)^2}{p_\leaf}\nonumber\\
      & = & -1 + \sum_{\leaf \in \leafset(h)} \frac{\left({n}^0_\leaf +{n}^l_\leaf \cdot \frac{1-p}{1-\tau}  +  {n}^r_\leaf \cdot \frac{p}{\tau}\right)^2}{p_\leaf}\nonumber\\
                                                                           & = & -1 +  \sum_{\leaf \in \leafset(h)} \frac{\left(l_\leaf + p \vardelta_\leaf\right)^2}{p_\leaf}\label{defCHI2-fork}\\
                                                                           & = & -1 + \mu_{\LL\LL} + 2p \mu_{\LL\D} + p^2 \mu_{\D\D},\label{defCHI2-L}
    \end{eqnarray}
    where we put $p$ in parameter of $\meas{N}'_{\calposterior}(.)$. Note we can also take a fork at \eqref{defCHI2-fork} and write instead:
    \begin{eqnarray}
      \chi^2\left(\meas{N}'_{\calposterior}(p)||\meas{P}_{\calposterior}\right) & = & -1 +  \sum_{\leaf \in \leafset(h)} \frac{\left(r_\leaf - (1-p) \vardelta_\leaf\right)^2}{p_\leaf}\label{defCHI2-fork}\\
                                                                           & = & -1 + \mu_{\R\R} - 2(1-p) \mu_{\R\D} + (1-p)^2 \mu_{\D\D}.\label{defCHI2-R}
    \end{eqnarray}
    Three values of $p$ are of interest:
    \begin{itemize}
      \item for $p=\tau$, we get
    \begin{eqnarray}
      \chi^2\left(\meas{N}'_{\calposterior}(\tau)||\meas{P}_{\calposterior}\right) = \chi^2\left(\meas{N}_{\calposterior}||\meas{P}_{\calposterior}\right),\label{chi2tau}
      \end{eqnarray}
      and there is no change in $\chi^2$ after split.
      \item for $p=0$, we get
\begin{eqnarray}
      \chi^2\left(\meas{N}'_{\calposterior}(0)||\meas{P}_{\calposterior}\right) = -1 + \mu_{\LL\LL},\label{chi2zero}
\end{eqnarray}
which corresponds to discarding support on $X^r$ and yields $n'_\leaf = l_\leaf$;
\item for $p=1$, we get
\begin{eqnarray}
      \chi^2\left(\meas{N}'_{\calposterior}(1)||\meas{P}_{\calposterior}\right) = -1 + \mu_{\R\R}, \label{chi2un}
\end{eqnarray}
which corresponds to discarding support on $X^l$ and yields $n'_\leaf = r_\leaf$.
\end{itemize}
\noindent \textbf{Case 1 -- } suppose $\mu_{\D\D} > 0$. Define $f$ as in Lemma \ref{defF} using \eqref{defCHI2-L}, with $a \defeq \mu_{\D\D}, b \defeq 2 \mu_{\LL\D}, c\defeq \mu_{\LL\LL} - 1$. We have
\begin{eqnarray}
u^* & = & -\frac{ \mu_{\LL\D}}{\mu_{\D\D}} = \frac{\mu_{\LL\LL}-\mu_{\LL\R}}{\mu_{\LL\LL}+\mu_{\R\R}-2 \mu_{\LL\R}}\label{defUSTAR}
  \end{eqnarray}
  \noindent \textbf{Case 1.1 -- } suppose in addition $\mu_{\LL\LL}\geq \mu_{\LL\R} $ and $\mu_{\R\R}\geq \mu_{\LL\R}$. We have $u^* \in [0,1]$ and fix $p \defeq u^*$ . For the choice $v = \tau$, Lemma \ref{defF} says that
  \begin{eqnarray}
      |\tau-p| \geq \epsilon & \Rightarrow & \chi^2\left(\meas{N}'_{\calposterior}(p)||\meas{P}_{\calposterior}\right) \leq \frac{1}{1+\epsilon^2 Q} \cdot  \chi^2\left(\meas{N}_{\calposterior}||\meas{P}_{\calposterior}\right),
  \end{eqnarray}
  with
  \begin{eqnarray}
Q & = & \frac{\mu_{\D\D}(\mu_{\LL\LL} - 1) - \mu_{\LL\D}^2}{\mu_{\D\D}} = -1 + \mu_{\LL\LL} - \frac{\mu_{\LL\D}^2}{\mu_{\D\D}}.
  \end{eqnarray}
  We also remark that with the value of $p$ as in \eqref{defUSTAR}, $\chi^2\left(\meas{N}'_{\calposterior}(p)||\meas{P}_{\calposterior}\right)$ turns out to be
  \begin{eqnarray}
\chi^2\left(\meas{N}'_{\calposterior}(p)||\meas{P}_{\calposterior}\right) & = & -1 + \mu_{\LL\LL} - 2 \cdot \frac{ \mu_{\LL\D}}{\mu_{\D\D}} \cdot \mu_{\LL\D} + \left(\frac{ \mu_{\LL\D}}{\mu_{\D\D}}\right)^2 \mu_{\D\D} = Q.
  \end{eqnarray}
  Hence, for any $\updelta > 0$, as long as $\chi^2\left(\meas{N}'_{\calposterior}(.)||\meas{P}_{\calposterior}\right) \geq \updelta$, whenever $|\tau-p| \geq \epsilon$, one step of \topdowngen~achieves geometric convergence with rate $1/(1+\updelta \epsilon^2)$.\\
  \noindent \textbf{Case 1.2 -- } suppose now $\mu_{\LL\LL} < \mu_{\LL\R} $, which implies $u^* < 0$. Pick $p=0$. From \eqref{chi2zero}, \eqref{chi2tau} and \eqref{defCHI2-L}, to get $\chi^2\left(\meas{N}'_{\calposterior}(0)||\meas{P}_{\calposterior}\right) \leq (1/(1+\epsilon))\cdot \chi^2\left(\meas{N}_{\calposterior} ||\meas{P}_{\calposterior}\right)$, we equivalently need
  \begin{eqnarray}
      \tau^2 \mu_{\D\D}+ 2\tau \mu_{\LL\D} - \epsilon \mu_{\LL\LL} +\epsilon& \geq & 0,\label{eqTARGET0}
    \end{eqnarray}
  which, expressed using the fact that $\mu_{\D\D} = \mu_{\R\R} + \mu_{\LL\LL} - 2 \mu_{\LL\R}$ and $\mu_{\LL\D} =  \mu_{\LL\R} - \mu_{\LL\LL}$, yields:
\begin{eqnarray}
      (\tau^2 -2\tau - \epsilon) \mu_{\LL\LL} + \tau^2 \mu_{\R\R} + 2\tau(1-\tau) \mu_{\LL\R}  +\epsilon & \geq & 0,\label{eqTARGET021}
    \end{eqnarray}
    The Weak Generative Assumption, $\mu_{\D\D} \geq \updelta \cdot \max\{\mu_{\LL\LL}, \mu_{\RR\RR}\}$ implies $\mu_{\R\R} \geq 2 \mu_{\LL\R} - \mu_{\LL\LL} + \updelta \cdot \mu_{\LL\LL}$, and with Case 1.2's assumption, $\mu_{\LL\R} > \mu_{\LL\LL}$, yields $\mu_{\R\R} \geq 2 \mu_{\LL\LL} - \mu_{\LL\LL} + \updelta \cdot \mu_{\LL\LL} = (1+\updelta)\cdot \mu_{\LL\LL}$, so we get
    \begin{eqnarray}
      (\tau^2 -2\tau - \epsilon) \mu_{\LL\LL} + \tau^2 \mu_{\R\R} + 2\tau(1-\tau) \mu_{\LL\R}  +\epsilon & > & (\tau^2 -2\tau - \epsilon) \mu_{\LL\LL} + (1+\updelta)  \tau^2 \cdot \mu_{\LL\LL} \nonumber\\
                                                                                                         & & + 2\tau(1-\tau) \mu_{\LL\LL}  +\epsilon\nonumber\\
      & & = (\updelta  \tau^2 - \epsilon) \cdot \mu_{\LL\LL} + \epsilon.
      \end{eqnarray}
      Fixing $\epsilon = \updelta  \tau^2$ thus brings \eqref{eqTARGET0} and we conclude with
      \begin{eqnarray}
\chi^2\left(\meas{N}'_{\calposterior}(0)||\meas{P}_{\calposterior}\right) \leq \frac{1}{1+\updelta  \tau^2}\cdot \chi^2\left(\meas{N}_{\calposterior} ||\meas{P}_{\calposterior}\right).\label{eqC12}
        \end{eqnarray}
    \noindent \textbf{Case 1.3 -- } suppose now $\mu_{\RR\RR} < \mu_{\LL\R} $, which implies $u^* > 1$ (the denominator of $u^*$ is always $\geq 0$). Pick $p=1$. From \eqref{chi2un}, \eqref{chi2tau} and \eqref{defCHI2-R}, to get $\chi^2\left(\meas{N}'_{\calposterior}(1)||\meas{P}_{\calposterior}\right) \leq (1/(1+\epsilon))\cdot \chi^2\left(\meas{N}_{\calposterior} ||\meas{P}_{\calposterior}\right)$, we equivalently need
  \begin{eqnarray}
      (1-\tau)^2 \mu_{\D\D}- 2(1-\tau) \mu_{\R\D} - \epsilon \mu_{\R\R} +\epsilon& \geq & 0.\label{eqTARGET0-R}
  \end{eqnarray}
  Using $\mu_{\D\D} = \mu_{\R\R} + \mu_{\LL\LL} - 2 \mu_{\LL\R}$ and $\mu_{\R\D} =  \mu_{\R\R} - \mu_{\LL\R}$, we break this down to:
    \begin{eqnarray}
      (1-\tau)^2 \mu_{\LL\LL} -(1- \tau^2 +\epsilon) \mu_{\R\R} + 2\tau(1-\tau) \mu_{\LL\R}  +\epsilon & \geq & 0,\label{eqTARGET021}
    \end{eqnarray}
    The Weak Generative Assumption yields this time $\mu_{\LL\LL} \geq 2 \mu_{\LL\R} - \mu_{\R\R} + \updelta \cdot \mu_{\RR\RR}$, which, together with Case 1.3's assumption, $\mu_{\LL\R} > \mu_{\R\R}$, yields $\mu_{\LL\LL} \geq (1+\updelta)\cdot \mu_{\R\R}$, so we get this time
    \begin{eqnarray}
      (1-\tau)^2 \mu_{\LL\LL} -(1- \tau^2 +\epsilon) \mu_{\R\R} + 2\tau(1-\tau) \mu_{\LL\R}  +\epsilon & > & (1-\tau)^2  (1+\updelta)\cdot \mu_{\R\R} -(1- \tau^2 +\epsilon) \mu_{\R\R} \nonumber\\
                                                                                                       & & + 2\tau(1-\tau) \mu_{\R\R}  +\epsilon\nonumber\\
      & & = (\updelta  (1-\tau)^2 - \epsilon) \cdot \mu_{\R\R} + \epsilon.
    \end{eqnarray}
    Fixing $\epsilon = \updelta  (1-\tau)^2$ thus brings \eqref{eqTARGET0} and we conclude with
      \begin{eqnarray}
\chi^2\left(\meas{N}'_{\calposterior}(0)||\meas{P}_{\calposterior}\right) \leq \frac{1}{1+\updelta  (1-\tau)^2}\cdot \chi^2\left(\meas{N}_{\calposterior} ||\meas{P}_{\calposterior}\right). \label{eqC13}
      \end{eqnarray}
      Cases 1.2 and 1.3 can be summarised as
      \begin{eqnarray}
\chi^2\left(\meas{N}'_{\calposterior}(p)||\meas{P}_{\calposterior}\right) \leq \frac{1}{1+\updelta  (\tau + (1-2\tau)p)^2}\cdot \chi^2\left(\meas{N}_{\calposterior} ||\meas{P}_{\calposterior}\right), p \in \{0,1\}. \label{eqC123}
      \end{eqnarray}
      \noindent \textbf{Case 2 -- } suppose $\mu_{\D\D} = 0$. We remark in this case that the optimal $p$ to minimize $\chi^2\left(\meas{N}'_{\calposterior}(p)||\meas{P}_{\calposterior}\right)$ as in \eqref{defCHI2-L} or \eqref{defCHI2-R} is in $\{0,1\}$, which brings us to the case where the Weak Generative Assumption holds and therefore make that Case 2 does not happen for the analysis of \topdowngen.

\subsection{Proof of Lemma \ref{it-to-dr}}\label{proof-it-to-dr}
We note that we have for any $\leaf \in \leafset(h_T)$ and $\ve{x}$ reaching $\leaf$,
\begin{eqnarray}
\frac{\prior \dmeas{P}^\top}{\prior \dmeas{P}^\top + (1-\prior) \dmeas{U}}(\ve{x}) & = & \frac{\prior p_\leaf}{\prior p_\leaf + (1-\prior) u_\leaf},
\end{eqnarray}
where we recall $p_\leaf \defeq \int_\leaf \dmeas{P}$ and $u_\leaf \defeq \int_\leaf \dmeas{U}$. We get from the scaled Bregman Theorem \citep[Theorem 1]{nmoAS} that since $g$ is affine, the perspective transform $\perspectivepobayesrisk$ is convex and the first equality holds in:
    \begin{eqnarray*}
      \lefteqn{\prior \cdot \expect_{\meas{U}} \left[ B_{\perspectivepobayesrisk}\left(\frac{\dmeas{P}}{\dmeas{U}}\left\|\frac{\dmeas{P}^T}{\dmeas{U}}\right. \right) \right]}\nonumber\\
      & = & \prior \cdot \int_{\mathcal{X}} g\left(\frac{\dmeas{P}}{\dmeas{U}}\right)\cdot B_{-\poibayesrisk} \left(\frac{\frac{\dmeas{P}}{\dmeas{U}}}{g(\frac{\dmeas{P}}{\dmeas{U}})} \left\| \frac{\frac{\dmeas{P}^T}{\dmeas{U}}}{g\left(\frac{\dmeas{P}^T}{\dmeas{U}}\right)} \right.\right) \cdot \dmeas{U}\\
      & = & \underbrace{\int_{\mathcal{X}} (\prior \dmeas{P} + (1-\prior) \dmeas{U}) \cdot (-\poibayesrisk) \left(\frac{\prior \dmeas{P}}{\prior \dmeas{P} + (1-\prior) \dmeas{U}}\right)}_{\defeq A}\\
      & & -  \underbrace{\int_{\mathcal{X}} (\prior \dmeas{P} + (1-\prior) \dmeas{U}) \cdot (-\poibayesrisk) \left(\frac{\prior \dmeas{P}^\top}{\prior \dmeas{P}^\top + (1-\prior) \dmeas{U}}\right)}_{\defeq B}\\
      & & - \underbrace{\int_{\mathcal{X}} (\prior \dmeas{P} + (1-\prior) \dmeas{U}) \cdot \left(\frac{\prior \dmeas{P}}{\prior \dmeas{P} + (1-\prior) \dmeas{U}}-\frac{\prior \dmeas{P}^\top}{\prior \dmeas{P}^\top + (1-\prior) \dmeas{U}}\right)\cdot (-\poibayesrisk)' \left(\frac{\prior \dmeas{P}^\top}{\prior \dmeas{P}^\top + (1-\prior) \dmeas{U}}\right)}_{\defeq C},
    \end{eqnarray*}
    and the second equality follows from the definition of Bregman divergences. Since leaves in $h_T$ induce a partition of $\mathcal{X}$, $C$ simplifies to:
    \begin{eqnarray*}
      C & = & \sum_{\leaf \in \leafset(h_T)} (-\poibayesrisk)' \left(\frac{\prior p_\leaf}{\prior p_\leaf + (1-\prior) u_\leaf}\right) \cdot \left(\prior \int_\leaf \dmeas{P} - \frac{\prior p_\leaf}{\prior p_\leaf + (1-\prior) u_\leaf} \cdot \int_\leaf \left(\prior \dmeas{P} + (1-\prior) \dmeas{U}\right)\right)\\
        & = &  \sum_{\leaf \in \leafset(h_T)} (-\poibayesrisk)' \left(\frac{\prior p_\leaf}{\prior p_\leaf + (1-\prior) u_\leaf}\right) \cdot \underbrace{\left(\prior \int_\leaf \dmeas{P} - \prior p_\leaf\right)}_{=0}\\
      & = & 0.
    \end{eqnarray*}
    We can also reformulate $A-B$:
    \begin{eqnarray*}
      A - B & = & \poibayesrisk(\prior) - \int_{\mathcal{X}} (\prior \dmeas{P} + (1-\prior) \dmeas{U}) \cdot \poibayesrisk \left(\frac{\prior \dmeas{P}}{\prior \dmeas{P} + (1-\prior) \dmeas{U}}\right)\\
            && - \left(\poibayesrisk(\prior) - \sum_{\leaf \in \leafset(h_T)} (\prior p_\leaf + (1-\prior) u_\leaf) \cdot \poibayesrisk \left(\frac{\prior p_\leaf}{\prior p_\leaf + (1-\prior) u_\leaf}\right)\right)\\
      & = & \idiv{\fprior}(\meas{P}, \meas{U}) - \idiv{\fprior}(\meas{P}_{\calposterior_T}, \meas{U}_{\calposterior_T}),
    \end{eqnarray*}
    which leads to the statement of the Lemma.\\
    
\begin{figure}
  \centering
    \includegraphics[trim=60bp 25bp 110bp 30bp,clip, width=0.6\textwidth]{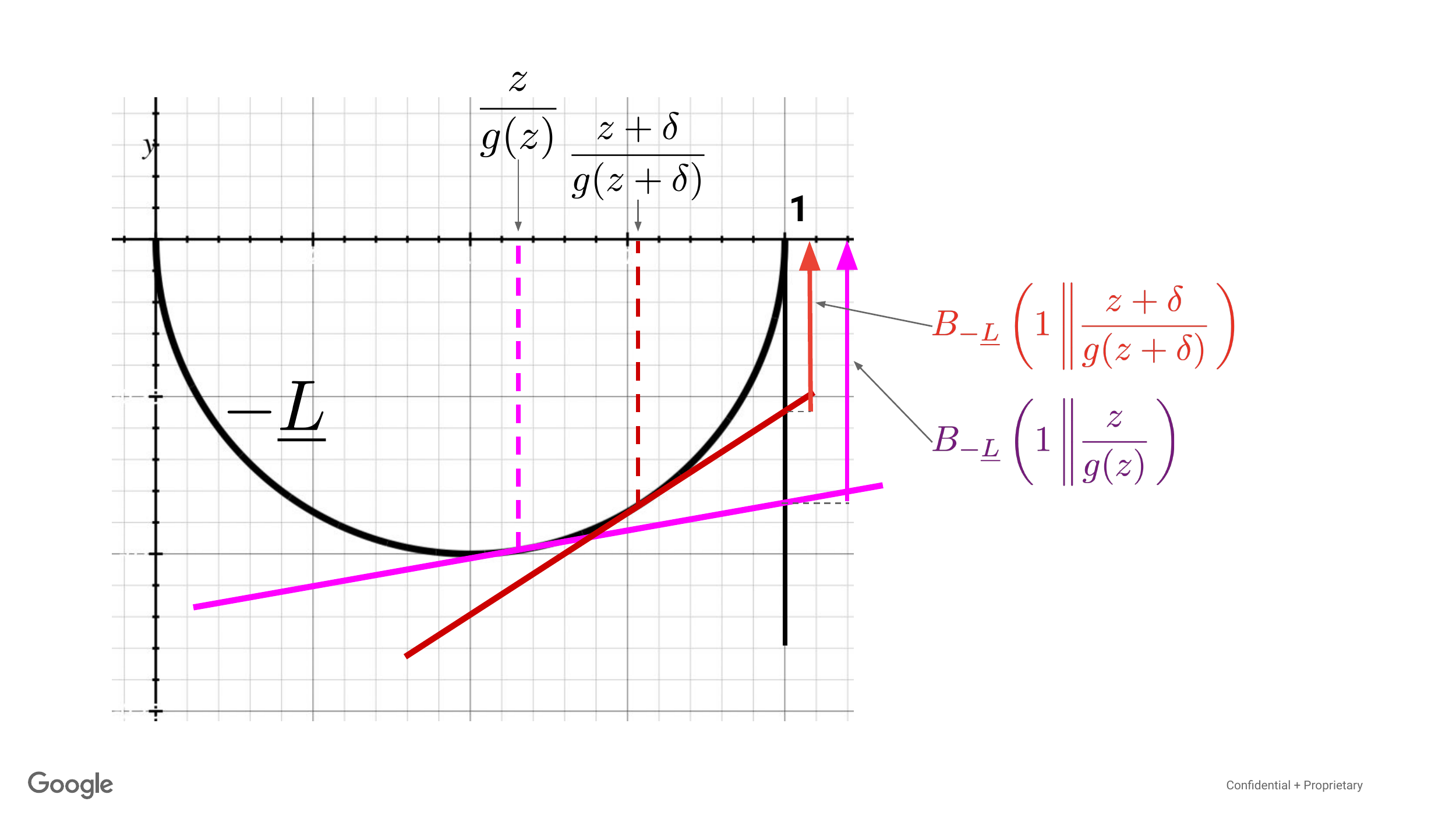}
    \caption{Depiction of the two Bregman divergences in \eqref{eq-breg-cvx}.}
    \label{f-breg}
  \end{figure}

    \noindent \textbf{Remark}: there is a simple argument to show the strict convexity of $\perspectivepobayesrisk$ that relates its derivative to negative a Bregman divergence:
    \begin{eqnarray*}
      \perspectivepobayesrisk'(z) & = & (-\poibayesrisk)\left(\frac{z}{g(z)}\right) + \left(1-\frac{z}{g(z)}\right)\cdot (-\poibayesrisk)'\left(\frac{z}{g(z)}\right)\\
                                  & = & (-\poibayesrisk)\left(1\right) -\left((-\poibayesrisk)\left(1\right) -(-\poibayesrisk)\left(\frac{z}{g(z)}\right) - \left(1-\frac{z}{g(z)}\right)\cdot (-\poibayesrisk)'\left(\frac{z}{g(z)}\right)\right)\\
      & = & - B_{-\poibayesrisk} \left(1 \left\| \frac{z}{g(z)}\right.\right),
    \end{eqnarray*}
    because in our case we have $ (-\poibayesrisk)\left(1\right) = 0$. Let for short $g(z) \defeq z + K$ for $K>0$. We have $g(z) > z$ and so for any $\delta > 0$,
    \begin{eqnarray}
\frac{z+\delta}{g(z+\delta)} = \frac{z+\delta}{g(z)+\delta} > \frac{z}{g(z)}, 
      \end{eqnarray}
      and since $\poibayesrisk$ is strictly concave and $z/g(z) < 1$ for $z > 0$, Bregman divergences lead to:
      \begin{eqnarray}
B_{-\poibayesrisk} \left(1 \left\| \frac{z+\delta}{g(z+\delta)} \right.\right) & < & B_{-\poibayesrisk} \left(1 \left\| \frac{z}{g(z)}\right.\right)\label{eq-breg-cvx}
      \end{eqnarray}
      (See Figure \ref{f-breg} for a depiction of the quantities), hence $\perspectivepobayesrisk'(z+\delta) > \perspectivepobayesrisk'(z)$, showing the derivative of the perspective transform of negative the pointwise Bayes risk is strictly increasing and the function is therefore strictly convex.

\section{Appendix on experiments}\label{app-exps}

\subsection{Examples of generative trees}\label{sec-example}

\begin{figure}
  \centering
  \includegraphics[trim=0bp 0bp 0bp 0bp,clip,width=0.9\columnwidth]{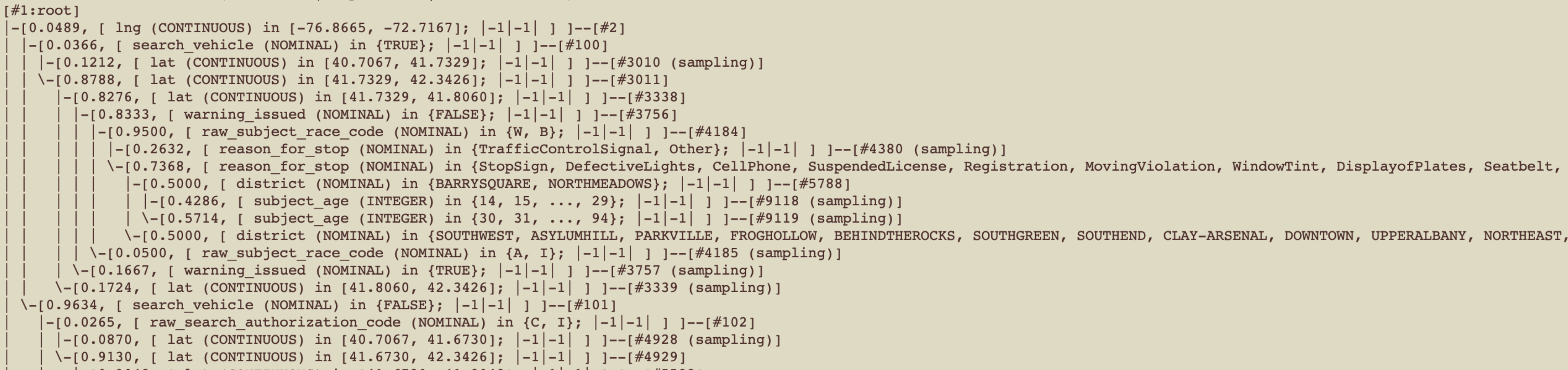} \\
 \includegraphics[trim=0bp 0bp 0bp 0bp,clip,width=0.9\columnwidth]{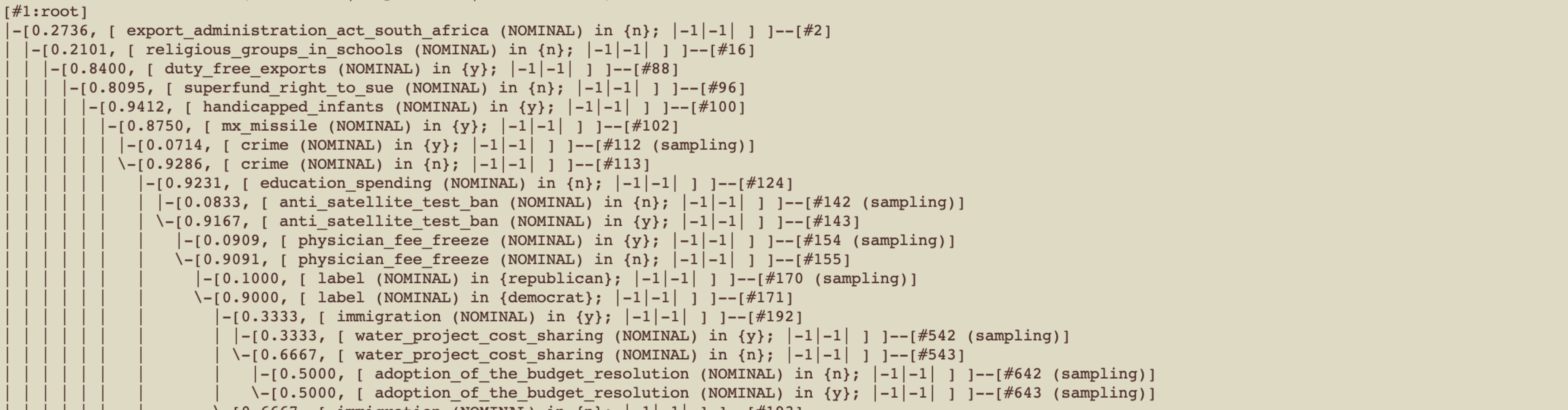} 
    \caption{Generative trees (crop) learned on a fold of Stanford \texttt{open policing} data for Hartford (top) and UCI \texttt{house votes} (bottom).}
    \label{fig:example-gt}
  \end{figure}

  Figure \ref{fig:example-gt} provides examples of subsets of generative trees learned on Stanford \texttt{open policing} data and UCI \texttt{house votes}.  Each node takes the form
  {\scriptsize
\begin{verbatim}
[prob value, [variable (nominal) in {set of nominal values}]; ... ]--[#node name]
[prob value, [variable (continuous) in [continuous interval]]; ... ]--[#node name]
[prob value, [variable (integer) in {int value n, n+1, ..., m}]; ... ]--[#node name]
\end{verbatim}
    }
\texttt{prob value} is the Bernoulli probability associated to the arc pointing to the node. If a node appears as
\begin{verbatim}
[ ... ]--[#node name (sampling)]
\end{verbatim}
then it is a leaf (sampling) node. The rest of the Figures should be self-explanatory. In \texttt{open policing}, notice sampling nodes \texttt{$\#$9118}, \texttt{$\#$9119}, inducing a higher probability of sampling a young person (age within 14 and 29) in the related part of the domain.

\subsection{Domains}\label{sec-doms}

\begin{table}[t]
\begin{center}
\begin{tabular}{|ccc|r|r|r|r|}
\hline \hline
  Domain & Source & Missing data $?$ & \multicolumn{1}{c|}{$m$} & \multicolumn{1}{c|}{$d$} & \multicolumn{1}{c|}{$\#$ Nom.}  & \multicolumn{1}{c|}{$\#$ Num.} \\ \hline
  \domainname{iris} & UCI & No & 150 & 5 & 1 & 4\\
  $^*$\domainname{ringGauss} & -- & No & 1 600 & 2 & -- & 2\\
  $^*$\domainname{circGauss} & -- & No & 2 200 & 2 & -- & 2\\
  $^*$\domainname{gridGauss} & -- & No & 2 500 & 2 & -- & 2\\
  \domainname{house-votes}'84 & UCI & Yes & 435 & 16 & 16 & --\\
  $^*$\domainname{randGauss} & -- & No & 3 800 & 2 & -- & 2\\
  \domainname{led} & UCI & No & 1 000 & 8 & -- & 8\\
\domainname{tictactoe} & UCI & No & 958 & 9  & 9 & --\\
  \domainname{winered} & UCI & No & 1 599 & 12 & 1 & 11 \\
  \domainname{led24} & -- & No & 1 000 & 25 & -- & 25\\
  \domainname{abalone} & UCI & No & 4 177 & 9 & 1 & 3 \\
  \domainname{winewhite} & UCI & No & 4 898 & 12 & 1 & 11 \\
  \domainname{sigma-cabs} & Kaggle & Yes & 5 000 & 13 & 5 & 8 \\
  \domainname{open-policing} & SOP$^{**}$ & Yes & 18 419 & 20 & 16 & 4 \\
  \domainname{dna} & UCI & No & 3 186 & 181 & 181 & -- \\
\hline\hline
\end{tabular}
\end{center}
\caption{Public domains considered in our experiments ($m=$ total number
  of examples, $d=$ number of features), ordered in
  increasing $m \times d$. "Nom." is a shorthand for nominal / ordinal / binary; "Num." stands for integers / reals. ($^*$) = simulated, ($^{**}$ = Hartford data from the Stanford Open Policing Project, \url{https://openpolicing.stanford.edu/}) (see text).}
  \label{t-s-uci}
\end{table}

\domainname{ringGauss} is the seminal 2D ring Gaussians appearing in numerous GAN papers \citep{xzzBG}; those are eight (8) spherical Gaussians with equal covariance, sampling size and centers located on sightlines regularly spaced (2-2 angular distance) and at equal distance from the origin. \domainname{gridGauss} was generated as a decently hard task from \cite{dbplamcAL}: it consists of 25 2D mixture spherical Gaussians with equal variance and sampled sizes, put on a regular grid. \domainname{circGauss} is a Gaussian mode surrounded by a circle, from \cite{xzzBG}. \domainname{randGauss} is a substantially harder version of \domainname{ringGauss} with 16 mixture components, in which covariance, sampling sizes and distances on sightlines from the origin are all random, which creates very substantial discrepances between modes.

\subsection{Algorithms configuration and choice of parameters}\label{sec-doms}

\paragraph{GTs} We have programmed the adversarial and copycat approaches in Java; to simplify the experiments, we only report comparisons with the copycat approach in which the discriminator is \citet{kmOT}'s greedy induction algorithm optimising Matusita's loss. The input of our algorithm to train a generator is a .csv file containing the training data \textit{without any further information}. In particular, each feature's domain is learned from the training data only; while this could surely and trivially be replaced by a user-informed domain for improved results (\textit{e.g.} indicating a proportion's domain as $[0\%,100\%]$, informing the complete list of socio-professional categories, etc.) --- and is in fact standard in some ML packages like \texttt{weka}'s ARFF files, we did not pick this option to alleviate all side information available to the GT learner. Technical details are:
\begin{enumerate}
\item features' domains are computed from training data; our software automatically recognizes three types of variables: nominal, integer and floating point represented\footnote{This is a difference with \texttt{mice} for which categorical variables need to be explicitly stated.};
  \item the discriminator's training follows the top-down induction blueprint in \citet{kmOT};
  \item we do not accept splits that will incur a branching probability $p\in \{0,1\}$ to prevent discarding support; leaves are split from the heaviest first; when a real examples with missing values branch in the decision tree on one of its missing values, the probability of following the left / right arc is \textit{not} 1/2 but takes into account the length of the left / right domains of the variable at the split (for example, if the left branch's variable domain is $\{A,B,C\}$ and the right one is $\{D\}$ for a nominal variable, then the left branching probability is 3/4);
  \item we consider three basic sizes of GTs, corresponding to 10, 300 and a \textsc{max} = 10 000 nodes splits (thus with tottal number of nodes equal to 21, 601 and no more than 20 001). In the \textsc{impute} experiment, we only use the \textsc{max} size. Note that the \textsc{max} size usually is smaller than 20 001 on small datasets because of the support constraint in [2].
  \end{enumerate}

  \paragraph{\texttt{mice}} We have used the R \texttt{mice} package V 3.13.0 with three choices of methods for the round robin (column-wise) prediction of missing values: \texttt{cart} \cite{bfosCA}, \texttt{norm} and random forests (\texttt{rf}) \cite{vgMM}. In that last case, we have replaced the default number of trees (10) by a larger number (100) to get better results. We use the default number of round-robin iterations (5).

  \paragraph{\textsc{CT-GAN}} We have used the Python implementation\footnote{\url{https://github.com/sdv-dev/CTGAN}} with default values.

  \paragraph{\textsc{TensorFlow}} To learn the additional Random Forests and Gradient Boosted Decision Trees involved in experiments \textsc{train-synth}, \textsc{synth-discrim} and \textsc{synth-aug}, we used Tensorflow Decision Forests library\footnote{\url{https://github.com/google/yggdrasil-decision-forests/blob/main/documentation/learners.md}}. The important points are:
  \begin{itemize}
\item for Random Forests, we use 300 trees with max depth 16. Attribute sampling: sqrt(number attributes) for classification problems, number attributes / 3 for regression problems (Breiman rule of thumb);
\item for Gradient Boosted Decision Trees, we use max 300 trees, with 10$\%$ of the training dataset for validation and early stopping. Max depth is 6 and there is no attribute sampling;
  \item in both cases, the min $\#$examples per leaf is 5, we use CART to find splits on numerical and categorical features (\textit{i.e.} we don't use one-hot encoding for categorical features). Induction is top-down.
    \end{itemize}

  \paragraph{Computers used} We ran part of the experiments on a Mac Book Pro 16 Gb RAM w/ 2 GHz Quad-Core Intel(R) Core i5(R) processor, and part on a desktop Intel(R) Xeon(R) 3.70GHz with 12 cores and 64 Gb RAM.

\subsection{Data generation experiments}\label{sec-datagen}

Figures \ref{fig:exp-randgauss}, \ref{fig:exp-gridgauss}, \ref{fig:exp-circgauss}, \ref{fig:exp-ringgauss}, \ref{fig:exp-iris-1}, \ref{fig:exp-iris-2}, \ref{fig:exp-winered-1}, \ref{fig:exp-winered-2}, \ref{fig:exp-sigma-cabs} present 2D heatmap of density learned by generative trees with shown total number of nodes. We have run a simple 10-fold CV experiment, each plot being the generator that minimizes over all folds an empirical $\chi^2$ between the training data and a set of generated data. For UCI domains, the variables plotted are indicated and we indicate the number ($m'$) of examples generated to build the plots.

\clearpage

\setlength\tabcolsep{0.5pt}
\bgroup
\def\arraystretch{0.2}
\begin{figure}
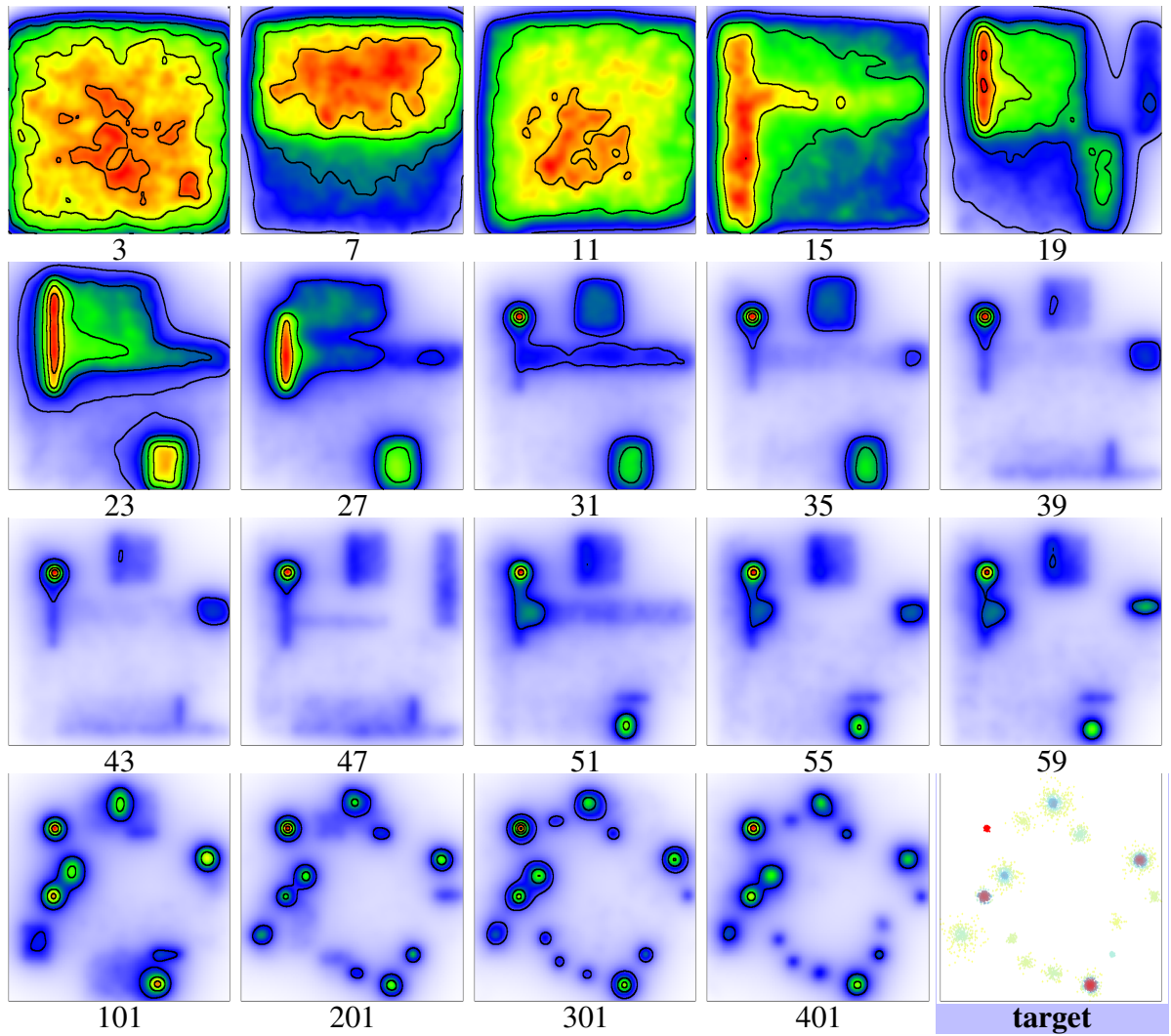

  \centering
  \begin{tabular}{ccccc}
    \imagebig{randgauss-density}{Sep_11th__7h_48m_51s}{3}{}{4000} & \imagebig{randgauss-density}{Sep_11th__7h_48m_51s}{7}{}{4000} & \imagebig{randgauss-density}{Sep_11th__7h_48m_51s}{11}{}{4000} & \imagebig{randgauss-density}{Sep_11th__7h_48m_51s}{15}{}{4000} & \imagebig{randgauss-density}{Sep_11th__7h_48m_51s}{19}{}{4000} \\
    3 & 7 & 11 & 15 & 19\\
    \imagebig{randgauss-density}{Sep_11th__7h_48m_51s}{23}{}{4000} & \imagebig{randgauss-density}{Sep_11th__7h_48m_51s}{27}{}{4000} & \imagebig{randgauss-density}{Sep_11th__7h_48m_51s}{31}{}{4000} & \imagebig{randgauss-density}{Sep_11th__7h_48m_51s}{35}{}{4000} & \imagebig{randgauss-density}{Sep_11th__7h_48m_51s}{39}{}{4000} \\
    23 & 27 & 31 & 35 & 39\\
    \imagebig{randgauss-density}{Sep_11th__7h_48m_51s}{43}{}{4000} & \imagebig{randgauss-density}{Sep_11th__7h_48m_51s}{47}{}{4000} & \imagebig{randgauss-density}{Sep_11th__7h_48m_51s}{51}{}{4000} & \imagebig{randgauss-density}{Sep_11th__7h_48m_51s}{55}{}{4000} & \imagebig{randgauss-density}{Sep_11th__7h_48m_51s}{59}{}{4000} \\
    43 & 47 & 51 & 55 & 59\\
    \imagebig{randgauss-density}{Sep_11th__7h_48m_51s}{101}{}{4000} & \imagebig{randgauss-density}{Sep_11th__7h_48m_51s}{201}{}{4000} & \imagebig{randgauss-density}{Sep_11th__7h_48m_51s}{301}{}{4000} & \imagebig{randgauss-density}{Sep_11th__7h_48m_51s}{401}{}{4000} & \cellcolor{blue!25}{\imagemodel{randgauss/data-randgauss-plot}} \\
     101 & 201 & 301 & 401 & \cellcolor{blue!25}{\textbf{target}}
   \end{tabular}
    \caption{Results on the \texttt{randGauss} simulated data (target in the bottom-right, $m = 3800$; colors indicate sampled density), for copycat training. Numbers are the total number of nodes of the generators; generators sampled for $m' = 4000$ points each, each plot shows results for one of the ten generators in the CV folds (not necessarily from the same fold); training method = copycat.}
    \label{fig:exp-randgauss}
  \end{figure}
\egroup

\clearpage

\bgroup
\def\arraystretch{0.2}
\begin{figure}
  \centering
  \begin{tabular}{ccccc}
    \imagebig{gridgauss-density}{Sep_10th__13h_15m_4s}{3}{}{4000} & \imagebig{gridgauss-density}{Sep_10th__13h_15m_4s}{7}{}{4000} & \imagebig{gridgauss-density}{Sep_10th__13h_15m_4s}{11}{}{4000} & \imagebig{gridgauss-density}{Sep_10th__13h_15m_4s}{15}{}{4000} & \imagebig{gridgauss-density}{Sep_10th__13h_15m_4s}{19}{}{4000} \\
    3 & 7 & 11 & 15 & 19\\
    \imagebig{gridgauss-density}{Sep_10th__13h_15m_4s}{23}{}{4000} & \imagebig{gridgauss-density}{Sep_10th__13h_15m_4s}{27}{}{4000} & \imagebig{gridgauss-density}{Sep_10th__13h_15m_4s}{31}{}{4000} & \imagebig{gridgauss-density}{Sep_10th__13h_15m_4s}{35}{}{4000} & \imagebig{gridgauss-density}{Sep_10th__13h_15m_4s}{39}{}{4000} \\
    23 & 27 & 31 & 35 & 39\\
    \imagebig{gridgauss-density}{Sep_10th__13h_15m_4s}{43}{}{4000} & \imagebig{gridgauss-density}{Sep_10th__13h_15m_4s}{47}{}{4000} & \imagebig{gridgauss-density}{Sep_10th__13h_15m_4s}{51}{}{4000} & \imagebig{gridgauss-density}{Sep_10th__13h_15m_4s}{55}{}{4000} & \imagebig{gridgauss-density}{Sep_10th__13h_15m_4s}{59}{}{4000} \\
    43 & 47 & 51 & 55 & 59\\
    \imagebig{gridgauss-density}{Sep_10th__13h_15m_4s}{91}{}{4000} & \imagebig{gridgauss-density}{Sep_10th__13h_15m_4s}{101}{}{4000} & \imagebig{gridgauss-density}{Sep_10th__13h_15m_4s}{201}{}{4000} & \imagebig{gridgauss-density}{Sep_10th__13h_15m_4s}{301}{}{4000} & \cellcolor{blue!25}{\imagemodel{gridgauss/data-gridgauss-plot}}\\
     91 & 101 & 201 & 301 & \cellcolor{blue!25}{\textbf{target}}
   \end{tabular}
    \caption{Results on the \texttt{gridGauss} simulated data (with $m = 2500, m' = 4000$), convention follows Fig. \ref{fig:exp-randgauss}.}
    \label{fig:exp-gridgauss}
  \end{figure}
  \egroup
  
\clearpage

\bgroup
\def\arraystretch{0.2} 
\begin{figure}
  \centering
  \begin{tabular}{ccccc}
    \imagebig{circgauss-density}{Sep_22th__6h_56m_9s}{3}{_X_X_Y_Y}{2200} & \imagebig{circgauss-density}{Sep_22th__6h_56m_9s}{7}{_X_X_Y_Y}{2200} & \imagebig{circgauss-density}{Sep_22th__6h_56m_9s}{11}{_X_X_Y_Y}{2200} & \imagebig{circgauss-density}{Sep_22th__6h_56m_9s}{15}{_X_X_Y_Y}{2200} & \imagebig{circgauss-density}{Sep_22th__6h_56m_9s}{19}{_X_X_Y_Y}{2200} \\
    3 & 7 & 11 & 15 & 19\\
    \imagebig{circgauss-density}{Sep_22th__6h_56m_9s}{23}{_X_X_Y_Y}{2200} & \imagebig{circgauss-density}{Sep_22th__6h_56m_9s}{27}{_X_X_Y_Y}{2200} & \imagebig{circgauss-density}{Sep_22th__6h_56m_9s}{31}{_X_X_Y_Y}{2200} & \imagebig{circgauss-density}{Sep_22th__6h_56m_9s}{35}{_X_X_Y_Y}{2200} & \imagebig{circgauss-density}{Sep_22th__6h_56m_9s}{39}{_X_X_Y_Y}{2200} \\
    23 & 27 & 31 & 35 & 39\\
    \imagebig{circgauss-density}{Sep_22th__6h_56m_9s}{43}{_X_X_Y_Y}{2200} & \imagebig{circgauss-density}{Sep_22th__6h_56m_9s}{47}{_X_X_Y_Y}{2200} & \imagebig{circgauss-density}{Sep_22th__6h_56m_9s}{51}{_X_X_Y_Y}{2200} & \imagebig{circgauss-density}{Sep_22th__6h_56m_9s}{55}{_X_X_Y_Y}{2200} & \imagebig{circgauss-density}{Sep_22th__6h_56m_9s}{59}{_X_X_Y_Y}{2200} \\
    43 & 47 & 51 & 55 & 59\\
    \imagebig{circgauss-density}{Sep_22th__6h_56m_9s}{91}{_X_X_Y_Y}{2200} & \imagebig{circgauss-density}{Sep_22th__6h_56m_9s}{101}{_X_X_Y_Y}{2200} & \imagebig{circgauss-density}{Sep_22th__6h_56m_9s}{201}{_X_X_Y_Y}{2200} & \imagebig{circgauss-density}{Sep_22th__6h_56m_9s}{301}{_X_X_Y_Y}{2200} & \cellcolor{blue!25}{\imagemodel{circgauss-density/data-circgauss-plot}}\\
     91 & 101 & 201 & 301 & \cellcolor{blue!25}{\textbf{target}}
   \end{tabular}
    \caption{Results on the \texttt{circGauss} simulated data (with $m = m' = 2200$), convention follows Fig. \ref{fig:exp-randgauss}.}
    \label{fig:exp-circgauss}
  \end{figure}
\egroup

\clearpage

\bgroup
\def\arraystretch{0.2}
\begin{figure}
  \centering
  \begin{tabular}{ccccc}
    \imagebig{ringgauss-density}{Sep_10th__16h_38m_55s}{3}{}{4000} & \imagebig{ringgauss-density}{Sep_10th__16h_38m_55s}{7}{}{4000} & \imagebig{ringgauss-density}{Sep_10th__16h_38m_55s}{11}{}{4000} & \imagebig{ringgauss-density}{Sep_10th__16h_38m_55s}{15}{}{4000} & \imagebig{ringgauss-density}{Sep_10th__16h_38m_55s}{19}{}{4000} \\
    3 & 7 & 11 & 15 & 19\\
    \imagebig{ringgauss-density}{Sep_10th__16h_38m_55s}{23}{}{4000} & \imagebig{ringgauss-density}{Sep_10th__16h_38m_55s}{27}{}{4000} & \imagebig{ringgauss-density}{Sep_10th__16h_38m_55s}{31}{}{4000} & \imagebig{ringgauss-density}{Sep_10th__16h_38m_55s}{35}{}{4000} & \imagebig{ringgauss-density}{Sep_10th__16h_38m_55s}{39}{}{4000} \\
    23 & 27 & 31 & 35 & 39\\
    \imagebig{ringgauss-density}{Sep_10th__16h_38m_55s}{43}{}{4000} & \imagebig{ringgauss-density}{Sep_10th__16h_38m_55s}{47}{}{4000} & \imagebig{ringgauss-density}{Sep_10th__16h_38m_55s}{51}{}{4000} & \imagebig{ringgauss-density}{Sep_10th__16h_38m_55s}{55}{}{4000} & \imagebig{ringgauss-density}{Sep_10th__16h_38m_55s}{59}{}{4000} \\
    43 & 47 & 51 & 55 & 59\\
    \imagebig{ringgauss-density}{Sep_10th__16h_38m_55s}{91}{}{4000} & \imagebig{ringgauss-density}{Sep_10th__16h_38m_55s}{101}{}{4000} & \imagebig{ringgauss-density}{Sep_10th__16h_38m_55s}{201}{}{4000} & \imagebig{ringgauss-density}{Sep_10th__16h_38m_55s}{301}{}{4000} & \cellcolor{blue!25}{\imagemodel{ringgauss/data-ringgauss-plot}}\\
     91 & 101 & 201 & 301 & \cellcolor{blue!25}{\textbf{target}}
   \end{tabular}
    \caption{Results on the \texttt{ringGauss} simulated data (with $m = 1500, m' = 4000$), convention follows Fig. \ref{fig:exp-randgauss}.}
    \label{fig:exp-ringgauss}
  \end{figure}
\egroup

\clearpage


\bgroup
\def\arraystretch{0.2}
\begin{figure}
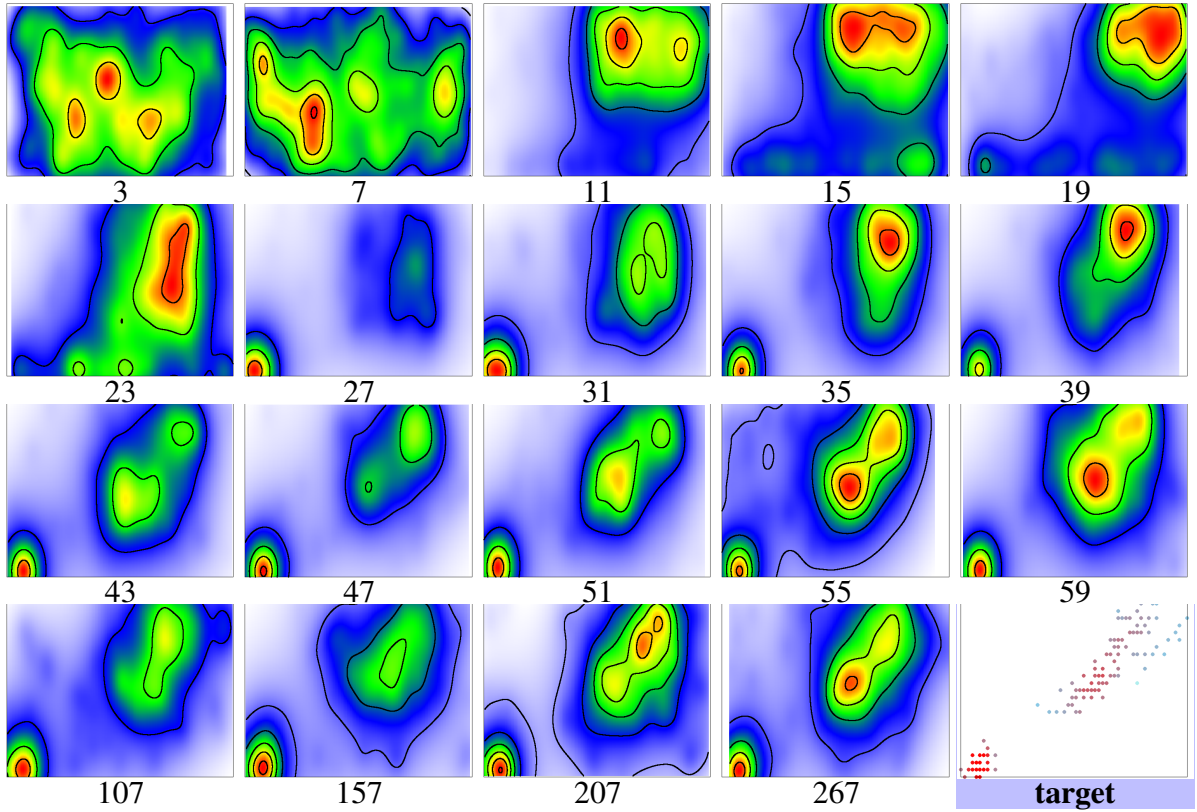

  \centering
  \begin{tabular}{ccccc}
    \imagebignosquare{iris}{Sep_15th__11h_35m_51s}{3}{_X_petal_length_Y_petal_width}{150} & \imagebignosquare{iris}{Sep_15th__11h_35m_51s}{7}{_X_petal_length_Y_petal_width}{150} & \imagebignosquare{iris}{Sep_15th__11h_35m_51s}{11}{_X_petal_length_Y_petal_width}{150} & \imagebignosquare{iris}{Sep_15th__11h_35m_51s}{15}{_X_petal_length_Y_petal_width}{150} & \imagebignosquare{iris}{Sep_15th__11h_35m_51s}{19}{_X_petal_length_Y_petal_width}{150} \\
    3 & 7 & 11 & 15 & 19\\
    \imagebignosquare{iris}{Sep_15th__11h_35m_51s}{23}{_X_petal_length_Y_petal_width}{150} & \imagebignosquare{iris}{Sep_15th__11h_35m_51s}{27}{_X_petal_length_Y_petal_width}{150} & \imagebignosquare{iris}{Sep_15th__11h_35m_51s}{31}{_X_petal_length_Y_petal_width}{150} & \imagebignosquare{iris}{Sep_15th__11h_35m_51s}{35}{_X_petal_length_Y_petal_width}{150} & \imagebignosquare{iris}{Sep_15th__11h_35m_51s}{39}{_X_petal_length_Y_petal_width}{150} \\
    23 & 27 & 31 & 35 & 39\\
    \imagebignosquare{iris}{Sep_15th__11h_35m_51s}{43}{_X_petal_length_Y_petal_width}{150} & \imagebignosquare{iris}{Sep_15th__11h_35m_51s}{47}{_X_petal_length_Y_petal_width}{150} & \imagebignosquare{iris}{Sep_15th__11h_35m_51s}{51}{_X_petal_length_Y_petal_width}{150} & \imagebignosquare{iris}{Sep_15th__11h_35m_51s}{55}{_X_petal_length_Y_petal_width}{150} & \imagebignosquare{iris}{Sep_15th__11h_35m_51s}{59}{_X_petal_length_Y_petal_width}{150} \\
    43 & 47 & 51 & 55 & 59\\
    \imagebignosquare{iris}{Sep_15th__11h_35m_51s}{107}{_X_petal_length_Y_petal_width}{150} & \imagebignosquare{iris}{Sep_15th__11h_35m_51s}{157}{_X_petal_length_Y_petal_width}{150} & \imagebignosquare{iris}{Sep_15th__11h_35m_51s}{207}{_X_petal_length_Y_petal_width}{150} & \imagebignosquare{iris}{Sep_15th__11h_35m_51s}{267}{_X_petal_length_Y_petal_width}{150} & \cellcolor{blue!25}{\imagemodelnosquare{iris/iris-petal-length-petal-width}}\\
     107 & 157 & 207 & 267 & \cellcolor{blue!25}{\textbf{target}}
   \end{tabular}
    \caption{Results on UCI \texttt{iris} domain (with $m = 150, m' = 150$) for the 2D plane \texttt{petal-length} $\times$ \texttt{petal-width}, convention follows Fig. \ref{fig:exp-randgauss}.}
    \label{fig:exp-iris-1}
  \end{figure}
\egroup

\bgroup
\def\arraystretch{0.2}
\begin{figure}
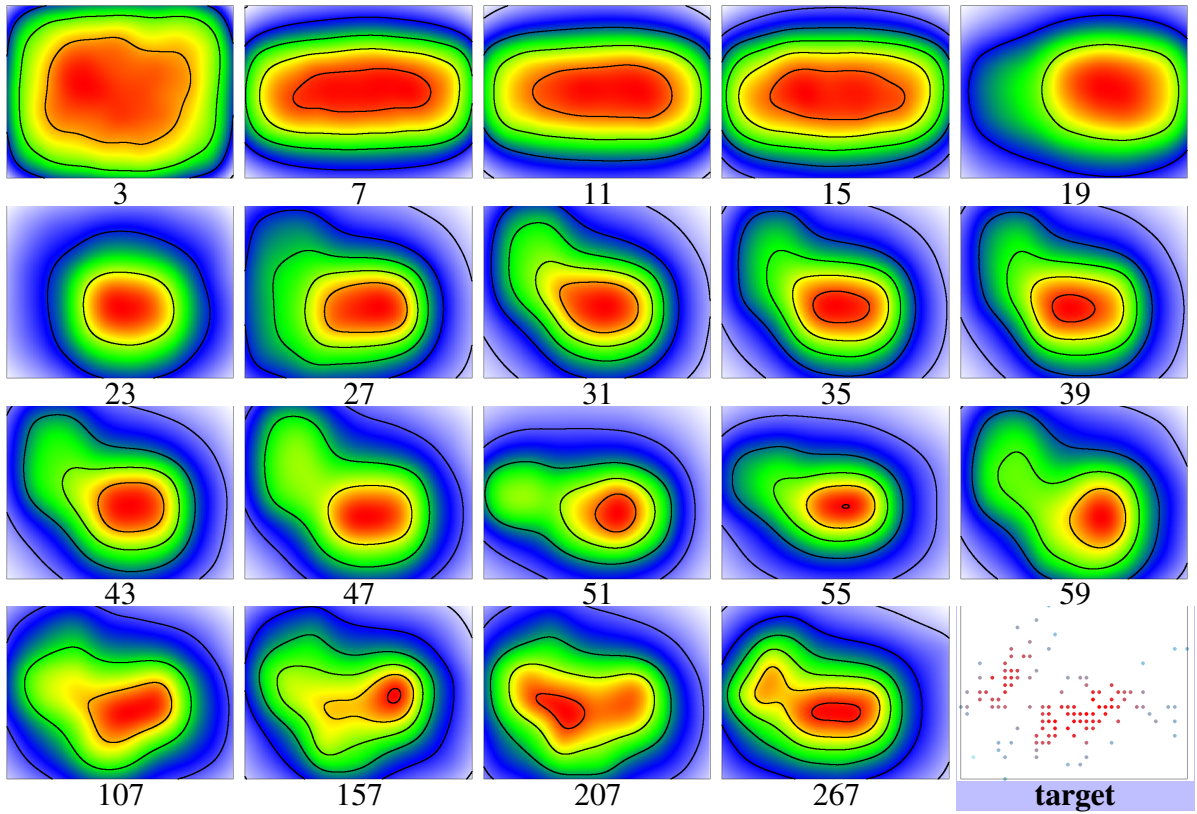

  \centering
  \begin{tabular}{ccccc}
    \imagebignosquare{iris}{Sep_15th__11h_35m_51s}{3}{_X_sepal_length_Y_sepal_width}{4000} & \imagebignosquare{iris}{Sep_15th__11h_35m_51s}{7}{_X_sepal_length_Y_sepal_width}{4000} & \imagebignosquare{iris}{Sep_15th__11h_35m_51s}{11}{_X_sepal_length_Y_sepal_width}{4000} & \imagebignosquare{iris}{Sep_15th__11h_35m_51s}{15}{_X_sepal_length_Y_sepal_width}{4000} & \imagebignosquare{iris}{Sep_15th__11h_35m_51s}{19}{_X_sepal_length_Y_sepal_width}{4000} \\
    3 & 7 & 11 & 15 & 19\\
    \imagebignosquare{iris}{Sep_15th__11h_35m_51s}{23}{_X_sepal_length_Y_sepal_width}{4000} & \imagebignosquare{iris}{Sep_15th__11h_35m_51s}{27}{_X_sepal_length_Y_sepal_width}{4000} & \imagebignosquare{iris}{Sep_15th__11h_35m_51s}{31}{_X_sepal_length_Y_sepal_width}{4000} & \imagebignosquare{iris}{Sep_15th__11h_35m_51s}{35}{_X_sepal_length_Y_sepal_width}{4000} & \imagebignosquare{iris}{Sep_15th__11h_35m_51s}{39}{_X_sepal_length_Y_sepal_width}{4000} \\
    23 & 27 & 31 & 35 & 39\\
    \imagebignosquare{iris}{Sep_15th__11h_35m_51s}{43}{_X_sepal_length_Y_sepal_width}{4000} & \imagebignosquare{iris}{Sep_15th__11h_35m_51s}{47}{_X_sepal_length_Y_sepal_width}{4000} & \imagebignosquare{iris}{Sep_15th__11h_35m_51s}{51}{_X_sepal_length_Y_sepal_width}{4000} & \imagebignosquare{iris}{Sep_15th__11h_35m_51s}{55}{_X_sepal_length_Y_sepal_width}{4000} & \imagebignosquare{iris}{Sep_15th__11h_35m_51s}{59}{_X_sepal_length_Y_sepal_width}{4000} \\
    43 & 47 & 51 & 55 & 59\\
    \imagebignosquare{iris}{Sep_15th__11h_35m_51s}{107}{_X_sepal_length_Y_sepal_width}{4000} & \imagebignosquare{iris}{Sep_15th__11h_35m_51s}{157}{_X_sepal_length_Y_sepal_width}{4000} & \imagebignosquare{iris}{Sep_15th__11h_35m_51s}{207}{_X_sepal_length_Y_sepal_width}{4000} & \imagebignosquare{iris}{Sep_15th__11h_35m_51s}{267}{_X_sepal_length_Y_sepal_width}{4000} & \cellcolor{blue!25}{\imagemodelnosquare{iris/iris-sepal-length-sepal-width}}\\
     107 & 157 & 207 & 267 & \cellcolor{blue!25}{\textbf{target}}
   \end{tabular}
    \caption{Results on UCI \texttt{iris} domain (with $m = 150, m' = 4000$) for the 2D plane \texttt{sepal-length} $\times$ \texttt{sepal-width}, convention follows Fig. \ref{fig:exp-randgauss}.}
    \label{fig:exp-iris-2}
  \end{figure}
\egroup

\clearpage

\bgroup
\def\arraystretch{0.2}
\begin{figure}
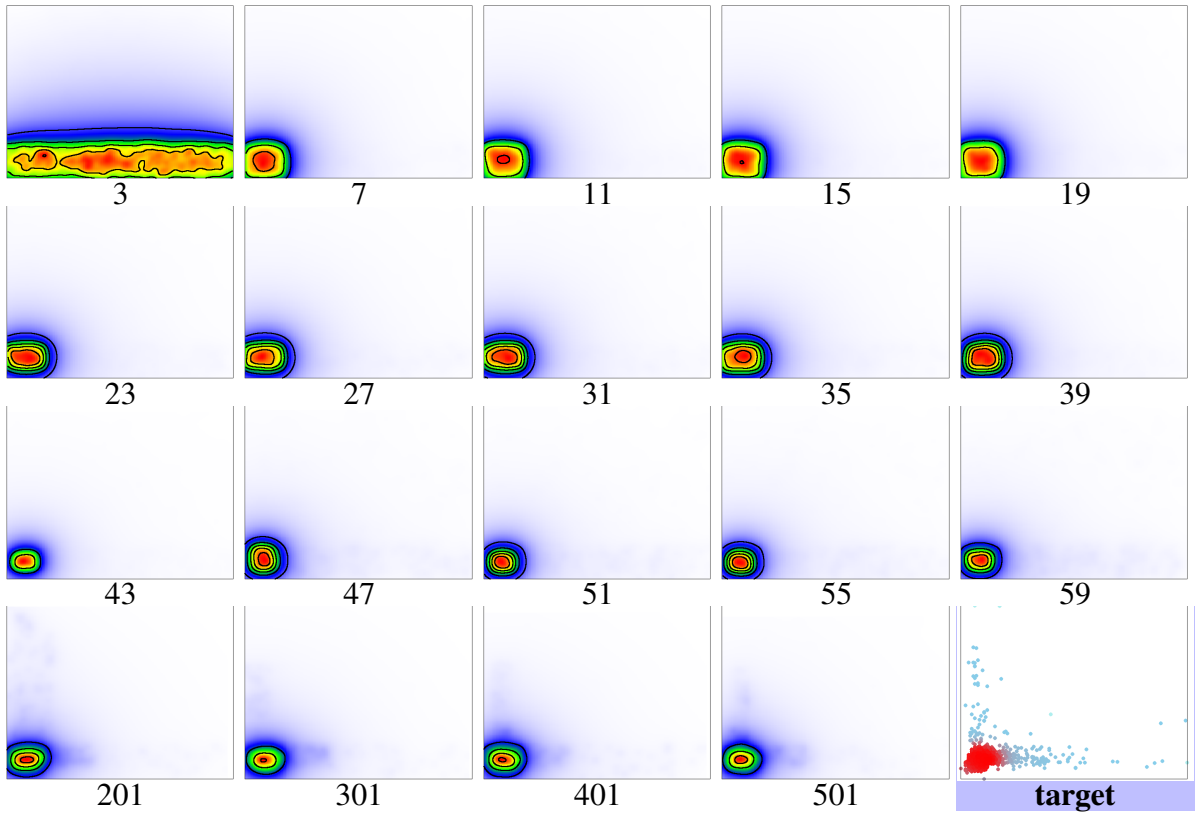

  \centering
  \begin{tabular}{ccccc}
    \imagebignosquare{winered}{Sep_15th__14h_24m_17s}{3}{_X_residual_sugar_Y_chlorides}{1599} & \imagebignosquare{winered}{Sep_15th__14h_24m_17s}{7}{_X_residual_sugar_Y_chlorides}{1599} & \imagebignosquare{winered}{Sep_15th__14h_24m_17s}{11}{_X_residual_sugar_Y_chlorides}{1599} & \imagebignosquare{winered}{Sep_15th__14h_24m_17s}{15}{_X_residual_sugar_Y_chlorides}{1599} & \imagebignosquare{winered}{Sep_15th__14h_24m_17s}{19}{_X_residual_sugar_Y_chlorides}{1599} \\
    3 & 7 & 11 & 15 & 19\\
    \imagebignosquare{winered}{Sep_15th__14h_24m_17s}{23}{_X_residual_sugar_Y_chlorides}{1599} & \imagebignosquare{winered}{Sep_15th__14h_24m_17s}{27}{_X_residual_sugar_Y_chlorides}{1599} & \imagebignosquare{winered}{Sep_15th__14h_24m_17s}{31}{_X_residual_sugar_Y_chlorides}{1599} & \imagebignosquare{winered}{Sep_15th__14h_24m_17s}{35}{_X_residual_sugar_Y_chlorides}{1599} & \imagebignosquare{winered}{Sep_15th__14h_24m_17s}{39}{_X_residual_sugar_Y_chlorides}{1599} \\
    23 & 27 & 31 & 35 & 39\\
    \imagebignosquare{winered}{Sep_15th__14h_24m_17s}{43}{_X_residual_sugar_Y_chlorides}{1599} & \imagebignosquare{winered}{Sep_15th__14h_24m_17s}{47}{_X_residual_sugar_Y_chlorides}{1599} & \imagebignosquare{winered}{Sep_15th__14h_24m_17s}{51}{_X_residual_sugar_Y_chlorides}{1599} & \imagebignosquare{winered}{Sep_15th__14h_24m_17s}{55}{_X_residual_sugar_Y_chlorides}{1599} & \imagebignosquare{winered}{Sep_15th__14h_24m_17s}{59}{_X_residual_sugar_Y_chlorides}{1599} \\
    43 & 47 & 51 & 55 & 59\\
    \imagebignosquare{winered}{Sep_15th__14h_24m_17s}{201}{_X_residual_sugar_Y_chlorides}{1599} & \imagebignosquare{winered}{Sep_15th__14h_24m_17s}{301}{_X_residual_sugar_Y_chlorides}{1599} & \imagebignosquare{winered}{Sep_15th__14h_24m_17s}{401}{_X_residual_sugar_Y_chlorides}{1599} & \imagebignosquare{winered}{Sep_15th__14h_24m_17s}{501}{_X_residual_sugar_Y_chlorides}{1599} & \cellcolor{blue!25}{\imagemodelnosquare{winered/winered-x-residual-sugar-y-chlorides}}\\
     201 & 301 & 401 & 501 & \cellcolor{blue!25}{\textbf{target}}
   \end{tabular}
    \caption{Results on UCI \texttt{winered} domain (with $m = m' = 1599$) for the 2D plane \texttt{residual-sugar} $\times$ \texttt{chlorides}, convention follows Fig. \ref{fig:exp-randgauss}.}
    \label{fig:exp-winered-1}
  \end{figure}
\egroup

\bgroup
\def\arraystretch{0.2}
\begin{figure}
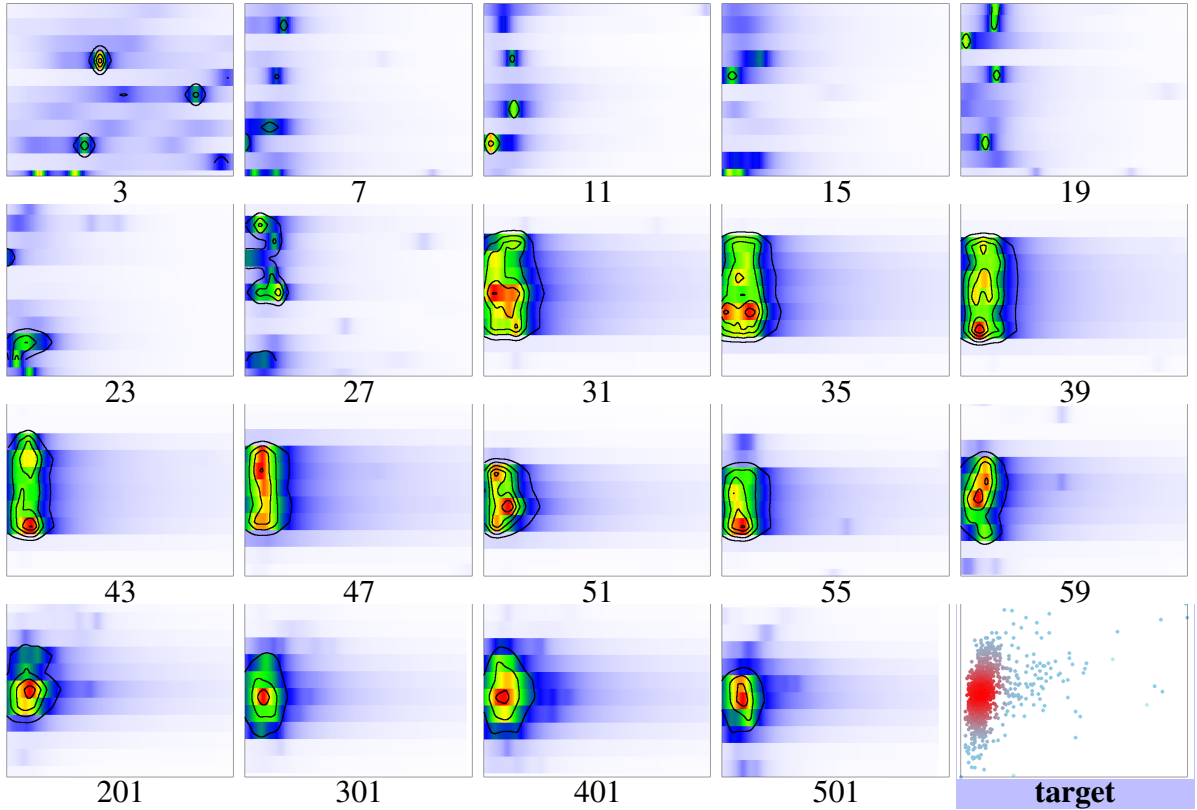

  \centering
  \begin{tabular}{ccccc}
    \imagebignosquare{winered}{Sep_15th__14h_24m_17s}{3}{_X_residual_sugar_Y_density}{1599} & \imagebignosquare{winered}{Sep_15th__14h_24m_17s}{7}{_X_residual_sugar_Y_density}{1599} & \imagebignosquare{winered}{Sep_15th__14h_24m_17s}{11}{_X_residual_sugar_Y_density}{1599} & \imagebignosquare{winered}{Sep_15th__14h_24m_17s}{15}{_X_residual_sugar_Y_density}{1599} & \imagebignosquare{winered}{Sep_15th__14h_24m_17s}{19}{_X_residual_sugar_Y_density}{1599} \\
    3 & 7 & 11 & 15 & 19\\
    \imagebignosquare{winered}{Sep_15th__14h_24m_17s}{23}{_X_residual_sugar_Y_density}{1599} & \imagebignosquare{winered}{Sep_15th__14h_24m_17s}{27}{_X_residual_sugar_Y_density}{1599} & \imagebignosquare{winered}{Sep_15th__14h_24m_17s}{31}{_X_residual_sugar_Y_density}{1599} & \imagebignosquare{winered}{Sep_15th__14h_24m_17s}{35}{_X_residual_sugar_Y_density}{1599} & \imagebignosquare{winered}{Sep_15th__14h_24m_17s}{39}{_X_residual_sugar_Y_density}{1599} \\
    23 & 27 & 31 & 35 & 39\\
    \imagebignosquare{winered}{Sep_15th__14h_24m_17s}{43}{_X_residual_sugar_Y_density}{1599} & \imagebignosquare{winered}{Sep_15th__14h_24m_17s}{47}{_X_residual_sugar_Y_density}{1599} & \imagebignosquare{winered}{Sep_15th__14h_24m_17s}{51}{_X_residual_sugar_Y_density}{1599} & \imagebignosquare{winered}{Sep_15th__14h_24m_17s}{55}{_X_residual_sugar_Y_density}{1599} & \imagebignosquare{winered}{Sep_15th__14h_24m_17s}{59}{_X_residual_sugar_Y_density}{1599} \\
    43 & 47 & 51 & 55 & 59\\
    \imagebignosquare{winered}{Sep_15th__14h_24m_17s}{201}{_X_residual_sugar_Y_density}{1599} & \imagebignosquare{winered}{Sep_15th__14h_24m_17s}{301}{_X_residual_sugar_Y_density}{1599} & \imagebignosquare{winered}{Sep_15th__14h_24m_17s}{401}{_X_residual_sugar_Y_density}{1599} & \imagebignosquare{winered}{Sep_15th__14h_24m_17s}{501}{_X_residual_sugar_Y_density}{1599} & \cellcolor{blue!25}{\imagemodelnosquare{winered/winered-x-residual-sugar-y-density}}\\
     201 & 301 & 401 & 501 & \cellcolor{blue!25}{\textbf{target}}
   \end{tabular}
    \caption{Results on UCI \texttt{winered} domain (with $m = m' = 1599$) for the 2D plane \texttt{residual-sugar} $\times$ \texttt{density}, convention follows Fig. \ref{fig:exp-randgauss}. The discontinuous strip-look is due to the huge difference in scales for the variables.}
    \label{fig:exp-winered-2}
  \end{figure}
\egroup

\clearpage


\bgroup 
\def\arraystretch{0.2} 
\begin{figure} 
  \centering 
  \begin{tabular}{ccccc}
    \imagebignosquare{sigma-cabs}{Sep_18th__11h_43m_4s}{3}{_X_Trip_Distance_Y_Life_Style_Index}{649} & \imagebignosquare{sigma-cabs}{Sep_18th__11h_43m_4s}{7}{_X_Trip_Distance_Y_Life_Style_Index}{649} & \imagebignosquare{sigma-cabs}{Sep_18th__11h_43m_4s}{11}{_X_Trip_Distance_Y_Life_Style_Index}{649} & \imagebignosquare{sigma-cabs}{Sep_18th__11h_43m_4s}{15}{_X_Trip_Distance_Y_Life_Style_Index}{649} & \imagebignosquare{sigma-cabs}{Sep_18th__11h_43m_4s}{19}{_X_Trip_Distance_Y_Life_Style_Index}{649} \\
    3 & 7 & 11 & 15 & 19\\
    \imagebignosquare{sigma-cabs}{Sep_18th__11h_43m_4s}{23}{_X_Trip_Distance_Y_Life_Style_Index}{649} & \imagebignosquare{sigma-cabs}{Sep_18th__11h_43m_4s}{27}{_X_Trip_Distance_Y_Life_Style_Index}{649} & \imagebignosquare{sigma-cabs}{Sep_18th__11h_43m_4s}{31}{_X_Trip_Distance_Y_Life_Style_Index}{649} & \imagebignosquare{sigma-cabs}{Sep_18th__11h_43m_4s}{35}{_X_Trip_Distance_Y_Life_Style_Index}{649} & \imagebignosquare{sigma-cabs}{Sep_18th__11h_43m_4s}{39}{_X_Trip_Distance_Y_Life_Style_Index}{649} \\
    23 & 27 & 31 & 35 & 39\\
    \imagebignosquare{sigma-cabs}{Sep_18th__11h_43m_4s}{43}{_X_Trip_Distance_Y_Life_Style_Index}{649} & \imagebignosquare{sigma-cabs}{Sep_18th__11h_43m_4s}{47}{_X_Trip_Distance_Y_Life_Style_Index}{649} & \imagebignosquare{sigma-cabs}{Sep_18th__11h_43m_4s}{51}{_X_Trip_Distance_Y_Life_Style_Index}{649} & \imagebignosquare{sigma-cabs}{Sep_18th__11h_43m_4s}{55}{_X_Trip_Distance_Y_Life_Style_Index}{649} & \imagebignosquare{sigma-cabs}{Sep_18th__11h_43m_4s}{59}{_X_Trip_Distance_Y_Life_Style_Index}{649} \\
    43 & 47 & 51 & 55 & 59\\
    \imagebignosquare{sigma-cabs}{Sep_18th__11h_43m_4s}{91}{_X_Trip_Distance_Y_Life_Style_Index}{649} & \imagebignosquare{sigma-cabs}{Sep_18th__11h_43m_4s}{101}{_X_Trip_Distance_Y_Life_Style_Index}{649} & \imagebignosquare{sigma-cabs}{Sep_18th__11h_43m_4s}{201}{_X_Trip_Distance_Y_Life_Style_Index}{649} & \imagebignosquare{sigma-cabs}{Sep_18th__11h_43m_4s}{301}{_X_Trip_Distance_Y_Life_Style_Index}{649} & \cellcolor{blue!25}{\imagemodelnosquare{sigma-cabs/sigma-cabs-trip-distance-life-style-index}}\\
     91 & 101 & 201 & 301 & \cellcolor{blue!25}{\textbf{target}}
   \end{tabular}
    \caption{Results on UCI \texttt{sigma-cabs} domain (with $m = m' = 5000$) for the 2D plane \texttt{trip-distance} $\times$ \texttt{life-style-index}, convention follows Fig. \ref{fig:exp-randgauss}.}
    \label{fig:exp-sigma-cabs}
  \end{figure}
\egroup

\clearpage

\subsection{Missing data imputation experiments (\textsc{impute})}\label{sec-impute}

\noindent\textbf{Objective} Imputing missing values in data is an important process \cite{mjbcMD}. Classically, imputation methods are \textit{specifically} designed for the task \citep{vFIM}, even when they rely on generative models \citep{ygsGM}. In our case, a general purpose GT $G$ can trivially be used for missing data imputation: we constrain the support of the tree to the observed variables and then sample in the region(s) of maximal density, for a process that takes no more than $\mathcal{O}(|\leafset(G)|)$ per observation. We have tested this simple procedure against the state of the art tree-based methods in the \texttt{mice} R package \cite{vgMM} (and one non-tree based but known to be a good fit for normal data). The methods we have considered in \texttt{mice} all have a commonpoint: they carry out round-robin imputation \citep{mjbcMD}. After having imputed the missing values with an initial guess, they circle round the attributes, repeatedly updating the prevision of one attribute by predicting it from all the current others. Notice the potentially huge number of classifiers used. In our case,  on a domain like \texttt{dna} with 181 variables, the default number of iterations (5) with random forests of 100 trees each means imputing a dataset necessitates no less than \textbf{90 500} trees. In comparison, we rely on a \textbf{single} tree-based model to simultaneously impute all values. In particular on such domains, we cannot hope to beat such approaches, but our approach was rather to compare with SOTA over ranges of problem complexity, variable diversity and have specific simulated domains to further scrutinise differences for GTs used as 'basic' components of imputation methods.\\

\noindent\textbf{Experimental setting} We consider copycat training against a discriminator minimizing Matusita's loss \cite{kmOT}. We grow the GT to a \textsc{max} size (with limit 10 000 nodes, see Section \ref{sec-train-synt}) and prevent splits with $p\in \{0,1\}$, thus avoiding discarding support for data generation. For each domain, we generate data that is Missing Completely At Random (MCAR, \citet{vFIM}) by removing a fixed proportion of modalities $q \in \{5\%, 10\%, 20\%, 50\%\}$, embedded in a 5-fold cross validation for each $q$. In each fold, we thus impute a complete dataset and compare the resulting imputations to the observed values. In such a setting, classical per-observation metrics like RMSE are not necessarily the best choices: if after removing MCAR features two observations were then the same for the resulting features, a perfect imputation of the missing values resulting in a permutation of the observations in the dataset would incur non-zero RMSE, yet would arguably be correct. Similarly to \citet{mjbcMD}, we have thus opted for an optimal transport metric, Wasserstein's $W_2^2$. We use \texttt{mice} with \texttt{method} oracles in $\{$\textsc{cart}, \textsc{norm}, \textsc{random forests (RF)} (100 trees)$\}$. \textsc{norm} is not tree-based but a good alternative on normal data \cite{vFIM}. Notice that we have not used the default number of trees for random forests (10), which we considered too small for our purpose. Due to the difficulty of aggregating different types of variables to compute the Wasserstein distance without accidentally dimming the contribution of some to the total, we consider here only domains for which all variables have the same or closely-related types (\textit{e.g.} categorical with a close number of modalities). We compute our metric using the squared errors normalized to the variable domain for continuous variables, and the error (0/1 loss, multiclass single valued) of prediction for nominal variables (in all cases, the contribution to the distance of each variable is in $[0,1]$).\\

\noindent\textbf{Results} A key part of our experiments was to compare approaches on simulated data since we then know the ground truth, including data with non-trivial structure. Table \ref{tab:imput-vs-mice-gauss} provides all results on our simulated domains. Several conclusions dan be drawn: first, our GTs appear to be winning on at least half of the total domain $\times$ MCAR$\%$ combinations, regardless of the \texttt{mice} contender, even when heavy disparities appear depending on the domain. On \texttt{gridGauss} for example, we become competitive for large $\%$ MCAR (though not statistically significantly) while on \texttt{circGauss} we statistically significantly beat all \texttt{mice} contenders on \textit{all but one run}. We can also notice the quality of the imputations from the plots: \textsc{norm} in \texttt{mice} is clearly failing to impute mostly on the heavy dense regions of the domain. Our results are viauslly much closer to those of \textsc{cart} and \textsc{rf}. We suspect that our method has a different imputation 'quality' pattern vs \textsc{cart} and \textsc{rf}: such methods typically successfully impute near the modes while we can get a more balanced allocation of data \textit{among} modes. On domains like \texttt{circGauss}, we suspect this is the source of our better results. We then have tested what happens when the number of variables increases: to assess this, we used different kind of data (boolean / trinary valued) and domains with an oncreasing number of variables (from 8 to 181). Results are in Table \ref{tab:imput-vs-mice-led}, from which it comes that we are competitive on problems on up to 25 variables and while we are significantly beaten by \texttt{mice} on the largest problem, one has to keep in mind that on \texttt{dna}, we compete on imputation with a single tree model per fold when random forests aggregate \textbf{90 500} of them to do the same task. On may expect that this has impact as well on the time to complete the task. This turns out to be true: on \texttt{dna}, it takes less than 5 minutes to impute a fold with GTs (taking into account the training of the GT), while \texttt{mice} requires more than \textit{two hours} for the same task on random forests. The implementations of our algorithms (Java) and \texttt{mice}'s (R) require caution in comparing times, but we can safely say that on such domains with relatively large number of variables, our approach takes much less time to complete the task. Finally, \texttt{mice} is specifically designed for imputation while our GTs can be used for other purposes than just imputation itself.

\bgroup
\def\arraystretch{1.0}

\begin{table*}
  \centering
\begin{tabular}{cr|lll|cccc}\hline\hline
     \multicolumn{2}{c|}{\textsc{u}s vs \textsc{m}ice$|$} & \tagnorm & \tagcart & \tagrf & \multirow{5}{*}{\includegraphics[trim=60bp 0bp 60bp 0bp,clip,width=0.15\textwidth]{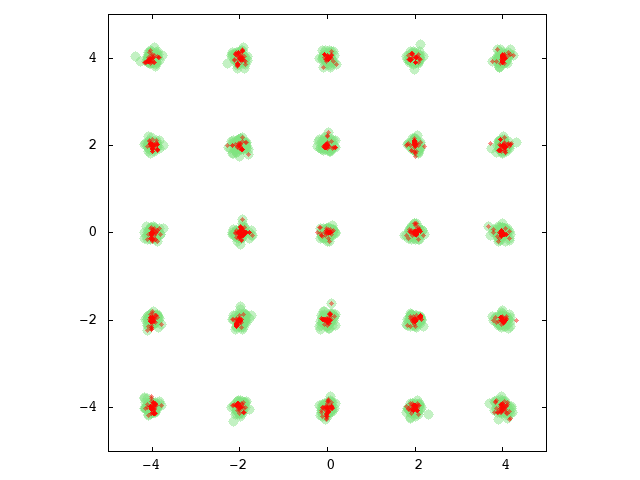}}& \multirow{5}{*}{\includegraphics[trim=60bp 0bp 60bp 0bp,clip,width=0.15\textwidth]{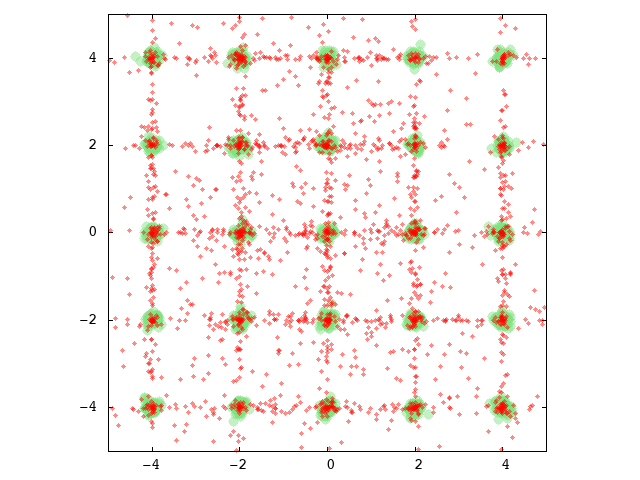}}& \multirow{5}{*}{\includegraphics[trim=60bp 0bp 60bp 0bp,clip,width=0.15\textwidth]{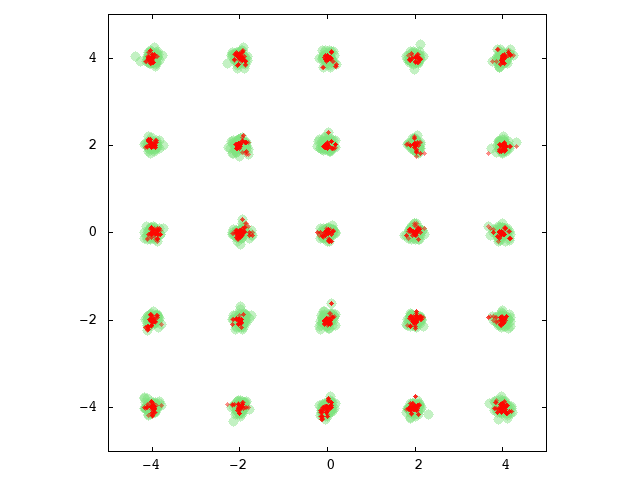}}& \multirow{5}{*}{\includegraphics[trim=60bp 0bp 60bp 0bp,clip,width=0.15\textwidth]{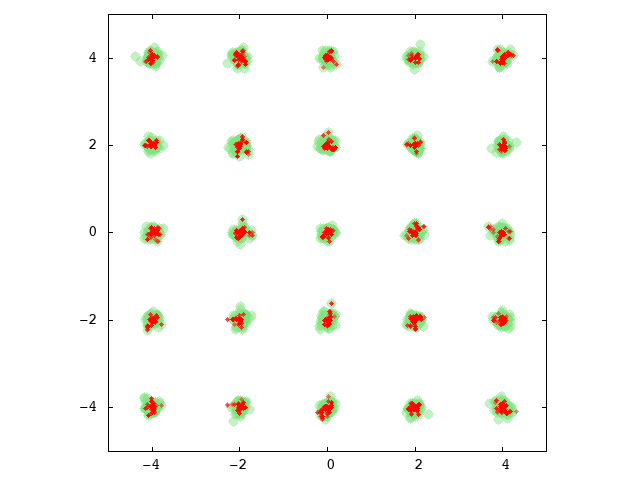}}\\
    \multicolumn{2}{c|}{$\#$trees} & N/A & 10 & 1 000 & & & & \\ \cline{1-5}
    \multirow{4}{*}{\parbox[h]{2mm}{\rotatebox[origin=c]{90}{{\footnotesize \texttt{gridgauss}}}}} & 5$\%$ & \tagthem & \tagthem & \tagthem & & & &  \\
   & 10$\%$ &  \tagthem & \tagthem& \tagthem & & & &  \\
   & 20$\%$ & \tagthem & \tagthem  & \tagthem  & & & &  \\
   & 50$\%$ & \tagus  & \tagus &\tagus & us & mice$|$\tagnorm & mice$|$\tagcart & mice$|$\tagrf\\ \hline\hline
     \multicolumn{2}{c|}{\textsc{u}s vs \textsc{m}ice$|$} & \tagnorm & \tagcart & \tagrf & \multirow{5}{*}{\includegraphics[trim=60bp 0bp 60bp 0bp,clip,width=0.15\textwidth]{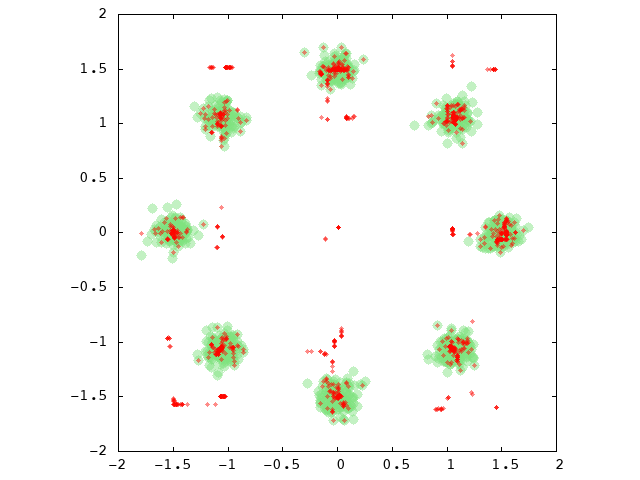}}& \multirow{5}{*}{\includegraphics[trim=60bp 0bp 60bp 0bp,clip,width=0.15\textwidth]{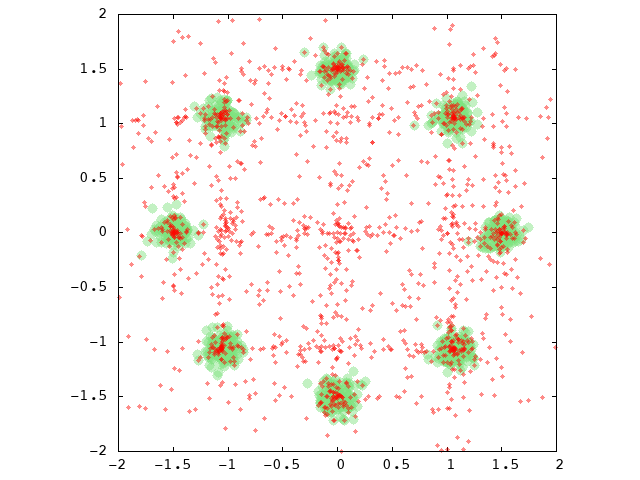}}& \multirow{5}{*}{\includegraphics[trim=60bp 0bp 60bp 0bp,clip,width=0.15\textwidth]{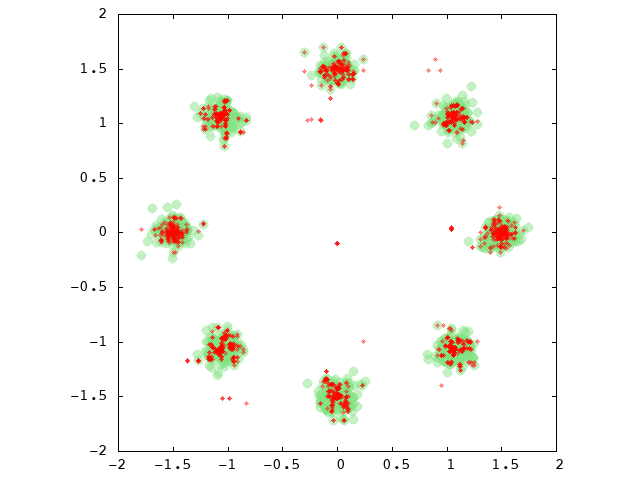}}& \multirow{5}{*}{\includegraphics[trim=60bp 0bp 60bp 0bp,clip,width=0.15\textwidth]{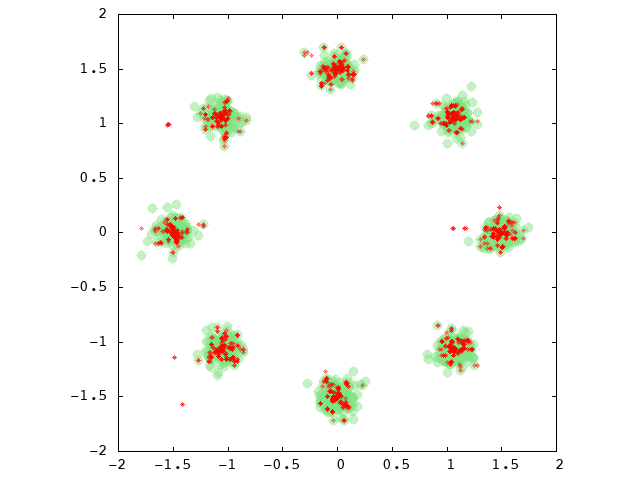}}\\
    \multicolumn{2}{c|}{$\#$trees} & N/A & 10 & 1 000 & & & & \\ \cline{1-5}
    \multirow{4}{*}{\parbox[h]{2mm}{\rotatebox[origin=c]{90}{{\footnotesize \texttt{ringgauss}}}}} & 5$\%$ & \tagus & \signif{\tagus}{0.07} & \tagthem & & & &  \\
   & 10$\%$ &  \tagus & \tagus& \tagthem & & & &  \\
   & 20$\%$ & \tagthem & \tagus  & \tagus  & & & &  \\
   & 50$\%$ & \tagthem  & \tagthem & \signif{\tagthem}{0.08} & us & mice$|$\tagnorm & mice$|$\tagcart & mice$|$\tagrf\\ \hline\hline
     \multicolumn{2}{c|}{\textsc{u}s vs \textsc{m}ice$|$} & \tagnorm & \tagcart & \tagrf & \multirow{5}{*}{\includegraphics[trim=60bp 0bp 60bp 0bp,clip,width=0.15\textwidth]{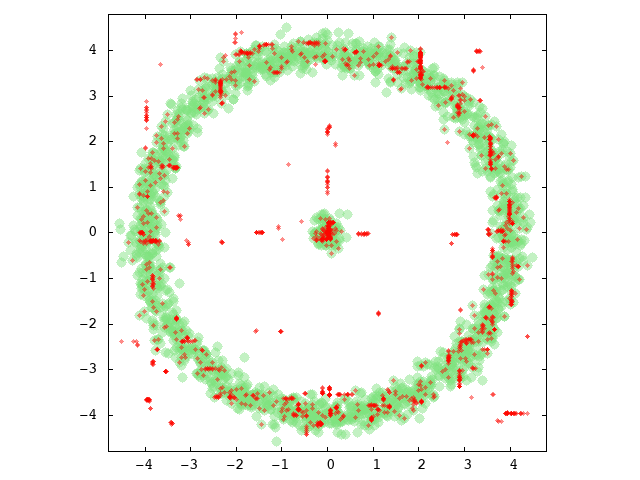}}& \multirow{5}{*}{\includegraphics[trim=60bp 0bp 60bp 0bp,clip,width=0.15\textwidth]{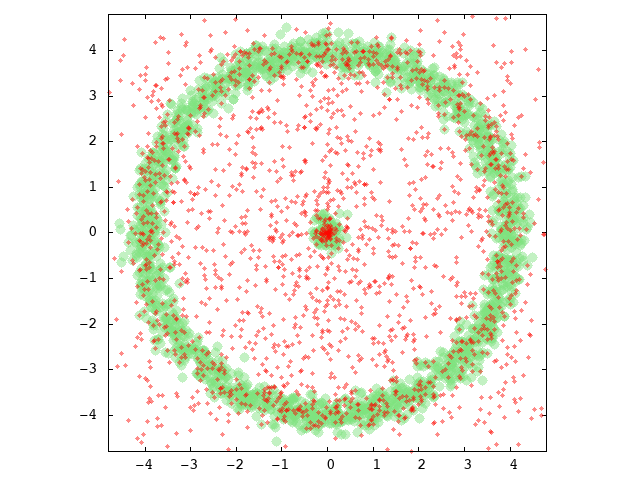}}& \multirow{5}{*}{\includegraphics[trim=60bp 0bp 60bp 0bp,clip,width=0.15\textwidth]{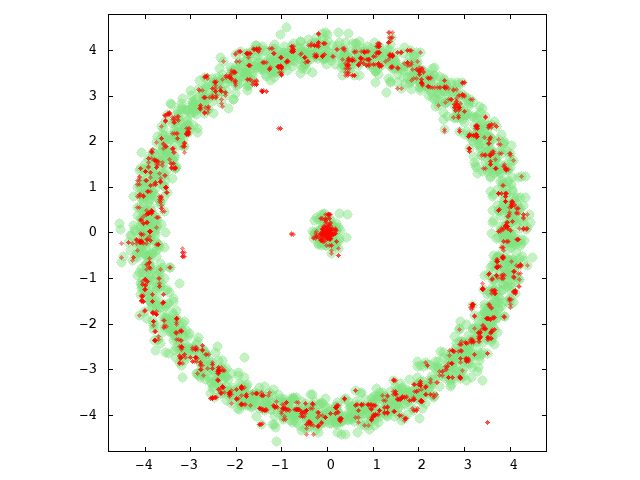}}& \multirow{5}{*}{\includegraphics[trim=60bp 0bp 60bp 0bp,clip,width=0.15\textwidth]{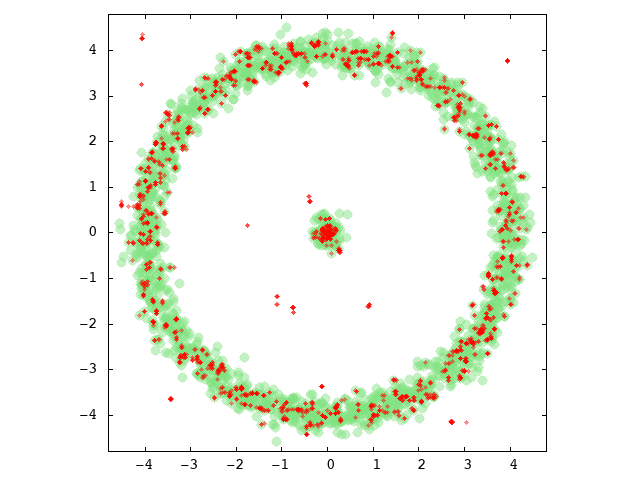}}\\
    \multicolumn{2}{c|}{$\#$trees} & N/A & 10 & 1 000 & & & & \\ \cline{1-5}
    \multirow{4}{*}{\parbox[h]{2mm}{\rotatebox[origin=c]{90}{{\footnotesize \texttt{circgauss}}}}} & 5$\%$ & \signif{\tagus}{0.04} & \tagus & \signif{\tagus}{0.09} & & & &  \\
   & 10$\%$ &  \signif{\tagus}{0.05} & \signif{\tagus}{0.0003} & \signif{\tagus}{0.0007} & & & &  \\
   & 20$\%$ & \signif{\tagus}{0.001} & \signif{\tagus}{0.001} & \signif{\tagus}{0.002}  & & & &  \\
   & 50$\%$ & \signif{\tagus}{0.04}  & \signif{\tagus}{0.05} & \signif{\tagus}{0.05} & us & mice$|$\tagnorm & mice$|$\tagcart & mice$|$\tagrf\\ \hline\hline
     \multicolumn{2}{c|}{\textsc{u}s vs \textsc{m}ice$|$} & \tagnorm & \tagcart & \tagrf & \multirow{5}{*}{\includegraphics[trim=60bp 0bp 60bp 0bp,clip,width=0.15\textwidth]{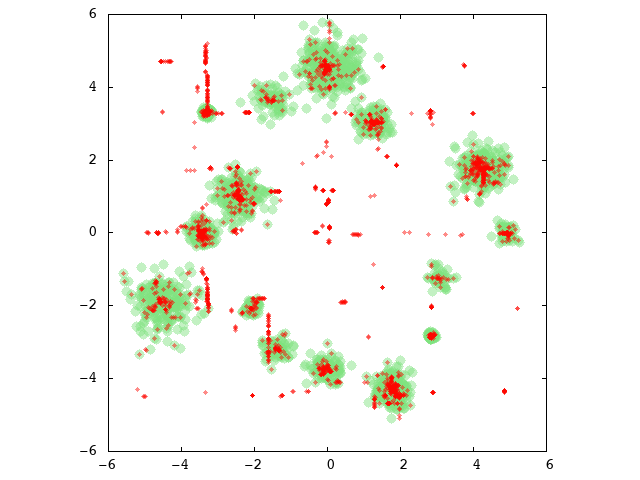}}& \multirow{5}{*}{\includegraphics[trim=60bp 0bp 60bp 0bp,clip,width=0.15\textwidth]{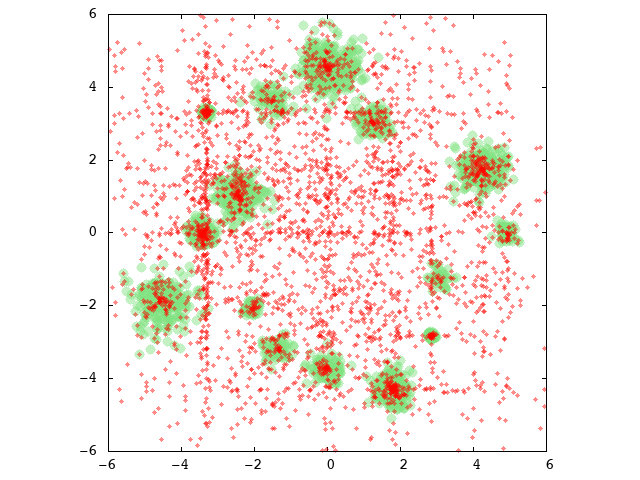}}& \multirow{5}{*}{\includegraphics[trim=60bp 0bp 60bp 0bp,clip,width=0.15\textwidth]{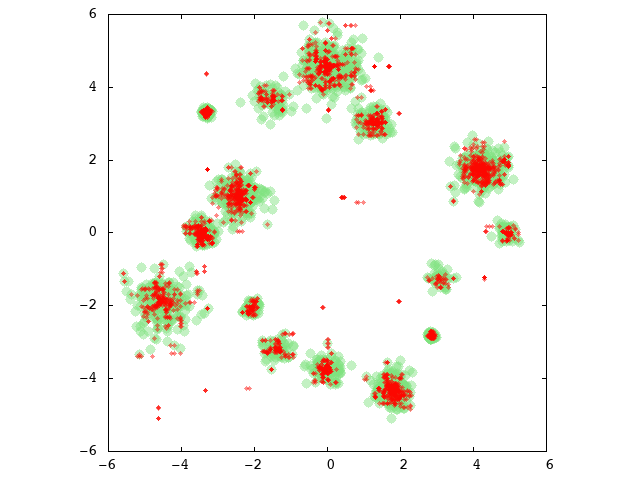}}& \multirow{5}{*}{\includegraphics[trim=60bp 0bp 60bp 0bp,clip,width=0.15\textwidth]{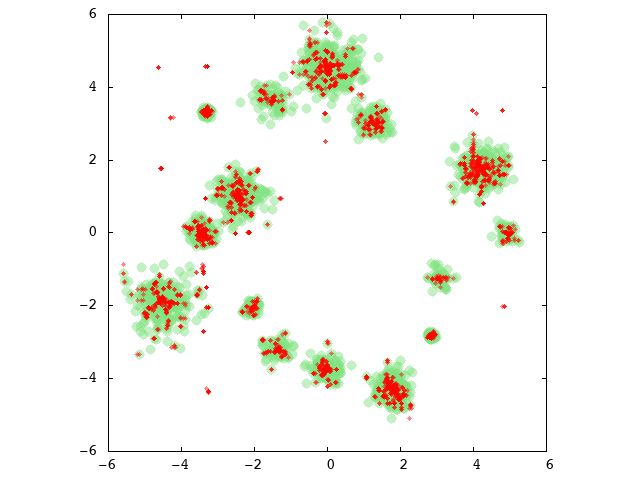}}\\
    \multicolumn{2}{c|}{$\#$trees} & N/A & 10 & 1 000 & & & & \\ \cline{1-5}
    \multirow{4}{*}{\parbox[h]{2mm}{\rotatebox[origin=c]{90}{{\footnotesize \texttt{randgauss}}}}} & 5$\%$ & \tagus & \tagus & \tagus & & & &  \\
   & 10$\%$ &  \tagthem & \signif{\tagus}{0.02} & \tagus & & & &  \\
   & 20$\%$ & \tagthem & \tagthem & \tagthem  & & & &  \\
   & 50$\%$ & \tagus & \tagthem & \tagthem & us & mice$|$\tagnorm & mice$|$\tagcart & mice$|$\tagrf\\ \hline\hline
\end{tabular}
\caption{Experiments \textsc{impute} on simulated domains \texttt{gridgauss}, \texttt{ringgauss}, \texttt{circgauss} and \texttt{randgauss}: comparison of copycat GT induction with maximal size (us) vs mice \citep{vgMM} with three prediction \texttt{method}s: \tagnorm, \tagcart~and \textsc{r}andom \textsc{f}orests. \textit{Left table}: for each $\%$ of missing completely at random (MCAR) variables (in $\{5, 10, 20, 50\}$), we indicate which of us (\tagus) or mice (\tagthem) achieves the lowest average Wasserstein metric $W_2^2$. \textbf{Bold faces} indicate when a Student paired $t$-test returns a $p$-val $<0.1$, in which case the $p$-values is indicated in parenthesis. \textit{Right plots}: examples imputation results (red dots) on top of the domain's data (green) for $50\%$ MCAR.}
    \label{tab:imput-vs-mice-gauss}
  \end{table*}

   \begin{table*}
     \centering
     {\small
\begin{tabular}{cr|ll}\hline\hline
     \multicolumn{2}{c|}{\textsc{u}s vs \textsc{m}ice$|$} & \tagcart & \tagrf \\
    \multicolumn{2}{c|}{$\#$trees} & 40 & 4 000 \\ \hline
    \multirow{4}{*}{\parbox[h]{2mm}{\rotatebox[origin=c]{90}{{\footnotesize \texttt{led}}}}} & 5$\%$ & \signif{\tagus}{0.04} & \tagus \\
   & 10$\%$ &  \signif{\tagus}{0.04} & \tagthem  \\
   & 20$\%$ & \tagus & \tagus  \\
   & 50$\%$ & \tagthem & \tagus  \\ \hline\hline
\end{tabular}
\begin{tabular}{cr|ll}\hline\hline
     \multicolumn{2}{c|}{\textsc{u}s vs \textsc{m}ice$|$} & \tagcart & \tagrf \\
    \multicolumn{2}{c|}{$\#$trees} & 45 & 4 500 \\ \cline{1-4}
    \multirow{4}{*}{\parbox[h]{2mm}{\rotatebox[origin=c]{90}{{\footnotesize \texttt{tictactoe}}}}} & 5$\%$ & \tagus & \signif{\tagthem}{0.07}  \\
   & 10$\%$ &  \tagus  & \tagthem \\
   & 20$\%$ & \tagthem & \signif{\tagthem}{0.05}  \\
   & 50$\%$ & \tagus & \tagthem \\ \hline\hline
\end{tabular}
\begin{tabular}{cr|ll}\hline\hline
     \multicolumn{2}{c|}{\textsc{u}s vs \textsc{m}ice$|$} & \tagcart & \tagrf \\
    \multicolumn{2}{c|}{$\#$trees} & 125 & 12 500 \\ \hline
    \multirow{4}{*}{\parbox[h]{2mm}{\rotatebox[origin=c]{90}{{\footnotesize \texttt{led24}}}}} & 5$\%$ & \tagthem & \tagus \\
   & 10$\%$ &  \tagus & \tagthem \\
   & 20$\%$ & \tagus & \tagthem  \\
   & 50$\%$ & \signif{\tagthem}{0.01} & \signif{\tagthem}{0.07}  \\ \hline\hline
\end{tabular}
\begin{tabular}{cr|ll}\hline\hline
     \multicolumn{2}{c|}{\textsc{u}s vs \textsc{m}ice$|$} & \tagcart & \tagrf \\
    \multicolumn{2}{c|}{$\#$trees} & 905 & 90 500 \\ \hline
    \multirow{4}{*}{\parbox[h]{2mm}{\rotatebox[origin=c]{90}{{\footnotesize \texttt{dna}}}}} & 5$\%$ & \signif{\tagthem}{0.03} & \signif{\tagthem}{0.02} \\
   & 10$\%$ &  \signif{\tagthem}{0.02} & \signif{\tagthem}{0.004} \\
   & 20$\%$ & \signif{\tagthem}{0.001} & \signif{\tagthem}{0.002}  \\
   & 50$\%$ & \signif{\tagthem}{0.002} & \signif{\tagthem}{0.001}  \\ \hline\hline
\end{tabular}
}
\caption{Experiments \textsc{impute} on binary/trinary valued domains \texttt{led}, \texttt{tictactoe}, \texttt{led24}, \texttt{dna}, with increasing number of nominal description variables. Notations follow Table \ref{tab:imput-vs-mice-gauss} (\tagnorm~not shown as it does not impute all NAs). While we are competitive on domains with the smallest number of variables, we are clearly beaten by \texttt{mice} when the number of variables substantially increases like for \texttt{dna}. Those numbers have to be read keeping in mind the number of trees involved in imputing a single fold: while we compete with 1 tree against 40 (\textsc{cart}) and 4 000 (\textsc{rf}) on \texttt{led}, we compete with \textbf{1} tree against \textbf{905} (\textsc{cart}) and \textbf{90 500} (\textsc{rf}) on \texttt{dna}.}
    \label{tab:imput-vs-mice-led}
  \end{table*}

 \egroup

\clearpage

\subsection{'Training on synthetic' experiment (\textsc{train-synth})}\label{sec-train-synt}

\bgroup
\def\arraystretch{1.5}
  \begin{table*}
  \centering
  \begin{tabular}{c|cccccccc}\hline\hline
    Domain & $\#$1 & $\#$2& $\#$3& $\#$4& $\#$5& $\#$6& $\#$7& $\#$8\\ \hline
  \texttt{abalone} & \tagcop & \taggt{\textsc{max}} & \taggt{300} & \tagctgan{1K} & \tagctgan{300} & \tagctgan{10} & \taggt{10} & \taguni \\     \texttt{dna} & \tagcop & \taggt{300} & \taggt{\textsc{max}} & \taguni & \tagctgan{10} & \taggt{10} & \tagctgan{1K} & \tagctgan{300} \\   \texttt{house votes} & \tagcop & \taggt{300} & \taggt{\textsc{max}} & \tagctgan{1K} & \taggt{10}  & \tagctgan{10} & \tagctgan{300} & \taguni \\   \texttt{iris} & \tagcop & \tagctgan{1K} & \taggt{300} & \taggt{\textsc{max}} & \tagctgan{10} & \tagctgan{300} & \taggt{10}  & \taguni \\   \texttt{led24} & \tagcop & \taggt{300} & \taggt{\textsc{max}} & \tagctgan{10} & \taggt{10}  & \tagctgan{1K} & \taguni & \tagctgan{300} \\   \texttt{led} & \tagcop & \taggt{300} & \taggt{\textsc{max}} & \tagctgan{1K} & \taggt{10}  & \tagctgan{10} & \tagctgan{300} & \taguni \\
  \texttt{winered} & \tagcop & \taggt{300} & \taggt{\textsc{max}} & \tagctgan{1K} & \taguni & \taggt{10}  & \tagctgan{300} & \tagctgan{10} \\
  \texttt{winewhite} & \tagcop & \taggt{\textsc{max}} & \taggt{300} & \taggt{10}  & \taguni & \tagctgan{300} & \tagctgan{10} & \tagctgan{1K} \\
  \texttt{sigma-cabs} & \tagcop & \tagctgan{1K} & \tagctgan{300} & \tagctgan{10}  & \taggt{\textsc{max}} & \taggt{300} & \taguni & \taggt{10}  \\
  \texttt{open-policing} & \tagcop & \taggt{\textsc{max}} & \tagctgan{1K} & \taggt{300} & \tagctgan{300} & \tagctgan{10} & \taggt{10}   & \taguni \\
  \hline\hline
\end{tabular}
\caption{Ranking results on experiment \textsc{train-synth}, showing for each domain the order (left to right: from best to worst) of the three sizes of runs of our GTs (\taggt{.}, number in parenthesis = number of splits; \textsc{max} = up to 10 000 splits), the three runs of \ctgan~with different epoch numbers (\tagctgan{.}), the \textsc{copy} approach (we use the original data) and the \textsc{unif}(orm) approach (we use a random sample). Those two last methods have their cells shaded to locate them.}
    \label{tab:train-synth-1}
  \end{table*}

  \begin{table*}
  \centering
\begin{tabular}{cccccccc}\hline\hline
 \tagcop & \taggt{300} & \taggt{\textsc{max}} & \tagctgan{1K} & \tagctgan{10} & \taggt{10} & \tagctgan{300} & \taguni\\
 1 & 2.9 & 3 & 4.5 & 5.7 & 6 & 6.2 & 6.8 \\  \hline\hline
\end{tabular}
\caption{Average rank for each approach in the \textsc{train-synth}, as collected in Table \ref{tab:train-synth-1}, ordered in increasing average rank.}
    \label{tab:train-synth-2}
  \end{table*}
  
\begin{table*}
  \centering
\begin{tabular}{c|c|c|c}\hline\hline
  \backslashbox{\taggt{.}}{\tagctgan{.}} & 10{\color{red}$\star$} & 300{\color{red}$\star$} & 1K{\color{red}$\star$}{\color{red}$\star$} \\  \hline\hline
  10 & \wtl{1}{7}{2} & \wtl{3}{4}{3} & \wtl{2}{3}{5}\\ \hline
  300 & \wtl{8}{1}{1} & \wtl{8}{1}{1} & \wtl{4}{5}{1}\\ \hline
  \textsc{max} & \wtl{8}{1}{1} & \wtl{8}{1}{1} & \wtl{7}{2}{1} \\ \hline\hline
\end{tabular}
\caption{Experiment \textsc{train-synth}: statistical wins / ties / statistical losses for us (\taggt{})~vs CT-GAN (\tagctgan{}). Statistical = significant for $p \leq 0.01$. For example, \wtl{$a$}{$b$}{$c$} means we statistically win $a$ times, lose $c$ times and there is no statistical difference $b$ times. Each red star ({\color{red}$\star$}) indicates a domain for which the related technique performed statistically \textit{worse} than uniform sampling (\textsc{unif}) for $p=0.05$ (See Table \ref{tab:train-synth-1} to spot those domains).}
    \label{tab:train-synth-3}
  \end{table*}

\begin{table*}
  \centering
  \begin{tabular}{c|rrr|rrr}\hline\hline
    Domain & \taggt{10} & \taggt{300} & \taggt{\textsc{max}} & \tagctgan{10} & \tagctgan{300} & \tagctgan{1K}\\ \hline
    \texttt{abalone} & 23 & 90 & 193 & 14 & 62 & 215\\
    \texttt{dna} & 2 & 7 & 38 & 143 & 1 051 & 3 128\\
    \texttt{house-votes} & $\epsilon$ & $\epsilon$ & $\epsilon$ & 7 & 20 & 51\\
    \texttt{iris} & $\epsilon$ & $\epsilon$ & $\epsilon$ & 7 & 17 & 42\\
    \texttt{led24}& $\epsilon$ & 1 & 2 & 6 & 22 & 62 \\
    \texttt{led}& $\epsilon$ & 1 & 1 & 6 & 14 & 46\\
  \texttt{winered} & 3 & 8 & 14 & 13 & 24 & 62\\
  \texttt{winewhite} & 17 & 46 & 131 & 17 & 52 & 210 \\
  \texttt{sigma-cabs} & 19 & 60 & 286 & 13 & 86 & 419 \\
  \texttt{open-policing} & 168 & 311 & 933 & 24 & 338 & 1 316\\
  \hline\hline
\end{tabular}
\caption{Average training times to get the generated training sample on experiment \textsc{train-synth}, in seconds, rounded to the nearest second. '$\epsilon$' means average $<0.5$s.}
    \label{tab:train-synth-4}
  \end{table*}

\begin{table*}
  \centering
  \begin{tabular}{c|rrr|rrr}\hline\hline
    Domain & $|\mbox{GT}|/|\mathcal{X}|\cdot 100$ ($\%$) \\ \hline
    \texttt{abalone} &  3.99 \\
    \texttt{dna} &  0.26 \\
    \texttt{house-votes} & 21.57\\
    \texttt{iris} &  200.13\\
    \texttt{led24}&  6.00\\
    \texttt{led}&  18.76\\
  \texttt{winered} & 7.82\\
  \texttt{winewhite} &  2.55\\
  \texttt{sigma-cabs} &  2.31 \\
  \texttt{open-policing} &  0.41\\
  \hline\hline
\end{tabular}
\caption{Experiment \textsc{train-synth}: average sizes of the GT obtained using \taggt{300} relative to the domain size (see text).}
    \label{tab:train-synth-5}
  \end{table*}
  
  \egroup

  \bgroup
\def\arraystretch{0.2}
\begin{figure}
  \centering
  \begin{tabular}{c||ccc||ccc}
    ground truth & \taggt{10} & \taggt{300} & \taggt{\textsc{max}} & \tagctgan{10} & \tagctgan{300} & \tagctgan{1K}\\\hline
    \includegraphics[trim=0bp 0bp 2600bp 0bp,clip,width=0.14\textwidth]{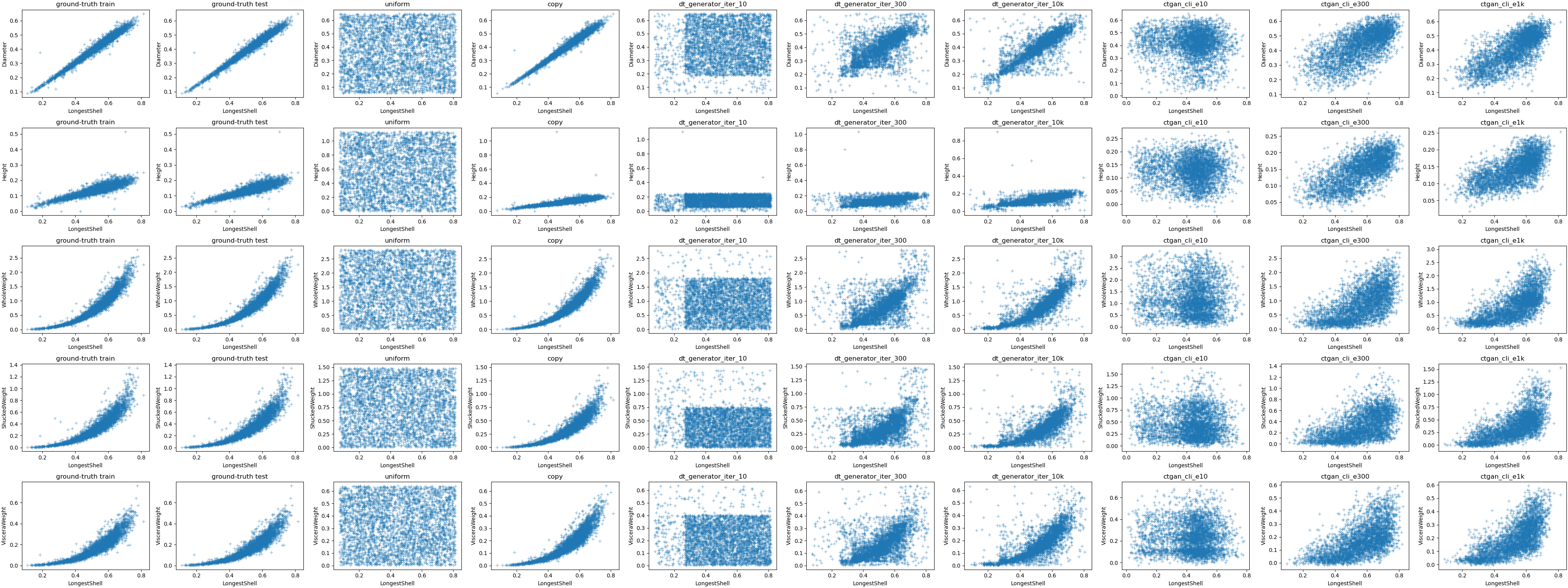} & \includegraphics[trim=1150bp 0bp 1450bp 0bp,clip,width=0.14\textwidth]{Figs/experiments/abalone/abalone_distributions.png} & \includegraphics[trim=1440bp 0bp 1160bp 0bp,clip,width=0.14\textwidth]{Figs/experiments/abalone/abalone_distributions.png} & \includegraphics[trim=1730bp 0bp 870bp 0bp,clip,width=0.14\textwidth]{Figs/experiments/abalone/abalone_distributions.png} & \includegraphics[trim=2020bp 0bp 580bp 0bp,clip,width=0.14\textwidth]{Figs/experiments/abalone/abalone_distributions.png} & \includegraphics[trim=2310bp 0bp 290bp 0bp,clip,width=0.14\textwidth]{Figs/experiments/abalone/abalone_distributions.png} & \includegraphics[trim=2600bp 0bp 0bp 0bp,clip,width=0.14\textwidth]{Figs/experiments/abalone/abalone_distributions.png} \\\hline\hline
   \end{tabular}
    \caption{Experiment \textsc{train-synth}: 2D distribution plots for domain \texttt{abalone} (warning: scales vary).}
    \label{fig:train-synth-1}
  \end{figure}
  \egroup
  
  \bgroup
\def\arraystretch{0.2}
\begin{figure}
  \centering
  \begin{tabular}{c||ccc||ccc}
    ground truth & \taggt{10} & \taggt{300} & \taggt{\textsc{max}} & \tagctgan{10} & \tagctgan{300} & \tagctgan{1K}\\\hline
    \includegraphics[trim=0bp 0bp 2600bp 0bp,clip,width=0.14\textwidth]{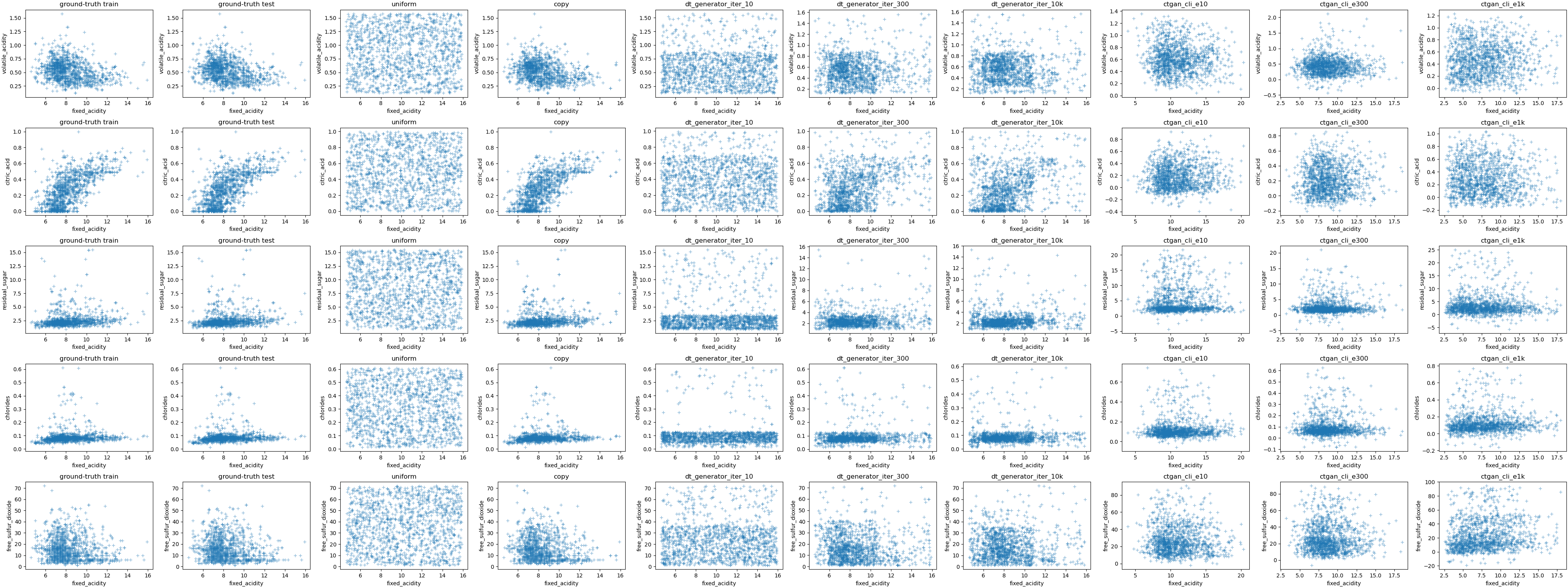} & \includegraphics[trim=1150bp 0bp 1450bp 0bp,clip,width=0.14\textwidth]{Figs/experiments/winered/winered_distributions.png} & \includegraphics[trim=1440bp 0bp 1160bp 0bp,clip,width=0.14\textwidth]{Figs/experiments/winered/winered_distributions.png} & \includegraphics[trim=1730bp 0bp 870bp 0bp,clip,width=0.14\textwidth]{Figs/experiments/winered/winered_distributions.png} & \includegraphics[trim=2020bp 0bp 580bp 0bp,clip,width=0.14\textwidth]{Figs/experiments/winered/winered_distributions.png} & \includegraphics[trim=2310bp 0bp 290bp 0bp,clip,width=0.14\textwidth]{Figs/experiments/winered/winered_distributions.png} & \includegraphics[trim=2600bp 0bp 0bp 0bp,clip,width=0.14\textwidth]{Figs/experiments/winered/winered_distributions.png} \\\hline\hline
   \end{tabular}
    \caption{Experiment \textsc{train-synth}: 2D distribution plots for domain \texttt{winered} (warning: scales vary). Notice that GTs manage to capture complex distribution shapes (such as the 'wavy' 2nd row) that neural nets do not necessarily capture, and neural networks can generate data clearly outside admissible bounds (negative values for some variables)}
    \label{fig:train-synth-2}
  \end{figure}
  \egroup

\noindent\textbf{Objective} In the context of tabular data, there can be several reasons to replace a training sample with a data generator: (i) the storing size of the generated model, in particular for big databases and / or with substantial redundancy \citep{rcDS}, (ii) privacy (disregarding the privacy model), (iii) security (with respect to classical databases), etc. . The objective of this experiment is the replacement of the training data by a data generator \textit{to address the supervised learning problem} related to the training data. For example, for domain \texttt{iris}, the objective is the prediction of a flower's variety among three.\\

\noindent\textbf{Experimental setting} For each domain, we carry out a 5-fold CV, leaving 20$\%$ of the data for testing and the rest for training. We train a data generator with the training data, then generate a training sample the size of the generator's training sample. This training sample is then used \textit{in lieu of} the original training sample to train a model following the original data's task. The model is then used to predict the class of the testing fold's examples. To alleviate the bias on the choice of this classifier and its training algorithm, we pick two types of classifiers / training models: random forests (RFs) and gradient boosted decision trees (GBDTs), that we thus run on each dataset. We compute the accuracy for label prediction and root mean square error for regression problems. For each domain and each generator type, we thus obtain 10 statistics (5 for RFs, 5 for GBDTs), to which we add the 10 running times and use them for comparisons between methods. Speaking methods, we consider for generative trees (GTs) copycat training against a discriminator minimizing Matusita's loss \cite{kmOT}. We grow generative trees with three different sizes: very small (10 splits, \textit{i.e.} 21 total nodes), medium (300 splits, \textit{i.e.} 601 total nodes) and 'maximal'. In this last case, we provide a limit size of 10 000 splits (\textit{i.e.} 20 001 total nodes). Notice that this maximal size is usually not reached, in particular when a training fold contains less than 10 000 training examples. We compare our three GT training flavours to three training neural network based methods relying on the state of the art CT-GANs \cite{xscvMT}. We use CT-GANs code\footnote{\url{https://github.com/sdv-dev/CTGAN}} with default parameters and varying number of epochs, choosing small (10), medium (300) and large (1K) training epochs. We also use two more contenders: the first is the \textsc{copy} contender, which uses the original training data as training data. The second is the \textsc{unif}orm contender, which just generates uniform data in the features domains. While we expect \textsc{copy} to lead the pack of algorithms, contender \textsc{unif} is used to point algorithms failing at learning anything 'substantial' about the domain when it significantly beats them. \\

\noindent\textbf{Results} Table \ref{tab:train-synth-1} provides the \textit{ranking} results among all eight contenders for each domain, where the average of the metric (accuracy or RMSE) is used to rank from best to worst. Table \ref{tab:train-synth-2} provides a more synthetic view by computing the average rank for each contender. There are several conclusions to draw: (i) as perhaps expected, \textsc{copy} is always the best approach; what is perhaps less expected is that (ii) \textsc{unif} is far from always being the worst, CT-GANs being the most frequently beaten contenders (albeit not necessarily statistically significantly, see below) on four out of ten domains. More importantly, (iii) our approach largely performs the best in all non-\textsc{copy} approaches \textit{unless} small GTs are used (\taggt{10}). In eight out of ten domains, \taggt{300} is in the top-three contenders, and top-two if we exclude \textsc{copy}. An interesting observation is that \taggt{\textsc{max}} is \textit{also} in the top-three contenders in eight out of ten domains, showing that there is reduced overfitting effect due to the large tree size (which we attribute in part to the constraint that $p \not\in \{0,1\}$ for GT splits). Obviously, in a context where explainability would be key, the smaller size option (\taggt{300}) would be a preferred choice. To dig further in comparing our method to CT-GAN, we have computed the number of domains one \textit{significantly} ($p=0.01$) beats the other, adding to those statistics the number of times \textsc{unif} does \textit{significantly} ($p=0.05$) beat some contender(s). All results are summarized in Table \ref{tab:train-synth-3}. The results display the superiority of our GT-based approach (unless, again small trees are used), but they also display that CT-GANs are, in few cases, significantly beaten by \textsc{unif} -- and this can happen for all three epoch numbers. This aligns with the observation that dealing with tabular data with neural nets forces sophisticated choices for the design \citep{xscvMT} and probably has as consequence that not optimizing sufficiently hyperparameters can result in worse performances than uniform generation. We clearly do not have this problem and believe that this is due to the fact that the tree (graphs) used in GTs bring the same convenience for data generation as the tree (graphs) used in decision trees have for discrimination. Figures \ref{fig:train-synth-1} and \ref{fig:train-synth-2} provide two examples of sets of 2D plots showing the distribution of generated examples according to different sets of couples of real-values variables. The absence of overfitting as well as the capturing of sophisticated features of the data's distribution is quite apparent, also in comparison with neural nets. Last, we have computed the average computation time to get the generated datasets for us and CT-GAN -- thus, inclusive of the training time for the generator. Results are provided in Table \ref{tab:train-synth-4}. One must be cautious in comparing numbers as our implementation of our algorithms is in Java while CT-GAN's is in Python, but at least one conclusion seems fair to draw: we are in general -- and especially for bigger models / longer training -- achieving much better results than CT-GAN. The \texttt{dna} domain, for which the imputation experiments were already displaying our superiority in terms of training time (Section \ref{secexp}), is a clear example of reduction in training time that can be of order $10\times$ --- $100\times$ with GTs compared to neural networks.

To put our results in perspective, we have also computed the relative size of the GT learned with respect to each domain size, for each experiment: we use as GT size $|\mbox{GT}|$ the total number of vertices and arcs, which equals ($1+5 \cdot$split number) and as domain size, $|\mathcal{X}|$, the total size of the dataset used (number of examples times number of variables). Table \ref{tab:train-synth-5} presents the results obtained for \taggt{300}, from which it appears that the GT learned can represent a tiny proportion of a domain's size --- at most a few percents in most cases.

\clearpage

\subsection{'Synthetic discrimination' experiment (\textsc{synth-discrim})}\label{sec-synt-disc}

\bgroup
\def\arraystretch{1.5}
\begin{figure*}
  \centering
  \includegraphics[trim=10bp 30bp 500bp 0bp,clip,width=0.55\textwidth]{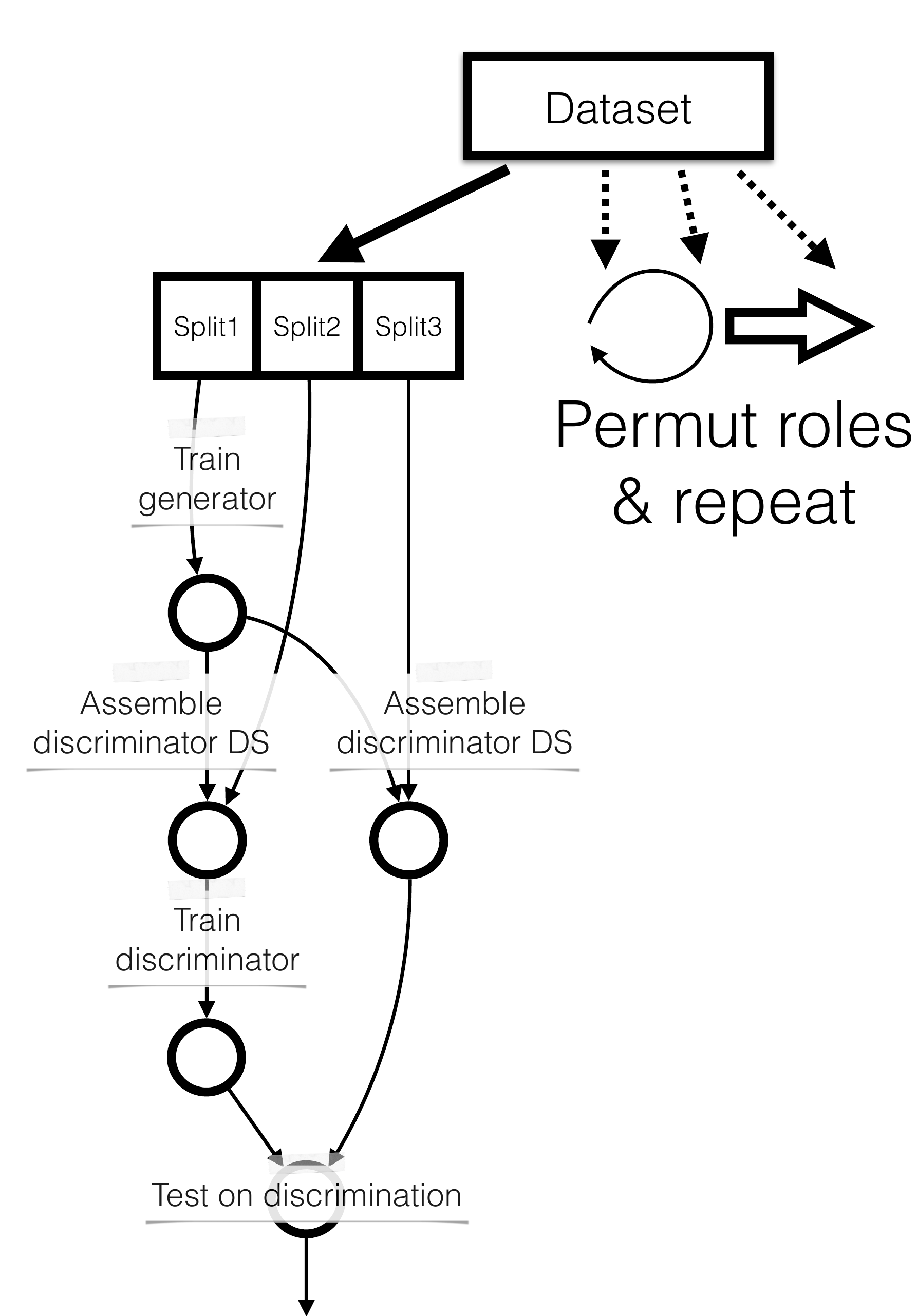}
\caption{Experiment \textsc{synth-discrim}: General overview of the pipeline, designed to avoid rewarding generators that would just copy their training sample.}
    \label{fig:synt-disc-1}
  \end{figure*}

\begin{table*}
  \centering
  {\footnotesize
  \begin{tabular}{c|l|l}\hline\hline
    Domain & \textbf{wins} vs \textsc{copy} & \textbf{loses} vs \textsc{unif} \\ \hline
    \texttt{circgauss} &  \taggt{300} \pval{0.01}, \taggt{\textsc{max}} \pval{0.02} & None \\
    \texttt{randgauss} &  \taggt{300} \pval{0.00004}, \taggt{\textsc{max}} \pval{0.00002} & \tagctgan{10} \pval{0.01} \\
    \texttt{ringgauss} &  \taggt{300} \pval{0.002}, \taggt{\textsc{max}} \pval{0.001} & None \\
    \texttt{abalone} &  None & \tagctgan{10} \pval{0.0006}  \\
    \texttt{dna} &  None & None \\
    \texttt{house-votes} &  None & \tagctgan{10} \pval{0.0002}\\
    \texttt{iris} &  None & \tagctgan{10} \pval{0.007}, \tagctgan{300} \pval{0.0002} \\
    \texttt{led24} &  None & None \\
    \texttt{led} &  None & \tagctgan{1K} \pval{0.03} \\
    \texttt{winered} &  None & \taggt{300} \pval{0.003}, \taggt{\textsc{max}} \pval{0.001}\\
    \texttt{winewhite} &  None & \taggt{300} \pval{0.006}, \taggt{\textsc{max}} \pval{0.001}\\
    \texttt{sigma cabs} &  None & \taggt{10} \pval{0.001}, \taggt{300} \pval{0.001}, \taggt{\textsc{max}} \pval{0.001}, \tagctgan{10} \pval{0.001}, \tagctgan{300} \pval{0.003}\\
    \texttt{open-policing} &  None & None \\
   \hline\hline
  \end{tabular}
  }
\caption{Experiment \textsc{synth-discrim}: for each domain, we compute the list of contenders in our method and CT-GAN statistically \textit{winning} against \textsc{copy} and statistically \textit{losing} against \textsc{unif} (for $p \leq 0.05$, indicated in parenthesis).}
    \label{tab:synt-disc-2}
  \end{table*}
  
  \begin{table*}
  \centering
\begin{tabular}{c|c|c|c}\hline\hline
  \backslashbox{\taggt{.}}{\tagctgan{.}} & 10 & 300  & 1K  \\  \hline\hline
  10 & \wtl{7}{4}{2} & \wtl{6}{5}{2} & \wtl{6}{5}{2}\\ \hline
  300 & \wtl{8}{4}{1} & \wtl{8}{3}{2} & \wtl{8}{3}{2}\\ \hline
  \textsc{max} & \wtl{8}{4}{1} & \wtl{8}{3}{2} & \wtl{8}{3}{2} \\ \hline\hline
\end{tabular}
\caption{Experiment \textsc{synth-discrim}: statistical wins / ties / statistical losses for us (\taggt{})~vs CT-GAN (\tagctgan{}). Statistical = significant for $p \leq 0.01$. For example, \wtl{$a$}{$b$}{$c$} means we statistically win $a$ times, lose $c$ times and there is no statistical difference $b$ times. (See Table \ref{tab:synt-disc-2} to spot those domains).}
    \label{tab:synt-disc-3}
  \end{table*}
  \egroup

\clearpage

\noindent\textbf{Objective} The objective fits in a simple question: \textit{can generated examples look like real ones} ?\\

\noindent\textbf{Experimental setting} The question is simple but its treatment non trivial: we need in particular to avoid 'rewarding' generators that would just copy their training examples. Figure \ref{fig:synt-disc-1} provides the training pipeline we have designed, that we ran for each domain considered. In short, we split each domain in three equal sized parts, say ${\mathcal{S}}_1$, ${\mathcal{S}}_2$ and ${\mathcal{S}}_3$. One of these parts, say ${\mathcal{S}}_1$, is used to trained the generator, which then generates a sample $\tilde{\mathcal{S}}_1$ having the same size as ${\mathcal{S}}_1$. We then train a discriminator for the 2-classes supervised learning problem consisting in distinguishing real from fake, using $\tilde{\mathcal{S}}_1$ and another original part, say ${\mathcal{S}}_2$ as training samples. The discriminator is then \textit{tested} on the problem consisting in distinguishing $\tilde{\mathcal{S}}_1$ from the last original part (not yet used), ${\mathcal{S}}_3$ in this case. The smaller the final accuracy, the more 'realistic' is considered $\tilde{\mathcal{S}}_1$. We then permut the roles of the three samples and run the experiment again, ending up in $3! = 6$ accuracies for each domain. Considering that the discriminator is trained and tested on two different subsets of the original data as real data, there is an incentive to not just 'copy' the original data but capture features about the domain that generalise well for data generation. We have considered the same generators as in experiment 'textsc{train-synth}': CT-GANs with small (10), medium (300) and large (1K) number of training epochs; our method with 10, 300 and \textsc{max} splits for GTs (recall that \textsc{max} = training up to 10 000 splits); we also consider the \textsc{copy} and \textsc{unif}orm baselines, the former giving an idea of the accuracy for the original data and the latter giving the most 'blunt' baseline. We consider the same discriminators as in experiment '\textsc{train-synth}' (random forests and gradient boosted decision trees).\\

\noindent\textbf{Results} We first have a look at the \textit{extreme} results, \textit{i.e.} how our method and CT-GANs compare to \textsc{copy} and \textsc{unif}. Table \ref{tab:synt-disc-2} presents the detailed results obtained for each domain. In this table, we only look at statistically significant results --- for example, when the accuracies on testing were statistically significantly \textit{larger} than \textsc{unif} (which means that \textsc{unif} performed better), or when the accuracies on testing were statistically significantly \textit{smaller} than \textsc{copy} (which means that \textsc{copy} performed worse). The picture with respect to \textsc{unif} displays that some CT-GANs (disregarding the number of epochs) get worse results on almost half of the domains (6) while some of our methods gets worse results on 3 of them. When looking at \textsc{copy}, we see that on our simulated domains, our method actually gets \textit{systematically} significantly better results than \textsc{copy} when the number of splits is at least 300. This, we believe, signals the potential for GTs to be used as efficient data generators, eventually as parts of more complex generators for more complex domains than our generated domains. We have also drilled down in the comparison between our approach and CT-GANs in the same way as we did for experiment \textsc{train-synth}. Table \ref{tab:synt-disc-3} presents the aggregated results, whose formatting follows the same rules as for Table \ref{tab:train-synth-3}. The conclusion from this Table is that our approach does better at producing 'realistically looking' datasets than CT-GAN does. When the GTs are big enough (at least 300 splits), the picture displays that our approach wins against all CT-GANs alternatives on a large majority of domains.

\subsection{'Synthetic augmentation' experiment (\textsc{synth-aug})}\label{sec-synt-aug}

\bgroup
\def\arraystretch{1.5}
  \begin{table*}
  \centering
\begin{tabular}{c|c|c|c}\hline\hline
  \backslashbox{\taggt{.}}{\tagctgan{.}} & 10 & 300  & 1K  \\  \hline\hline
  10 & \wtl{6}{1}{3} & \wtl{6}{2}{2} & \wtl{4}{1}{5}\\ \hline
  300 & \wtl{9}{0}{1} & \wtl{9}{0}{1} & \wtl{6}{2}{2}\\ \hline
  \textsc{max} & \wtl{9}{0}{1} & \wtl{9}{0}{1} & \wtl{7}{1}{2} \\ \hline\hline
\end{tabular}
\caption{Experiment \textsc{synth-aug}: statistical wins / ties / statistical losses for us (\taggt{})~vs CT-GAN (\tagctgan{}). Statistical = significant for $p \leq 0.01$. For example, \wtl{$a$}{$b$}{$c$} means we statistically win $a$ times, lose $c$ times and there is no statistical difference $b$ times. }
    \label{tab:synt-aug-1}
  \end{table*}
  \egroup

\clearpage

\bgroup
\def\arraystretch{1.0}

\begin{table*}
  \centering
\begin{tabular}{c}\hline\hline
  \includegraphics[trim=0bp 0bp 0bp 300bp,clip,width=0.99\textwidth]{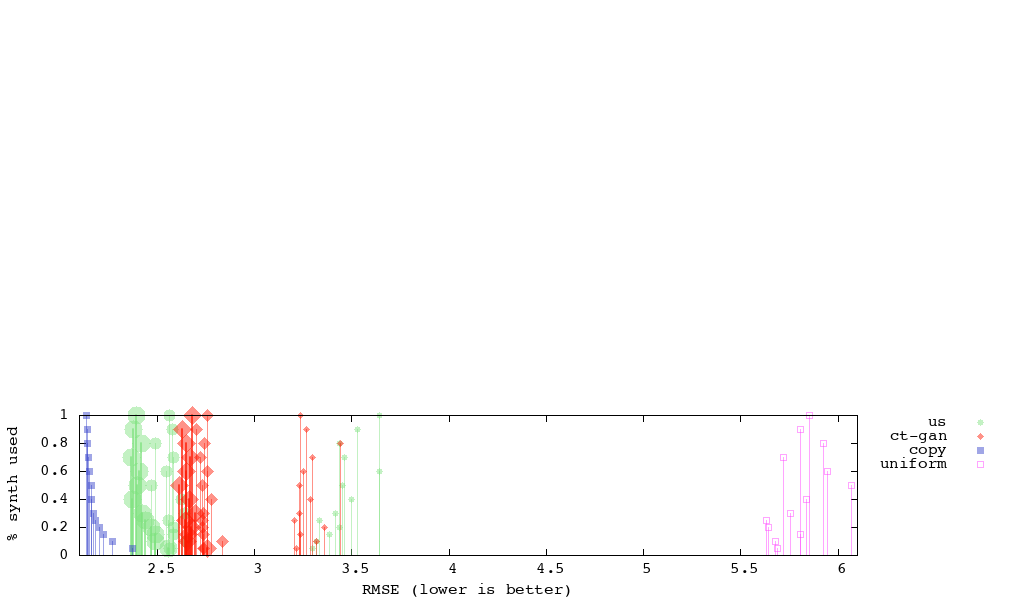} \\
  \texttt{abalone}\\ \hline
  \includegraphics[trim=0bp 0bp 0bp 300bp,clip,width=0.99\textwidth]{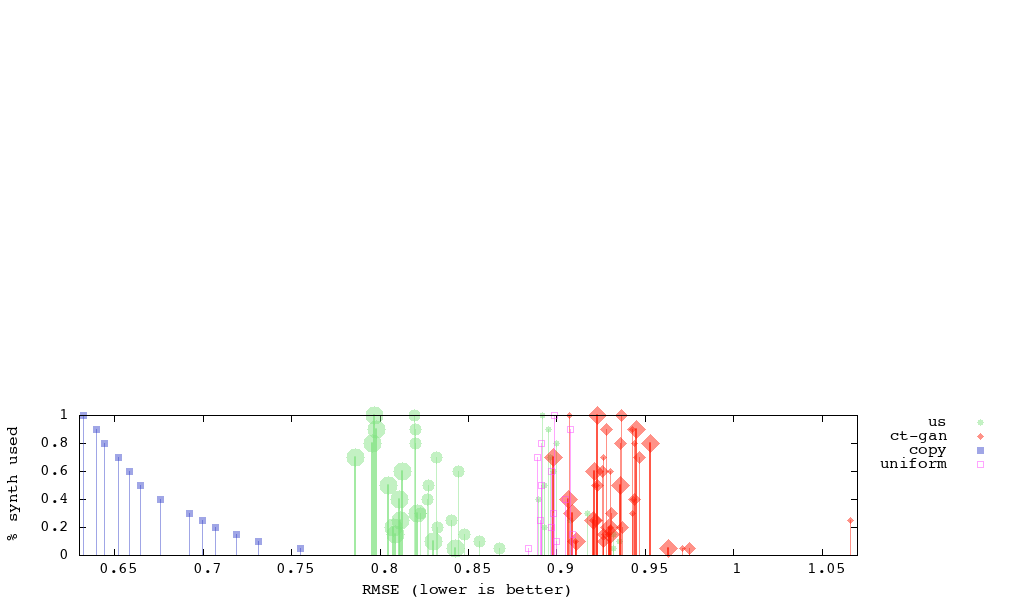} \\
  \texttt{winewhite}\\ \hline
  \includegraphics[trim=0bp 0bp 0bp 300bp,clip,width=0.99\textwidth]{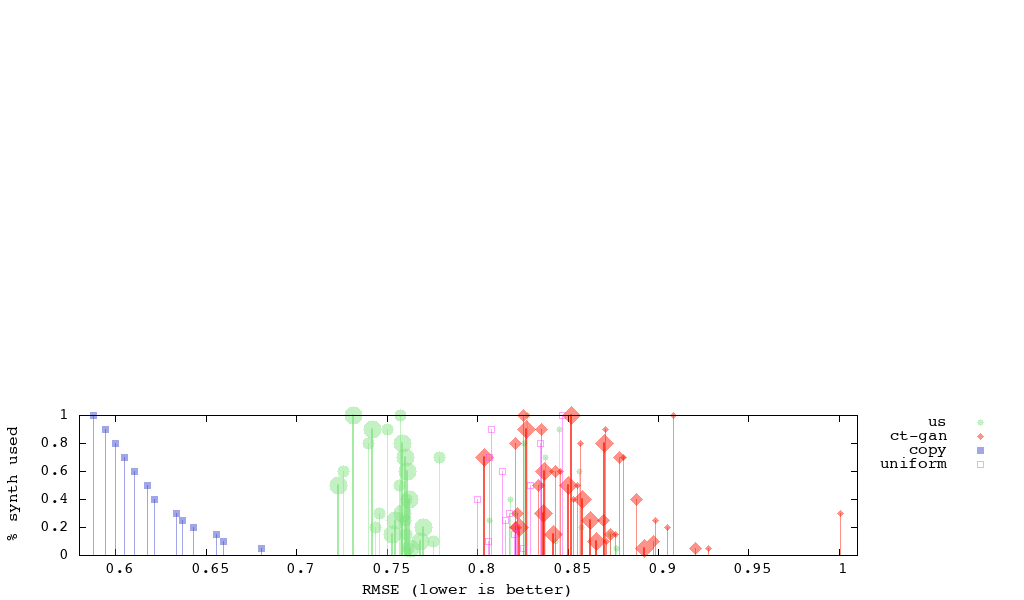} \\
  \texttt{winered}\\ \hline\hline
\end{tabular}
\caption{Experiment \textsc{synth-aug}: detailed results on three domains for which the metric is the RMSE. In each plot, the $x$ value of a vertical bar indicates a method's accuracy and the height along the $y$ axis indicates the $\%$ of real data that represents generated data used to train the final classifier (up to $100\%$). Green filled circles are GTs results, the size of the circle indicating the number of splits in the GTs (\tikzcircle[green, fill=green]{1pt} = 10, \tikzcircle[green, fill=green]{2pt} = 300, \tikzcircle[green, fill=green]{3pt} = 10K). Red filled diamonds are CT-GANs results, the size of the diamond indicating the number of epochs (\tikzdiamond[red, fill=red]{2pt} = 10, \tikzdiamond[red, fill=red]{4pt} = 300, \tikzdiamond[red, fill=red]{6pt} = 1K). Finally, empty pink squares (\tikzunif[pink]{3pt}) are \textsc{unif}'s results and filled blue squares (\tikzcopy[blue,fill=blue]{3pt}) are those of \textsc{copy}.}
    \label{tab:synth-aug-2-RMSE}
  \end{table*}

  \begin{table*}
  \centering
\begin{tabular}{c}\hline\hline
  \includegraphics[trim=0bp 0bp 0bp 300bp,clip,width=0.99\textwidth]{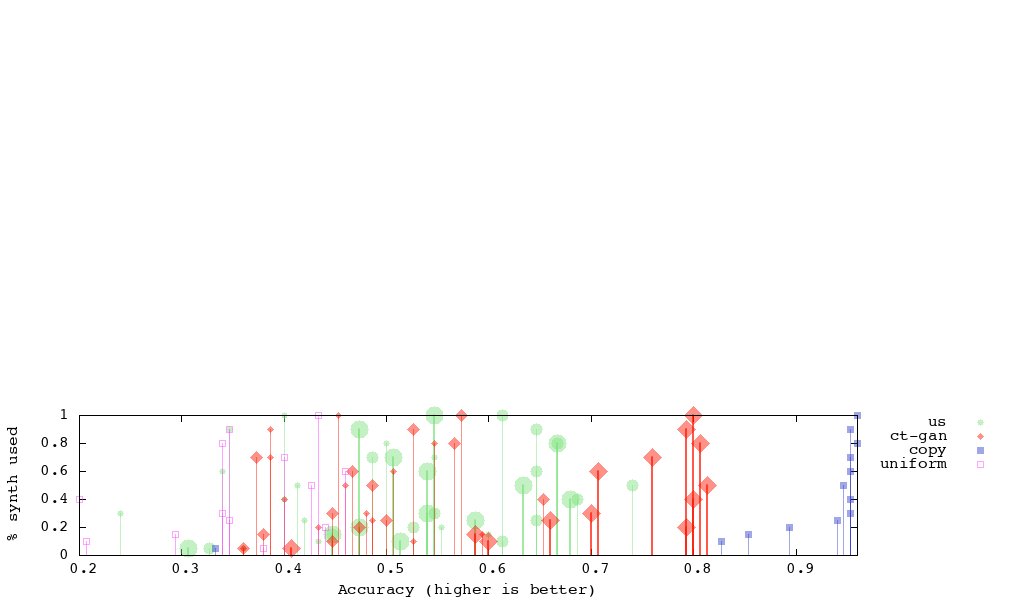} \\
  \texttt{iris}\\ \hline
  \includegraphics[trim=0bp 0bp 0bp 300bp,clip,width=0.99\textwidth]{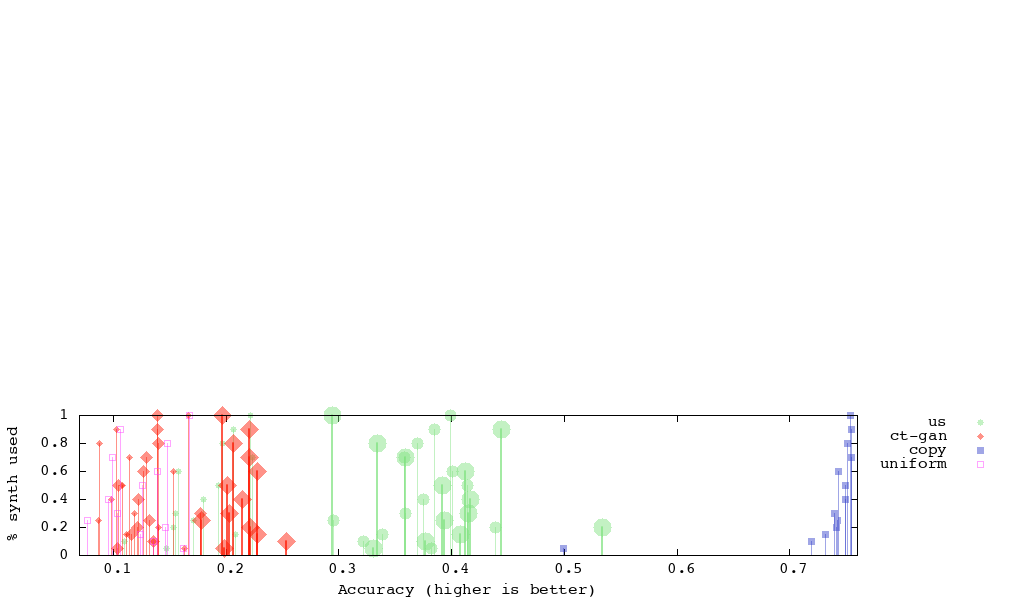} \\
  \texttt{led}\\ \hline
  \includegraphics[trim=0bp 0bp 0bp 300bp,clip,width=0.99\textwidth]{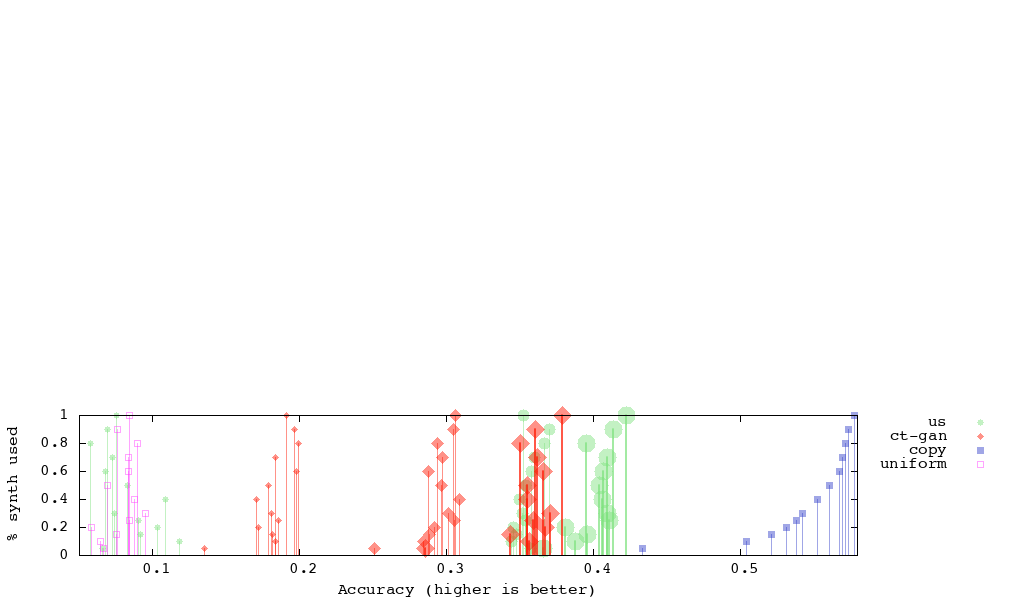} \\
  \texttt{open-policing}\\ \hline
    \includegraphics[trim=0bp 0bp 0bp 300bp,clip,width=0.99\textwidth]{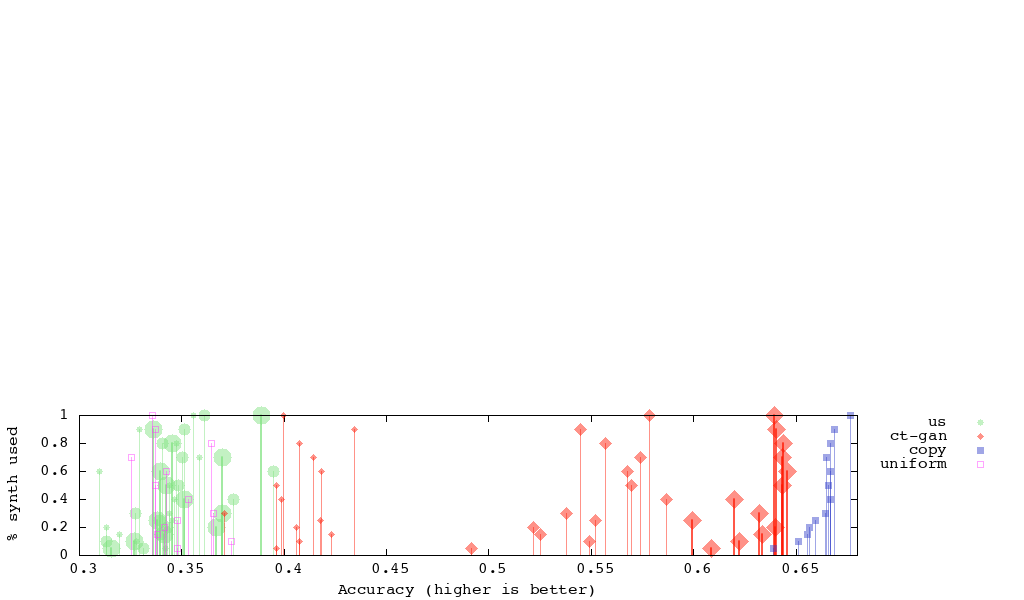} \\
  \texttt{sigma-cabs}\\ \hline\hline
\end{tabular}
\caption{Experiment \textsc{synth-aug}: detailed results on four domains for which the metric is the accuracy. Conventions follow Table \ref{tab:synth-aug-2-RMSE}.}
    \label{tab:synth-aug-2-ACC}
  \end{table*}

 \egroup
\clearpage

\noindent\textbf{Objective} Supplementing real data with additional 'faithful' generated data could be envisioned as a way to improve the performances of models trained from the whole data, a problem of substantial practical impact. Our objective was to test how generative models can perform in such a scenario.\\

\noindent\textbf{Experimental setting} The setting can be summarized as the equivalent of the \textsc{train-synth} experiment (Section \ref{sec-train-synt}), with the sole modification that we train
(supervised) models using generated data mixed with the training data of the generators instead of just the generated data alone. Equivalently, we train with \textsc{copy} + \textit{a} generated sample. This generated sample could be generated by CT-GANs or our GTs, but we also try the \textsc{copy} case (we add real examples) and the \textsc{unif}orm case (we add uniformly generated examples). We first tried the simplest experimental setting in which the size of the generated data was the same as the size of the training data, but the experiments were largely inconclusive in terms of who wins or loses. Hence, we have dug in this scenario, allowing a varying $\%$ of generated data to be added to the training data, for a $\%$ of generated data that would represent $5\%, 10\%, 15\%, ..., 100\%$ of the training (real) data. This represents a lot more experiments but also allows us to understand, for each generative technique, what is the effect of putting more generative examples in the training sample. Regardless of the comparison with the \textsc{copy} approach, a good generative approach should bring improved results as the number of generated examples increases. All other parameters, generative models and supervised classifiers considered are the same as in the \textsc{train-synth} experiment.\\

\noindent\textbf{Results} Table \ref{tab:synt-aug-1} summarises the results obtained, much in the same way as Tables \ref{tab:train-synth-3} and \ref{tab:synt-disc-3}. The results display that GTs tend to be a better fit for data augmentation than CT-GANs. Training more the neural nets does not yield an advantage over GTs, as even when comparing with GTs having a few dozen nodes, the picture is still quite balanced. Tables \ref{tab:synth-aug-2-RMSE} and \ref{tab:synth-aug-2-ACC} complete Table \ref{tab:synth-aug-2-ACC-main} for the remains domains used. We observe that while \texttt{dna} is clearly the domain where CT-GANs obtained the worst results, with \textit{all} accuracies (substantially) lower than any of \textsc{unif}, such bad performances also occur in domains \texttt{winewhite} and \texttt{winered}, which could signal that the issue is not linked to the domain being categorical (\texttt{winewhite} and \texttt{winered} are both real-valued). On \texttt{iris}, CT-GANs have a slight edge over GTs, an edge which much more significant on \texttt{sigma-cabs} where GT results basically cannot be distinguished from \textsc{unif}. We observe that in this domain, CT-GANs manage to get an improvement over $+5\%$ real data, in the same way as GTs manage to get an improvement over $+5\%$ real data on \texttt{abalone}. \texttt{open-policing}, which is our biggest domain, displays that training for a longer time is beneficial to both CT-GANs and GTs, but ultimately GTs get the best results. Two conclusions can be drawn: first, there is still work to do to beat adding real data, for both neural nets and generative trees, even when those latter models seem to overall get the best and most stable results. Second, we do not observe for generative trees the apparent overfitting pattern that follows CT-GANs on \texttt{dna}, \texttt{winewhite} and \texttt{winered}. Apart from the fact that we prevent discarding support in the induction of the GTs, we do not see currently any other reason for this observation to happen.







\end{document}